\documentclass[12pt]{article}
\usepackage{amsmath}
\usepackage{graphicx}
\usepackage{enumerate}
\usepackage{natbib}
\usepackage{tikz} 
\usepackage{url} 

\newcommand{\blind}{1}

\usepackage{latexsym, epsfig, amssymb, amsthm, mathrsfs, bbm, enumitem}
\usepackage[OT1]{fontenc}
\usepackage{xcolor}
\usepackage[colorlinks,citecolor=blue,linkcolor=blue,urlcolor=blue]{hyperref}
\usepackage{multirow,multicol,booktabs}

\usepackage[ruled,plain,linesnumbered]{algorithm2e}
\usepackage{tikz}
\usepackage{float}
\usepackage{booktabs}
\usepackage{adjustbox}
\usepackage{bbm}
\usepackage{cancel}

\DeclareMathOperator*{\argminA}{argmin}

\newcommand{\indep}{\perp \!\!\! \perp}

\DeclareMathOperator{\Var}{Var}
\DeclareMathOperator{\Cov}{Cov}

\newtheorem{assumption}{Assumption}
\newtheorem{lemma}{Lemma}
\newtheorem{proposition}{Proposition}
\newtheorem{theorem}{Theorem}
\newtheorem{definition}{Definition}
\newtheorem{remark}{Remark}
\newtheorem{example}{Example}
 
\renewcommand{\theequation}{
	\arabic{equation}%
}

\newcommand{\ignore}[1]{}{}

\newcommand{\ba}{\mbox{\bf a}}
\newcommand{\bb}{\mbox{\bf b}}

\newcommand{\be}{\mbox{\bf e}}

\newcommand{\bp}{\mbox{\bf p}}
\newcommand{\bnu}{{\bf \nu}}

\newcommand{\bu}{\mbox{\bf u}}
\newcommand{\bv}{\mbox{\bf v}}
\newcommand{\bw}{\mbox{\bf w}}
\newcommand{\bx}{\mbox{\bf x}}
\newcommand{\by}{\mbox{\bf y}}
\newcommand{\bz}{\mbox{\bf z}}
\newcommand{\bA}{\mbox{\bf A}}
\newcommand{\bB}{\mbox{\bf B}}
\newcommand{\bC}{\mbox{\bf C}}
\newcommand{\bD}{\mbox{\bf D}}

\newcommand{\bH}{\mbox{\bf H}}
\newcommand{\bM}{\mbox{\bf M}}

\newcommand{\bI}{\mbox{\bf I}}
\newcommand{\bL}{\mbox{\bf L}}

\newcommand{\bT}{\mbox{\bf T}}
\newcommand{\bU}{\mbox{\bf U}}

\newcommand{\bW}{\mbox{\bf W}}
\newcommand{\bX}{\mbox{\bf X}}
\newcommand{\bO}{\mbox{\bf O}}

\newcommand{\bY}{\mbox{\bf Y}}
\newcommand{\bZ}{\mbox{\bf Z}}
\newcommand{\bone}{\mbox{\bf 1}}
\newcommand{\bzero}{\mbox{\bf 0}}
\newcommand{\bveps}{\mbox{\boldmath $\varepsilon$}}

\newcommand{\bbeta}{\mbox{\boldmath $\beta$}}
\newcommand{\bepsilon}{\mbox{\boldmath $\epsilon$}}

\newcommand{\bdelta}{\mbox{\boldmath $\delta$}}
\newcommand{\bfeta}{\mbox{\boldmath $\eta$}}

\newcommand{\g}{\mbox{$\mathcal{G}$}}

\newcommand{\sbz}{\scriptsize\bz}

\newcommand{\Prob}{\mathbb{P}}

\newcommand{\tr}{ {\mathrm{tr}} }

\newcommand{\Expected}{\mathbb{E}}

\newcommand{\wt}{\wt}
\renewcommand{\wt}{\widetilde}

\def\independenT#1#2{\mathrel{\setbox0\hbox{$#1#2$}%
		\copy0\kern-\wd0\mkern4mu\box0}}

\renewcommand{\ldots}{\cdots}
\renewcommand{\hat}{\widehat}
\renewcommand{\tilde}{\widetilde}

\usepackage{xcolor}
\usepackage[draft,inline,nomargin,index]{fixme}
\fxsetup{theme=color,mode=multiuser}
\FXRegisterAuthor{yf}{ayf}{\color{magenta}}

\usepackage{ulem}

\graphicspath{{figs/}}

\begin{document}

\def\spacingset#1{\renewcommand{\baselinestretch}%
{#1}\small\normalsize} \spacingset{1}


\if1\blind
{
    \title{\bf HNCI: High-Dimensional Network Causal Inference%
    \thanks{
            Wenqin Du is Postdoctoral Scholar, Data Sciences and Operations Department, Marshall School of Business, University of Southern California, Los Angeles, CA 90089 (E-mail: \textit{wenqindu@marshall.usc.edu}). %
            Rundong Ding is Ph.D. candidate, Department of Mathematics, University of Southern California, Los Angeles, CA 90089 (E-mail: \textit{rundongd@usc.edu}). %
            Yingying Fan is Centennial Chair in Business Administration and Professor, Data Sciences and Operations Department, Marshall School of Business, University of Southern California, Los Angeles, CA 90089 (E-mail: \textit{fanyingy@marshall.usc.edu}). %
            Jinchi Lv is Kenneth King Stonier Chair in Business Administration and Professor, Data Sciences and Operations Department, Marshall School of Business, University of Southern California, Los Angeles, CA 90089 (E-mail: \textit{jinchilv@marshall.usc.edu}).
            This work was supported in part by NSF Grants EF-2125142 and DMS-2324490.
        }}
    \author{
        Wenqin Du, Rundong Ding, Yingying Fan and Jinchi Lv
        \vspace{0.1in}\\ 
        University of Southern California\\
        \vspace{-0.1in}
        \date{December 24, 2024}
    }
    \maketitle
} \fi

\if0\blind
{
  \bigskip
  \bigskip
  \bigskip
  \begin{center}
    {\LARGE\bf HNCI: High-Dimensional Network Causal Inference}
\end{center}
  \medskip
} \fi

\begin{abstract}	
The problem of evaluating the effectiveness of a treatment or policy commonly appears in causal inference applications under network interference. In this paper, we suggest the new method of high-dimensional network causal inference (HNCI) that provides both valid confidence interval on the average direct treatment effect on the treated (ADET) and valid confidence set for the neighborhood size for interference effect. We exploit the model setting in Belloni et al. (2022) and allow certain type of heterogeneity in node interference neighborhood sizes. We propose a linear regression formulation of potential outcomes, where the regression coefficients correspond to the underlying true interference function values of nodes and exhibit a latent homogeneous structure. Such a formulation allows us to leverage existing literature from linear regression and homogeneity pursuit to conduct valid statistical inferences with theoretical guarantees. The resulting confidence intervals for the ADET are formally justified through asymptotic normalities with estimable variances. We further provide the confidence set for the neighborhood size with theoretical guarantees exploiting the repro samples approach. The practical utilities of the newly suggested methods are demonstrated through simulation and real data examples.
\end{abstract}

\noindent%
{\it Keywords:} 
High-dimensional causal inference; 
Network interference; 
Average direct treatment effect on the treated; 
Confidence intervals; 
Neighborhood size confidence set

\spacingset{1.8} 

\section{Introduction} 
\label{section::intro}

Interaction is a common feature of complex systems and is prevalent in various domains such as social networks, economics, bioinformatics, and physical systems. 
This has sparked strong interest in the task of causal inference under network interference over the last decade, which challenges the “stable unit treatment value assumption” (SUTVA) that rules out interference in classical causal inference \citep{rubin1980randomization}.
Specifically, individual treatments may influence the outcomes of many individuals within a network, both directly and indirectly \citep{eckles2017design,li2022random}.

Various efforts have been made to accommodate network inference effect in causal inference. For example, the partial interference models have been proposed where nodes in a network are divided into groups, allowing interference within each group but not across groups \citep{liu2016inverse}.
To accommodate interference of unknown and arbitrary forms, another important stream of studies have focused on modeling interference through the exposure mapping \citep{baird2018optimal,BelloniFangVolfovsky2022,gao2023causal}.
In other studies, it is widely assumed that the outcomes of a unit in the network are affected only by its immediate neighbors \citep{sussman2017elements,awan2020almost,Jagadeesan2020Designs,forastiere2021identification,li2022random}.
However, as noted in \cite{eckles2017design}, such assumptions may be overly restrictive in practice as they only account for nodes directly connected to the ego.
Relaxing them is thus essential for improving model generality and enabling more flexible analysis of causal effects in complex networks.

To this end, the recent work of \citet{leung2022causal} allows the neighborhood size to go to infinity under an ``approximate neighborhood interference" (ANI) assumption, which states that the interference diminishes as the distance between the treated nodes and the ego increases.
We seek to balance the above approaches, where the neighborhood size is assumed to be either one or infinity, by conducting network causal inference when the depth of interference can be node-specific and may vary with the treatment assignments. 
Brief discussions on potential heterogeneity in interference effects can be found in some existing works \citep{eckles2017design,puelz2022graph}.
Our study follows the framework of \citet{BelloniFangVolfovsky2022} to exploit the heterogeneous depth of neighborhood across  units that depends on the network and treatment assignments for controlling the approximation errors. 

Specifically, we develop methods to estimate and infer the \textit{average direct treatment effect on the treated (ADET)} for network data, where the network size diverges to infinity.
We assess the estimation uncertainty in the presence of cross-unit interference and provide user-friendly practical inference procedures. A major innovation of ours is building our inference framework on a linear regression formulation of potential outcomes, where the regression coefficients exhibit latent homogeneity determined by the observed network and the unknown true interference depths of nodes. Such formulation allows us to leverage existing works in the linear regression and homogeneity pursuit literature to estimate the underlying true inference function values. Then, using a matching procedure, we obtain an estimate of ADET with confidence interval guarantees.   
Given that the interference experienced by each unit may vary with the size of its corresponding neighborhood, our approach relies only on a \textit{conservative upper bound} of the neighborhood size. 
To achieve this, we suggest an inference procedure based on the square-root fused clipped Lasso (SFL) extending the fused Lasso method \citep{shen2010grouping} 
to uncover the latent homogeneity in regression coefficients. 
While the choice of neighborhood size is rarely discussed in the literature, inspired by the work of \cite{wang2022finite} which originally targeted at model selection in high-dimensional regression, we offer an inference framework for the neighborhood size with theoretical guarantees.
This can be viewed either as a separate task of interest or as a preprocessing step in the network causal inference problem. We further validate the effectiveness of our methods through a range of numerical examples.

The rest of the paper is organized as follows. Section \ref{section::model} introduces the potential outcome model under network interference. We develop the new framework of HNCI for estimating and inferring the ADET in Section \ref{sec::inference-for-the-average-treatment-effect}. 
Section \ref{section::inference-on-the-neighborhood-size} provides an inference framework for the neighborhood size with both finite-sample and large-sample properties. Simulation and real data studies are in Sections \ref{section::numerical-experiments} and \ref{section::teenage-friends-and-lifestyle-study-analysis}, respectively. Section \ref{section::discussion} discusses some implications and extensions of our work. All the proofs, technical details, and additional numerical studies are provided in the Supplementary Material.

\section{Model setting under network interference}
\label{section::model}
\subsection{Model assumptions}
\label{section::model-assumptions}

Consider a sample of $n$ units indexed by $i \in [n] := \{1,2,\ldots,n\}$, connected through an interference network $G$, where each unit is randomly assigned a binary treatment $Z_i \sim \text{Bernoulli}(p_i)$ for some $p_i \in (0,1)$.
Let $\bz = (z_1, z_2, \cdots, z_n)^T \in \{0,1\}^n$ denote the treatment assignments, which serves as a realization of the random vector $\bZ = (Z_1, Z_2, \cdots, Z_n)^T$.
For example, $\bz$ could indicate that a tax incentive is offered to a specific subset of businesses in a region.
In the network setting, the units are referred to as nodes in $G$, which are rarely independent of each other. Hence, the effect of a tax incentive on a specific company may depend on whether its collaborators or competitors also receive the tax incentive.
As mentioned in Section \ref{section::intro}, we exploit the model framework introduced in  \cite{BelloniFangVolfovsky2022} for $n$ nodes connected through $G$.
The potential outcome of the $i$th node is defined as $Y_i := \wt Y_i(\bz) = \wt Y_i(z_i, \bz_{-i})$, where $\wt Y_i(\cdot): \{0,1\}^n \rightarrow \mathbb{R}$, and $z_i$ and $\bz_{-i}$ are the treatment assignments for the $i$th node and the remaining nodes, respectively.
In practice, we may observe node features $\{\bC_i\}_{i\in [n]}$.
Unlike \cite{li2022random}, our study will condition on $G$ and allow \textit{heterogeneous} propensity scores $p_i$ for each node, which may vary with $\bC_i$.

The direct treatment effect of each node $i \in [n]$ can have distinct values
$$ 
    \tau_i := \Expected \{\wt Y_i(1, \bzero_{-i}) - \wt Y_i(0, \bzero_{-i})|\bC_i\},
$$
where $\bzero_{-i}$ is an $(n-1)$-dimensional vector of zeros.
Throughout this paper, we assume the following unconfoundedness condition, where given the covariates, $\bZ$ is as good as random.

\begin{assumption}[Unconfoundedness] 
\label{assumption::Unconfoundedness}
For each $i \in [n]$, $
    Z_i \indep (\wt Y_i(0, \bZ_{-i}), \wt Y_i(1, \bZ_{-i})) | \bC_i 
    $.
\end{assumption}

We also impose the additivity condition on the individual direct treatment effects in the assumption below.
Intuitively, this assumption states the additivity of the direct treatment effects and peer influence effects in their contributions to the potential outcomes' mean, and that there is no interaction between nodes’ treatment indicators \citep{awan2020almost}.

\begin{assumption}[Additivity of main effects] 
\label{assumption::additivity-of-main-effects}
    For each $i \in [n]$ and $\bz\in \{0,1\}^n$, assume that 
    $
    \Expected \{\wt Y_i(1,\bzero_{-i}) - \wt Y(0, \bzero_{-i})|\bC_i\} = \Expected\{\wt Y_i(1,\bz_{-i}) - \wt Y(0, \bz_{-i})|\bC_i\}.
    $
\end{assumption}

Assumption \ref{assumption::additivity-of-main-effects} is common in the network interference literature \citep{sussman2017elements}.
Together with the fact that $\wt Y_i(\bz)=z_i\{\wt Y(z_i, \bz_{-i})-\wt Y_i(0, \bz_{-i})\} + \wt Y_i(0, \bz_{-i})$, we have
\begin{align}
    \wt Y(z_i, \bz_{-i}) = z_i(\tau_i + \epsilon_{1,i}) + \wt Y_i(0,\bz_{-i}) 
    \label{equation::y-decompose-1}
\end{align}
with $\epsilon_{1,i} = \wt Y_i(1, \bz_{-i}) - \wt Y(0, \bz_{-i}) - \Expected\{\wt Y_i(1, \bz_{-i}) - \wt Y(0, \bz_{-i})|\bC_i\}$, entailing that  $\Expected(\epsilon_{1,i}|\bC_i)=0$.

The interference network $G$ imposes constraints on how potential outcomes can vary with the treatment assignments.
In the spirit of the neighborhood interference assumption \citep{sussman2017elements,awan2020almost,BelloniFangVolfovsky2022}, we define the interference function to assess how $G$ contributes to the interference patterns in the potential outcomes. 
To achieve this, let us define $G_i^{\scriptsize\bz}(k)$ as the $k$-hop neighborhood of node $i$ in $G$ with labeled treatments $\bz \in \{0,1\}^n$. 
In this context, node $i$ is referred to as the \textit{ego}, and its $k$-hop neighborhood includes all nodes with depth from node $i$ at most $k$.
Accordingly, denote by $G_i^{\scriptsize\bz}$ the subgraph consisting of all nodes connected to the ego node $i$ and their corresponding treatments. 
For simplicity, assume that $G$ is connected in the sense that any node pair $(i, j)$ is linked by a path; our study can be generalized to the disconnected case with minimum efforts. Thus, $G_i^{\scriptsize\bz}$ represents the entire interference network under treatment assignments $\bz$, centered on ego $i$, without any constraints on the neighborhood size.

\begin{remark}
\label{remark::def-subgraph}
    We provide some discussions on $G_i^{\scriptsize\bz}$ with ego $i$.   
    Let $d_{i}(j)$ be the depth of node $j$ in $G_i^{\scriptsize\bz}$. 
    By definition, we have $d_{i}(j)\leq n$ for all $i,j\in [n]$ and the maximum depth $\max_{j \in [n]}d_{i}(j)$ can vary with $i$. 
    Consequently, $G_i^{\scriptsize\bz}(k)$ is well-defined only if $k\leq \max_{j \in [n]}d_{i}(j)$. To simplify the presentation, we set $G_i^{\scriptsize\bz}(k) = G_i^{\scriptsize\bz}(\max_{j \in [n]}d_{i}(j))$ for all $k>\max_{j \in [n]}d_{i}(j)$ so that $G_i^{\scriptsize\bz}(k)$ is well-defined for all nodes $i\in [n]$ and neighborhood sizes $k\in [n]$.  
\end{remark}

\begin{figure}[h!]
    \centering
    \includegraphics[width=0.75\textwidth]{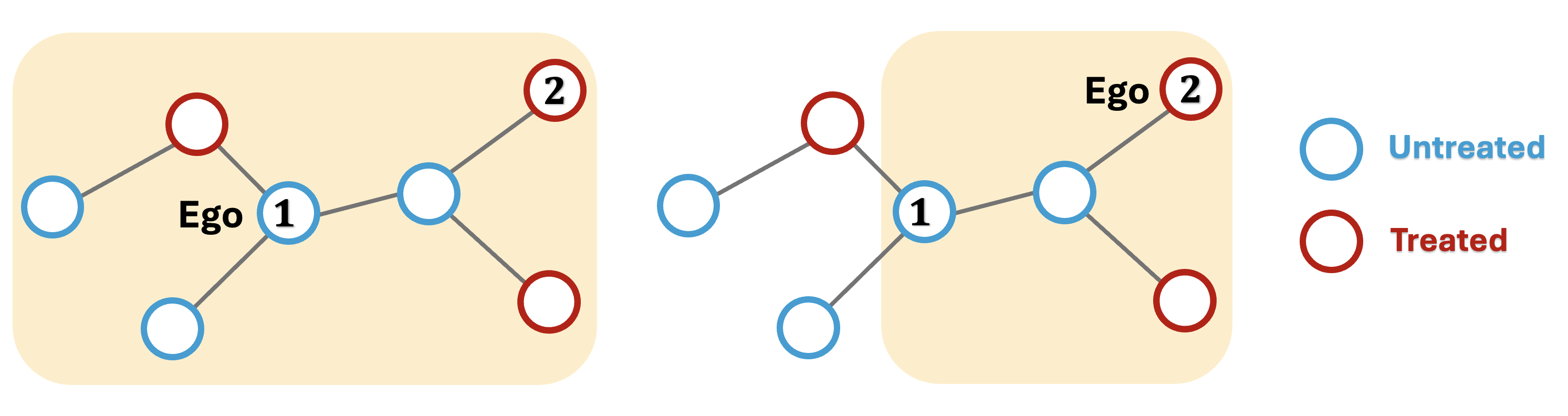}
    \vspace{-1.5em}
    \caption{An illustration network with $2$-hop neighborhood of nodes $1$ and $2$.
    }
    \label{fig::eg-heter-ki}
\end{figure}

To help understand Remark \ref{remark::def-subgraph}, we consider a small network illustration in Figure \ref{fig::eg-heter-ki}.
In this network, nodes $1$ and $2$ have the maximum depths equal to $2$ and $4$, respectively. By definition, $G_1^{\sbz}(2)$ and $G_2^{\sbz}(2)$ are the colored regions in the left and right plots of Figure \ref{fig::eg-heter-ki}, respectively.
Remark \ref{remark::def-subgraph} ensures that $G_1^{\sbz}(k)$ is well-defined and stays the same for all $k \geq 2$.

Note that the neighborhood size that can affect the outcome $Y_i$ is generally \textit{unknown} in practice.
As discussed in Section \ref{section::intro}, many existing works address the special case when $k=1$.
Yet, in practice, the neighborhood size could exceed $1$ and vary across nodes \citep{BelloniFangVolfovsky2022,leung2022causal}. 
This motivates us to consider the general scenario which does \textit{not} assume $k=1$ or known. 
Next, we introduce the neighborhood interference assumption.

\begin{assumption}[Neighborhood interference]
\label{assumption::NI} 
    There exists a known mapping $\gamma_0(\cdot)$ satisfying the \textit{nested matching} property that $\gamma_0(G_i^{\sbz}(k))=\gamma_0(G_j^{\sbz}(k))$ implies $\gamma_0(G_i^{\sbz}(k'))=\gamma_0(G_j^{\sbz}(k'))$ for all $k' \in [k]$. With such $\gamma_0(\cdot)$, for some unknown $k_0\in \mathbb N$ it holds that
    \begin{itemize}
    \item[i)]
    $\Expected \{\wt Y_i(0, \bz_{-i})\}
    = f\big(\gamma_0(G_i^{\sbz}(k_0))\big)$ for all $i \in [n]$ and some interference function $f(\cdot)$; 
     \item[ii)] for any treatment assignment vectors $\bz$ and $\bz'$ with $\gamma_0(G_i^{\sbz}(k_0)) = \gamma_0(G_i^{\sbz'}(k_0))$, we have
    $
    \Expected \{\wt Y_i(1, \bz_{-i})\} = \Expected \{\wt Y_i(1, \bz'_{-i})\}$ and $ \ \Expected \{\wt Y_i(0, \bz_{-i})\} = \Expected \{\wt Y_i(0, \bz'_{-i})\}.
    $ 
    \end{itemize}
    If $k_0$ is not unique, we define it as the smallest neighborhood size that satisfies $i)$ and $ii)$.
\end{assumption}

Observe that the interference function $f(\cdot)$ is independent of node labels so the $k_0$-hop neighborhood of each node is sufficient to determine the interference function values of nodes.  
Assumption \ref{assumption::NI} ensures that the expected potential outcome for node $i$ remains unchanged if the treatments assigned to $i$ and its $k_0$-hop neighbors are fixed, regardless of changes in other nodes' treatments. This motivates the matching idea based on the $k_0$-hop neighborhood of nodes, detailed in Section \ref{sec::matching-estimator}.
Although $k_0$ is independent of node labels, Assumption \ref{assumption::NI} can accommodate certain type of heterogeneity in the true neighborhood size that can affect the potential outcome $\tilde Y_i(0,\bz_{-i})$'s. We illustrate this with two examples.

\begin{example}\label{example::hetero-1}
    Assume that $\Expected \{\wt Y_i(0, \bz_{-i})\} = f\big(\gamma_0(G_i^{\sbz}(k_{0,i}))\big)$  for $k_{0,i} = \max_{j \in [n]} d_i(j)$, the maximum node depth in $G^{\sbz}_i$. Let $k_0 =\max_{i\in[n]} k_{0,i}$. Then with the convention specified in Remark \ref{remark::def-subgraph}, it holds that $k_0\geq k_{0,i}$ and $\gamma_0(G_i^{\sbz}(k_0)) = \gamma_0(G_i^{\sbz}(k_{0,i}))$. 
    It is seen that we have heterogeneous neighborhood sizes because $k_{0,i} = \max_{j\in[n]} d_i(j)$ can vary from node to node. 
\end{example}

\begin{example}\label{example::hetero-2}
    Assume that $\Expected \{\wt Y_i(0, \bz_{-i})\} = f\big(\gamma_0(G_i^{\sbz}(k_{0,i}))\big)$ for some $k_{0,i}\in \mathbb N$.
    A slightly different version of Example \ref{example::hetero-1} above is that for all nodes $i$ with $k_{0,i} = \max_{j\in[n]} d_i(j)$, it satisfies that $k_{0,i}\leq k_0$, while for all nodes $i$ with $ \max_{j\in[n]} d_i(j)> k_{0,i}$, it holds that $k_{0,i}\equiv k_0$. It is seen that here, some nodes are allowed to have heterogeneous neighborhood sizes.  
\end{example}

To summarize, a \textit{sufficient} condition for nodes to have heterogeneous neighborhood sizes under Assumption \ref{assumption::NI} is that $\gamma_0(G_i^{\bz}(k_{0,i})) = \gamma_0(G_i^{\bz}(k))$ for all $k_{0,i}\leq k\leq k_0$, where $i\in [n]$ with $k_{0,i}<k_0$.  
Examples \ref{example::hetero-1} and \ref{example::hetero-2} above enforce this sufficient condition by directly setting $G_i^{\bz}(k) = G_i^{\bz}(k_{0,i})$ for all $k_{0,i}\leq k\leq k_0$ and all $i\in [n]$ with $k_{0,i}<k_0$ (if any). Since mapping $\gamma_0(\cdot)$ can be many-to-one, such sufficient condition accommodates broader scenarios than those covered in Examples \ref{example::hetero-1} and \ref{example::hetero-2}.

We next provide an example on the mapping $\gamma_0(\cdot)$ and the interference function $f(\cdot)$. 

\begin{example}
\label{eg::mapping-treated-nodes}
    For any interference network $G_i^{\sbz}(k)$, let $\gamma_0(\cdot)$ be the vector of proportions of treated nodes at each depth $d\in [l]$, where $k$ and $l$ are potentially different integers. Specifically, $\gamma_0(G_i^{\sbz}(k)) = (T_{i,1}, T_{i,2}, \cdots, T_{i,l})$, where $T_{i,j}$ is the proportion of treated nodes at depth $j$ in $G_i^{\sbz}(k)$. We adopt the convention $0/0=0$ so if $l> \max_{j\in[n]} d_i(j)$, then  $T_{i,l}=0$.

    We connect to the example in Figure \ref{fig::eg-heter-ki} to ease the understanding. Assume that $k_0=l=3$ and consider $G^{\sbz}_i(k_0)$ for nodes $i=1$ and $2$. 
    By definition, we have $(T_{1,1}, T_{1,2}, T_{1,3})=(1/3,2/3,0)$ for node $1$, and $(T_{2,1}, T_{2,2}, T_{2,3})=(0,1/2,1/2)$ for node $2$.
    An example of interference function $f(\cdot):\mathbb{R}^{l} \rightarrow \mathbb{R}$ is given by $f\big(\gamma_0(G_i^{\sbz}(k_0))\big)=\sum_{j=1}^{l} T_{i,j} / 2^j$, which models the diminishing influence of treated nodes on node $i$ as their depth increases.
\end{example}

\begin{assumption}[Constant baseline]
\label{assumption::constant-baseline}
    For each $\bz \in \{0,1\}^n$ and $i \in [n]$, assume that $\Expected\{\wt Y_i(0, \bz_{-i})\} = \Expected\{\wt Y_i(0, \bz_{-i})|\bC_i\}$.
\end{assumption}

Assumption \ref{assumption::constant-baseline} above was introduced in \cite{BelloniFangVolfovsky2022} for ruling out the effect of covariates on the potential outcomes of untreated nodes, motivated by situations when a relationship between $\bC_i$ and $G$ is expected.
We refer to \citet{BelloniFangVolfovsky2022} for further discussions on this assumption.
From Assumptions \ref{assumption::NI} and \ref{assumption::constant-baseline}, it follows that $\wt Y_i(0, \bz_{-i}) = f\big(\gamma_0(G_i^{\bz}(k_0))\big) + \epsilon_{2,i}$, where $\epsilon_{2,i} = \wt Y(0, \bz_{-i}) - \Expected\{ \wt Y(0, \bz_{-i})\}$.
Combining this with \eqref{equation::y-decompose-1} and letting $\epsilon_{i}= z_i\epsilon_{1,i} + \epsilon_{2,i}$, we can obtain that
\begin{align} \label{def::potential-outcome-model}
    \wt Y_i(z_i, \bz_{-i}) = z_i\tau_i + f\big(\gamma_0(G_i^{\bz}(k_0))\big) + \epsilon_{i},
\end{align}
which implies that $\epsilon_{i} = \wt Y(0, \bz_{-i}) - \Expected\{\wt Y(0, \bz_{-i})\}$ when $z_i=0$ and $\epsilon_{i} = \wt Y(1, \bz_{-i}) - \Expected\{\wt Y(1, \bz_{-i})|\bC_i\}$ when $z_i=1$.

Under Assumptions \ref{assumption::Unconfoundedness}--\ref{assumption::constant-baseline}, the potential outcome model can be characterized as follows, with additional regularity condition on the error terms.

\begin{definition}[Potential outcome model]
\label{definition::potential-outcome-model}
    The potential outcome model satisfies \eqref{def::potential-outcome-model}, where $\{|\tau_i|+|f\big(\gamma_0(G_i^{\bz}(k_0))\big)|\}_{i \in [n]}$ are uniformly bounded for all $\bz\in \{0,1\}^n$ and the overlapping condition $c \leq \Prob\big(Z_i=1|\gamma_0(G_i^{\bz}(k_0)), \bC_i\big) \leq 1-c$ holds for some $c \in (0, 1)$.
\end{definition}

We will work under the model in Definition \ref{definition::potential-outcome-model} for the remaining part of this paper to estimate and infer the average direct treatment effect on the treated (ADET)
\begin{align}
    \tau : = \frac{1}{\sum_{i=1}^n Z_i}\sum_{i=1}^nZ_i \Expected\{\wt Y_i(1, \bzero_{-i}) - \wt Y_i(0, \bzero_{-i})|\bC_i\} = \frac{1}{\sum_{i=1}^n Z_i}\sum_{i=1}^nZ_i\tau_i,
    \label{def::average-treatment-effect}
\end{align}
which represents the average incremental response of treated units to their own treatments. 
Note that $\tau$ is defined with respect to the treatment assignments and thus we consider the estimation and inference problem conditional on $\bZ$.

Although $\tau_i$ and $\bZ_i$ may depend on $\bC_i$'s, model \eqref{def::potential-outcome-model} and $\tau$ do not directly include $\bC_i$'s as feature vectors. Therefore, our study does not directly utilize the node covariates $\bC_i$'s. Additionally, since the propensity score estimation has been a well-studied problem with many available methods, we consider settings with known propensity scores, for example, from the experimental design, to simplify the technical presentation. We note that our method and theory can be easily extended to accommodate estimated propensity scores.

\subsection{The matching estimator}
\label{sec::matching-estimator}

In this subsection, we outline our estimation and inference method for ADET $\tau$ in \eqref{def::average-treatment-effect}.
Inspired by \cite{BelloniFangVolfovsky2022}, our approach is rooted in a matching procedure.
Specifically, we estimate interference function values using untreated nodes and match them to treated ones based on their features, i.e., $\{\gamma_0(G_i^{\bz}(k_0))\}$.
This procedure provides estimates for $\tau_i$'s on treated nodes, which we substitute into \eqref{def::average-treatment-effect} to estimate $\tau$.
A key innovation of our method is framing the interference function estimation as a homogeneity pursuit problem in the linear regression setting. 
The new formulation allows us to leverage existing methods to address the underlying homogeneity structure of the interference function values.

For the matching procedure outlined above to work, we introduce the balanced feature assumption: the features of untreated nodes should be sufficiently rich to ensure each treated node can be matched to some untreated ones based on the features; that is, 
\begin{align}
    \{\gamma_0(G_i^{\sbz}(k_0)): z_i=1, i\in [n]\}\subset \{\gamma_0(G_i^{\sbz}(k_0)): z_i=0, i\in [n]\}.
    \label{equ::match-feature}
\end{align}
For sufficiently large networks, condition \eqref{equ::match-feature} is easily satisfied with high probability (with respect to the randomness in $\bZ$) under the overlapping condition in Definition \ref{definition::potential-outcome-model}.
Since $k_0$ is unknown in practice, we consider a slightly stronger condition than \eqref{equ::match-feature}.
To formalize this, we first introduce some necessary notation.
For each fixed $k$ and given $\gamma_0(\cdot)$, let
\begin{align}
      \hat E_{k} := \{g: \text{there exists some } i\in[n] \text{ such that } z_i=0, g=\gamma_0(G_i^{\bz}(k)) \}
     \label{def::E-hat}
\end{align}
denote the collection of distinct features of untreated nodes based on the $k$-hop neighborhood and let $d(k) :=|\hat E_{k}|$.
In general, the number of distinct features $d(k)$ is no larger than $n$ but can diverge with $n$. We are now ready to introduce the balanced feature assumption. 

\begin{assumption}[Balanced features] 
\label{assumption::Balanced-features}
    For any treated nodes with $z_i=1$ and any neighborhood size $k$ of interest, it holds that $\gamma_0(G_i^{\bz}(k)) \in \hat E_{k}$.
\end{assumption}
      
Given treatment assignments $\bz$, let $\{g_{k,1}, g_{k,2},\cdots, g_{k,d(k)}\}$ be the elements in $\hat E_{k}$. 
Define $S_{k,l} := \{j \in [n]: \gamma_0(G_j^{\bz}(k))=g_{k,l}, z_j=0\}$ as the set of untreated nodes with feature $g_{k,l}$ for $l\in [d(k)]$.
Then, $S_k:=\{S_{k,l}, l\in [d(k)]\}$ forms a partition of the untreated node set $\{j\in [n]: z_j=0\}$. 
By the nested matching property in Assumption \ref{assumption::NI}, partitions $S_k$'s have a hierarchical structure in that sets in $S_{k+1}$ are obtained by splitting some sets in $S_{k}$, and that larger $k$ gives a finer partition of the untreated node set. 
For each fixed $k \geq k_0$, partition $S_k$ implants a homogeneity structure in untreated nodes, with nodes in the same set $S_{k,l}$ sharing the same interference function value $f(\gamma_0(g))$ for $g \in \hat E_k$. 
When $k = k_0$, such homogeneity reflects the underlying ground truth based on network features.

Under Definition \ref{definition::potential-outcome-model}, for untreated nodes, we have  
$
    \wt Y_i(z_i, \bz_{-i}) = f\big(\gamma_0(G_i^{\bz}(k_0))\big) + \epsilon_i,
$
where $\epsilon_i \overset{\text{i.i.d.}}{\sim} N(0, \sigma_0^2)$.
However, for each $k > k_0$ and $i \in [n]$, $f\big(\gamma_0(G_i^{\bz}(k))\big)$ and $f\big(\gamma_0(G_i^{\bz}(k_0))\big)$ can differ in general.
Hence, the oracle interference function value of any node with feature $g_{k,l} \in \hat E_{k}$ depends on its $k_0$-hop neighborhood.
Specifically, for each $k\geq k_0$, let $\beta_{k,l}^0 := f\big(\gamma_0(G_j^{\bz}(k_0))\big)$ for any $j \in S_{k,l}$, where uniqueness of $\beta_{k,l}^0$ is guaranteed by Assumption \ref{assumption::NI}.
Then we define the vector of true interference function values over the node partition $S_k$ as 
\begin{align} 
    \bbeta_{k}^0 = (\beta_{k,1}^0, \beta_{k,2}^0, \cdots, \beta_{k,d(k)}^0)^T.
    \label{def::interaction-function-of-ture-k0}
\end{align}
It is seen that $\bbeta_{k}^0$ is a subvector of $\bbeta_{k'}^0$ for all $k' > k \geq k_0$. 
Denote by $n_0$ the total number of untreated nodes, which is a fixed quantity conditional on the treatment assignments $\bz$ of $n$ nodes.
Additionally, define $\bX_{k}\in\mathbb R^{n_0\times d(k)}$ as a matrix with the $i$th row the standard basis vector $\be_l$ if $i\in S_{k,l}$.
With such notation, for all $k \geq k_0$, the response vector $\by_{obs}\in \mathbb R^{n_0}$ of untreated nodes can be rewritten in the form of a linear regression model 
\begin{equation}
\label{equation::untreated-nodes-model}
  \by_{obs} = \bX_{k}\bbeta_{k}^0 + \bveps_0,
\end{equation}
where $\bveps_0 \sim \mathcal{N}(0, \sigma_0^2 \bI_{n_0})$ with $\bI_{n_0}$ denoting the $n_0 \times n_0$ identity matrix.

Model \eqref{equation::untreated-nodes-model} above allows us to utilize the existing results in the linear regression literature. 
Given an estimate $\hat \bbeta_{k}=(\hat\beta_{k,1}^0, \cdots, \hat\beta_{k,d(k)}^0)^T$ of $\bbeta_{k}^0$, it follows from Assumption \ref{assumption::Balanced-features} that for each treated node $j \in [n]$, there exist some $l \in [d(k)]$ and $g_{k,l} \in \hat E_{k}$ such that $\gamma_0(G^{\sbz}_j(k)) = g_{k,l}$.
This motivates us to estimate the interference function value $f\big(\gamma_0(G^{\sbz}_j(k_0))\big)$ of node $j$ as $\hat \beta_{k,l}^0$.
Thus, for each treated node $j$, we can form an estimate $\hat f_{j,k}$ of $f\big(\gamma_0(G^{\sbz}_j(k_0))\big)$. 

\begin{remark}[Matching procedure]
\label{remark::matching-procedure}
    Given an estimate $\hat \bbeta_{k}$ of $\bbeta_{k}^0$, 
    for each treated node $i$ with $z_i=1$, we match $\gamma_0(G_i^{\bz}(k)) = g_{k,l}$ for $l \in [d(k)]$ and set $\hat f_{i,k} = \hat \beta_{k,l}$. 
\end{remark}

The matching procedure introduced in Remark \ref{remark::matching-procedure} above is valid under Assumption \ref{assumption::Balanced-features}. 
When this assumption does not hold in practice, the trimming technique \citep{crump2009dealing,d2021overlap} can be employed. However, this will change the population distribution of data and hence affect the definition of the ADET. 
This is beyond the scope of our current paper and will be investigated in future work.

To estimate $\tau$, we exploit the popularly used outcome regression (OR) estimator and the doubly robust (DR) estimator \citep{robins1994estimation}, which, respectively, take the forms 
\begin{align} \label{equation::def-OR}
    \hat \tau^{OR} =&~ \frac{1}{\sum_{i=1}^n Z_i}\sum_{i=1}^nZ_i(Y_i - \hat f_{i,k}),\\ \label{equation::def-DR}
    \hat \tau^{DR} =&~ \frac{1}{\sum_{i=1}^n Z_i}\sum_{i=1}^n\Big\{Z_i(Y_i - \hat f_{i,k}) - \frac{(1-Z_i)(Y_i-\hat f_{i,k}) p_i }{1-p_i}\Big\}.
\end{align}

It is well-known that the DR estimator is consistent for $\tau$ if either the propensity score or the potential outcome model is correctly specified.
Hence, it is more robust in terms of bias if we do not have \textit{exact matching} in model \eqref{def::potential-outcome-model} but $\{p_i\}$ is correctly specified \citep{tan2006distributional}. 
Here, ``no exact matching" refers to settings where the interference function values of nodes with the same feature are \textit{close} but not necessarily identical. As noted in Section \ref{section::model-assumptions}, we assume that $\{p_i\}$ is given; the case when $\{p_i\}$ is unknown but can be well estimated can be accommodated by straightforwardly extending our method and theory. 

We conclude this section with some additional discussions. 
Although the linear representation \eqref{equation::untreated-nodes-model} holds for all $k\geq k_0$ and the dimensionality of $\bbeta^0_{k}$ is generally smaller than $n_0$, the dimensionality may still be unnecessarily high because of two reasons: i) the working parameter $k$ may exceed the ground truth $k_0$, and ii) the unknown function $f(\cdot)$ can be many-to-one. 
This implies potential latent homogeneity structure in $\bbeta^0_{k}$ and the OLS estimator may not produce a most efficient estimate of the vector of interference function values, making the estimator of the ADET based on the matching procedure inefficient. 
We illustrate the above reasoning i) in Section \ref{section::discussion-on-k0s-role-in-inferring-tau}, where the effect of using a neighborhood size that differs from $k_0$ will be investigated in a simplified model setting.
To leverage the potential homogeneity structure in $\bbeta^0_{k}$, we will adapt existing methods from the literature on homogeneity pursuit, as detailed in Section \ref{section::the-general-prior-assisted-inference-framework}. 
Moreover, we will suggest in Section \ref{section::inference-on-the-neighborhood-size} a method for inferring $k_0$, which can serve as a preprocessing step for ADET analysis.

\section{Inference on the ADET}
\label{sec::inference-for-the-average-treatment-effect}

\subsection{Understanding the role of $k_0$ in inferring $\tau$}
\label{section::discussion-on-k0s-role-in-inferring-tau}

We study in this subsection a simplified setting where $\tau_i = \tau$ for all $i \in [n]$ in model \eqref{def::potential-outcome-model} to explore the impact of neighborhood size on inferring $\tau$ under the OLS framework. The general case of heterogeneous $\tau_i$'s will be investigated in Section \ref{section::the-general-prior-assisted-inference-framework}. Here, we consider
\begin{align}
\label{def::potential-outcome-model-1-simplified}
    \wt Y_i(z_i, \bz_{-i}) = Z_i\tau + f\big(\gamma_0(G_i^{\bz}(k_0))\big) + \epsilon_i, \text{  where } \epsilon_i \overset{i.i.d.}{\sim} N(0, \sigma^2_0).
\end{align}

For any given $k$, we form matrix $\wt\bX_{k} \in \mathbb R^{n\times d(k)} $ analogous to $\bX_{k}$ in model \eqref{equation::untreated-nodes-model}, with the difference that all $n$ nodes are considered here.
If $k\geq k_0$, model \eqref{def::potential-outcome-model-1-simplified} can be written as
\begin{align}
\label{def::potential-outcome-model-1-simplified-matrix-form}
    \wt\by_{obs} =
      \tau \bz  
    + \wt\bX_{k}
        \bbeta_{k}^0
    + \bepsilon,
\end{align}
where $\wt\by_{obs}=(y_1, y_2, \cdots, y_n)^T$ is the vector of potential outcomes and $\bepsilon \in \mathbb{R}^n$.
This suggests that the ADET $\tau$ can be estimated by the OLS regression. In practice, $k_0$ is unknown. For a given $k$ that may be different from $k_0$, denote by $\hat \tau(k) = \be_1^T (\bD_k^T\bD_k)^{-1}\bD_k^T \wt\by_{obs}$ the OLS estimate of $\tau$, where $\bD_k = \big[ \bz ~ \wt\bX_{k} \big]$ is the augmented design matrix and $\be_1$ is the standard basis vector.      
The result in the proposition below shows that an under-specified neighborhood size $k$ (i.e., $k<k_0$) yields a biased estimate of $\tau$, while an over-specified neighborhood size (i.e., $k>k_0$) leads to larger uncertainty of the OLS estimator. 

\begin{proposition}
\label{proposition::infer-tau-model1-overfit-underfit}
    Under model \eqref{def::potential-outcome-model-1-simplified}, estimator $\hat \tau(k)$ for $\tau$ is unbiased only if $k \geq k_0$. 
    When $\kappa d(k)^{3/2} \rightarrow 0$ with $\kappa$ the maximum diagonal entry of $\bD_{k} (\bD_{k}^T \bD_{k})^{-1} \bD_{k}^T$, the asymptotic level $(1-\alpha)$ confidence interval (CI) for $\tau$ is $\hat \tau(k) \pm \boldsymbol{\Phi}^{-1}(1-\alpha/2) \sqrt{\hat\sigma^2_k}$, where
    \begin{align*}
         \hat\sigma^2_k := \hat\sigma^2_0 \be_1^T(\bD_k^T\bD_k)^{-1}\be_1 \text{ and } \hat\sigma^2_0 = \wt\by_{obs}^T (\bI_n -  \bD_k(\bD_k^T\bD_k)^{-1}\bD_k^T) \wt\by_{obs} / (n-d(k)-1),
    \end{align*}
    for each $\alpha \in (0,1)$. Here, $d(k)+1$ represents the number of columns in $\bD_{k}$ and $\boldsymbol{\Phi}(\cdot)$ is the cumulative distribution function of the standard normal distribution.
    For any $k_1 > k_2 \geq k_0$, when $\bD_{k_1}$ and $\bD_{k_2}$ are of full column rank, we have $\Expected(\hat\sigma^2_{k_1}) \geq \Expected(\hat\sigma^2_{k_2})$.
\end{proposition}

Proposition \ref{proposition::infer-tau-model1-overfit-underfit} above is built upon the results in \cite{yohai1979asymptotic} for the OLS estimator under diverging dimensionality.
To gain insights, let us consider a simple numerical example where 
we simulate data from model \eqref{def::potential-outcome-model-1-simplified} with $z_i \overset{i.i.d.}{\sim} \text{Bernoulli}(0.2)$ and $\epsilon_i \overset{i.i.d.}{\sim} N(0, 0.2^2)$. Graph $G$ has $1000$ nodes generated from the Erd{\H{o}}s--R{\'e}nyi (ER) model \citep{erdds1959random} with connectivity probability $0.005$. 
We set $\tau=0.6$ and define $\gamma_0(G_i^{\bz}(m))=(p_{i,1}, p_{i,2}, \ldots, p_{i,m}) $, where $p_{i,l}$ is the proportion of treated nodes at depth $l$ in $G^{\sbz}_i$ for $l=1, \cdots, m$. 
The interference function is $f\big(\gamma_0(G_i^{\bz}(3))\big)=\sum_{\ell=0}^3 (1/2)^\ell ~ \wt p_{i,\ell}/ \max\{p_{i,\ell}\}$ with $\wt p_{i,\ell}=\lceil (p_{i,\ell}/0.05) \rceil$.
By construction, we have $k_0=3$.
Notice that the interference function $f(\cdot)$ is unknown when inferring $\tau$.
We choose $\alpha=0.05$ as the significance level.
Since our aim is to study how varying neighborhood size affects the inference outcomes for $\tau$, we experiment with different values of $k$ in $\{0,1,2,3,4,5,6\}$ as the working values for $k_0$. 
Table \ref{tab::model-1-simulation-infer-tau} presents the results across $1000$ replications, each involving the generation of $G$ and treatment assignments.

\begin{table}[h]
    \caption {Empirical coverage probability and average width of 95\% CI under different values of $k$. The standard error of the average CI width is included in parentheses.}
    \vspace{0.5em}
    \centering
    \resizebox{\linewidth}{!}{
    \begin{tabular}{lccccccc}\toprule
    $k$ & 0 & 1 & 2 & 3 & 4 & 5 & 6 \\ \midrule 
    Coverage & 0.862 & 0.882 & 0.920 & 0.957 & 0.949 & 0.952 & 0.940 \\ 
    Width & 0.336($4\times10^{-4}$) & 0.129($6\times10^{-4}$) & 0.080($2\times10^{-4}$) & 0.072($8\times10^{-5}$) & 0.078($9\times10^{-5}$) & 0.083($1\times10^{-4}$) & 0.109($2\times10^{-4}$) \\ \bottomrule
    \end{tabular}
    }
\label{tab::model-1-simulation-infer-tau}
\end{table}

When $k=k_0=3$, the coverage probability aligns with the nominal level, exhibiting the smallest average CI width and  standard error. 
In contrast, under-specified neighborhood size ($k < 3$) results in poor coverage and much wider CIs. 
For $k > k_0$, the average CI width gradually increases and the standard error of CI width increases as $k$ deviates from $k_0$.

\begin{figure}[H]
    \centering
    \makebox[0pt][l]{\raisebox{0.7\height}{\hspace{-1.2em}\textbf{\footnotesize{(a)}}}}
    \includegraphics[width=0.547\textwidth]{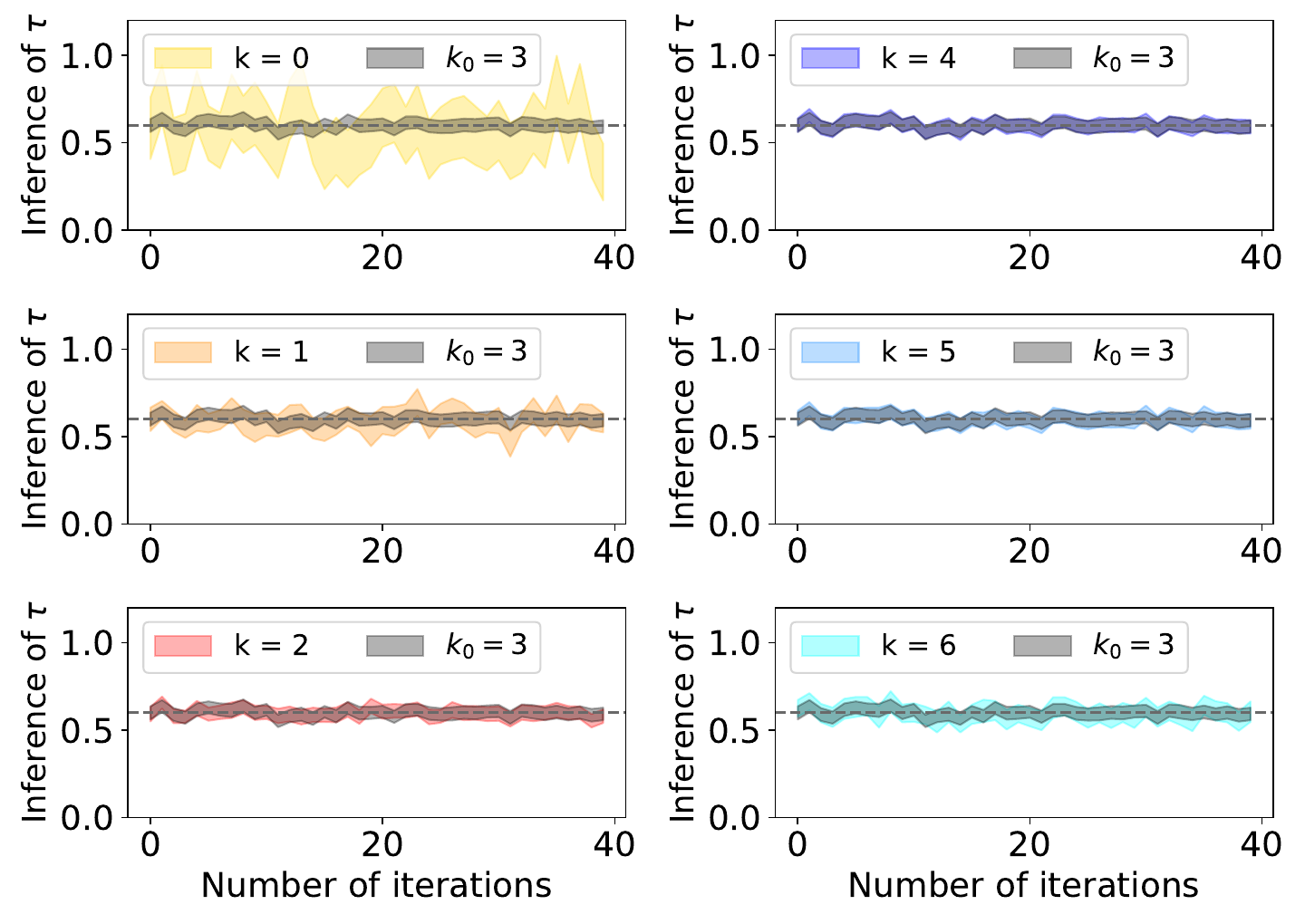}
    \hspace{1em}
    \makebox[0pt][l]{\raisebox{0.5\height}{\hspace{-1.2em}\textbf{\footnotesize{(b)}}}}
    \includegraphics[width=0.31\textwidth]{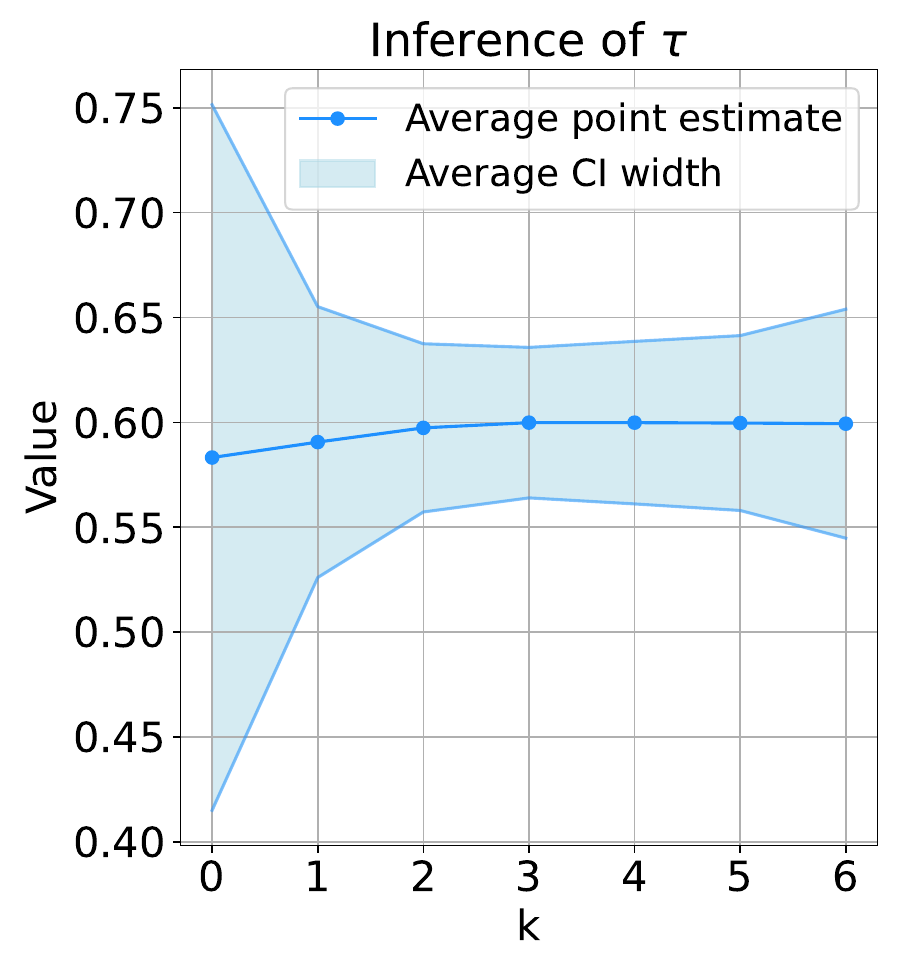}
    \vspace{-1em}
    \caption{(a) Confidence intervals across the first $40$ replications with $k_0=3$ and $\tau=0.6$. (b) Performance of the OLS estimator under varying $k$ values with $k_0=3$. For different values of $k$, the band is centered at the average point estimate of $\tau$ across $1000$ replications, and the corresponding width represents the average width of 95\% confidence intervals.}
    \label{fig::model1-CIs}
\end{figure}

Figure \ref{fig::model1-CIs} plots CIs for the first 40 replications and depicts the average width of CIs centered at the average point estimate of $\tau$ across $1000$ replications. 
It is seen that the OLS estimator is biased when $k < k_0$, but the bias diminishes when $k \geq k_0$, consistent with Proposition \ref{proposition::infer-tau-model1-overfit-underfit}. 
However, such reduction in bias when $k > k_0$ comes at the cost of wider CIs.
In general, the inference performance of CIs remains robust when $k$ is close to $k_0$.

\subsection{The general prior-assisted inference framework}
\label{section::the-general-prior-assisted-inference-framework}

As shown in Section \ref{section::discussion-on-k0s-role-in-inferring-tau}, the over-specified neighborhood size is of less concern than the under-specified case under the simplified model \eqref{def::potential-outcome-model-1-simplified}.
Motivated by such insight, we suggest the ``prior-assisted" approach for inferring $\tau$ under the general model \eqref{def::potential-outcome-model}, assuming that we know a conservative prior $k$ satisfying $k \geq k_0$.
Practical guidance on selecting $k$ with finite-sample guarantees will be provided in Section \ref{section::inference-on-the-neighborhood-size}.

Recall the high-level ideas behind constructing $\hat\tau^{OR}$ and $\hat\tau^{DR}$ that are outlined in Section \ref{sec::matching-estimator}. 
In this subsection, we will focus on introducing our suggested method for estimating the interference function values, and the subsequent estimation and inference of the ADET based on \eqref{equation::def-OR}--\eqref{equation::def-DR}.
Section \ref{section::OLS-estimation-of-interference-function} briefly introduces the OLS estimator, which performs well when $k_0$ is correctly specified. However, in overfitting scenarios where $k > k_0$, OLS-based inference tends to be conservative due to unnecessarily fine node partitioning.
To address such challenge, we will introduce a regularized regression approach in Section \ref{section::estimation-of-interference-function-using-fused-lasso-approach}, which can automatically adapt to the underlying latent homogeneity structure in $\bbeta_k^0$ for $k\geq k_0$.

\subsubsection{Inference with the OLS method}
\label{section::OLS-estimation-of-interference-function}

With heterogeneous treatment effects $\tau_i$'s, we can no longer write the potential outcomes for all $n$ nodes in  the combined linear regression model presented in \eqref{def::potential-outcome-model-1-simplified-matrix-form}. 
We thus estimate the interference function values based on model \eqref{equation::untreated-nodes-model} for untreated nodes. 
Specifically, for each given $k \geq k_0$, we have the OLS estimator
$\hat \bbeta_{k} = (\bX_{k}^T\bX_{k})^{-1} \bX_{k}^T \by_{obs}$.

Based on \eqref{equation::def-OR}--\eqref{equation::def-DR}, we can compute $\hat\tau^{OR}$ by rewriting $(\sum_{i=1}^n Z_i)^{-1}\sum_{i=1}^nZ_i \hat f_{i,k} = \bv^T \hat \bbeta_{k}$ with $\bv\in \mathbb{R}^{d(k)}$ a $d(k)$-dimensional vector whose components satisfy $v_l=c_l/\sum_{i=1}^n Z_i$ for $l \in [d(k)]$. Here, $c_l$ is the total number of treated nodes with feature $g_{k,l}$ according to the matching procedure in Remark \ref{remark::matching-procedure}.
Likewise, to get $\hat\tau^{DR}$, we rewrite $(\sum_{i=1}^n Z_i)^{-1}\sum_{i=1}^n \{Z_i \hat f_{i,k} - (1-Z_i)p_i \hat f_{i,k}/(1-p_i)\} = \bu^T \hat \bbeta_{k}$ for some vector $\bu \in \mathbb{R}^{d(k)}$ with components $u_l=\sum_{i\in E_{k,l}} \{Z_i-(1-Z_i)p_i/(1-p_i)\}/\sum_{i=1}^n Z_i$ for $l \in [d(k)]$, where $E_{k,l} := \{j \in [n]: \gamma_0(G_j^{\bz}(k))=g_{k,l}\}$ is the collection of all nodes with feature $g_{k,l}$, defined analogously to $S_{k,l}$ around Assumption \ref{assumption::Balanced-features}.
Formally, for each $\alpha \in (0,1)$, we can define the asymptotic level $(1-\alpha)$ CI for $\tau$ as
\begin{align} 
\label{equation::OLS-ORandDR-infer-tau}
    {\rm CI}_k := \Big[\hat\tau - \boldsymbol{\Phi}^{-1}(1-\alpha/2) w_k, \, \hat\tau + \boldsymbol{\Phi}^{-1}(1-\alpha/2) w_k\Big].
\end{align}
Consequently, both ${\rm CI}_k^{OR}$ and ${\rm CI}_k^{DR}$ using the OR and DR estimators can be obtained by substituting $\hat\tau$ in \eqref{equation::OLS-ORandDR-infer-tau} with $\hat\tau^{OR}$ and $\hat\tau^{DR}$, respectively, and replacing $w_k$ with
\allowdisplaybreaks
\begin{align}    
  \label{equation::OLS-OR-se}
  w_k^{OR} =&~ \Big[\{ \bv^T (\bX_{k}^T\bX_{k})^{-1} \bv + n_1^{-1} \} \frac{\by_{obs}^T (\bI_{n_0} -\bH_k) \by_{obs}}{n_0 - d(k)} \Big]^{1/2},\\
  \label{equation::OLS-DR-se}
  w_k^{DR} =&~ \Big[\Big\{ \bu^T (\bX_{k}^T\bX_{k})^{-1} \bu + n_1^{-1} + \sum_{i=1}^n\frac{(1-Z_i) p_i^2}{n_1^2 (1-p_i)^2 } 
  \Big\} \frac{\by_{obs}^T (\bI_{n_0} -\bH_k) \by_{obs}}{n_0 - d(k)}  \Big]^{1/2},
\end{align}
where $n_1$ is the number of treated nodes and $\bH_k := \bX_{k}(\bX_{k}^T \bX_{k})^{-1} \bX_{k}^T$ is the projection matrix of $\bX_{k}$.
Recall that $S_{k,l}$ is the set of untreated nodes with feature $g_{k,l}$ for $l\in [d(k)]$.
Theorem \ref{theorem::OLS-infer-tau-asy-cp} below demonstrates the unbiasedness of the estimated interference function values and the validity of the inference procedure.
We refer to Section \ref{sec.supp.ols} of the Supplementary Material for further details on the OLS-based method.

\begin{theorem}
\label{theorem::OLS-infer-tau-asy-cp}
    Under model \eqref{equation::untreated-nodes-model}, for each $k \geq k_0$, we have $\Expected(\hat f_{i,k}) = f_i$. 
    Furthermore, under model \eqref{def::potential-outcome-model}, given the potential outcomes, interference network $G$, treatments $\bZ$, mapping $\gamma_0(\cdot)$, and a conservative upper bound $k \geq k_0$ satisfying  $\kappa d(k)^{3/2} \rightarrow 0$ with $\kappa = \max_{l \in [d(k)]}\{1/|S_{k,l}|\}$, 
    it holds that for each $\alpha \in (0,1)$, $\Prob(\tau \in {\rm CI}_k^{OR}) \rightarrow 1 - \alpha$ and $\Prob(\tau \in {\rm CI}_k^{DR}) \rightarrow 1 - \alpha$ as $n_0 \rightarrow \infty$.
\end{theorem}

In our setting, if the partition of $n_0$ untreated nodes based on neighborhood size $k$ 
is balanced, we have $\kappa \asymp d(k)/n_0$.
While Theorem \ref{theorem::OLS-infer-tau-asy-cp} allows $d(k) \rightarrow \infty$ at certain rate for the suggested CIs to be asymptotically valid, in practice the inference procedure could be conservative in finite samples due to potential overfitting issue as mentioned at the end of Section \ref{sec::matching-estimator}.
Although $\by_{obs}^T (\bI_{n_0} -\bH_k) \by_{obs} / (n_0 - d(k))$ in \eqref{equation::OLS-OR-se}--\eqref{equation::OLS-DR-se} is an unbiased estimator of $\sigma_0^2$ when $k > k_0$, it is less efficient than that with $k=k_0$.
Thus, in Section \ref{section::estimation-of-interference-function-using-fused-lasso-approach} we will introduce an alternative approach to address these challenges.

\subsubsection{Inference with the square-root fused clipped Lasso (SFL) method}
\label{section::estimation-of-interference-function-using-fused-lasso-approach}

To automatically adapt to the latent homogeneity in $\bbeta^0_k$, a natural idea is to employ the fused Lasso approach \citep{shen2010grouping}, which enables learning the underlying homogeneity structure in the interference function values.
In our setting, such consideration leads to the exhaustive pairwise penalty $\sum_{1 \leq i<j \leq d(k)}$ $ \lambda |\beta_{k,i}-\beta_{k,j}|$ for $\bbeta_{k} =  (\beta_{k,i},\ldots,\beta_{k,d(k)})^T $.
Similar to the Lasso penalty, it introduces an intrinsic bias due to the excessive shrinkage.
To overcome such issue, we exploit the grouping pursuit approach in \cite{shen2010grouping}, which was shown to enjoy both asymptotic guarantees and computational efficiency using the difference of convex (DC) programming. 
We also incorporate the idea of the square-root Lasso approach \citep{belloni2011square} to eliminate the need to know or preestimate the error standard deviation $\sigma_0$ for choosing the regularization parameter.
Specifically, we consider estimating $\bbeta_{k}^0$ by minimizing the objective function for the square-root fused clipped Lasso (SFL) given by 
\begin{align}
    \underset{\scriptsize{\bbeta_{k}} \in \mathbb R^{d(k)}}{\rm argmin}
    \Big[ (2n_0)^{-1/2}\big\|\by_{obs}- \bX_{k}\bbeta_{k}\big\|_2 + \lambda_1 \sum_{1 \leq i<j \leq d(k)} \min\big\{\big|\beta_{k,i}-\beta_{k,j}\big|,\lambda_2\big\} \Big],
    \label{equation::fused-lasso}
\end{align}
where $\lambda_1$, $\lambda_2$ are nonnegative regularization parameters. 
The fused clipped Lasso penalty in \eqref{equation::fused-lasso} based on grouping pursuit distinguishes between large and small pairwise differences. 
 
Let $M^0$ be the number of distinct values in $\bbeta_{k}^0$, and $\bfeta^0 = (\eta_1^0, \ldots, \eta_{M^0}^0)^T \in \mathbb R^{M^0}$ the vector recording these distinct values. We then define $\g_i^0 := \{l: \beta_{k,l}^0 =\eta_i^{0}, l\in [d(k)]\}$ as the index set of components in $\bbeta_{k}^0$ whose values are equal to $\eta_i^{0}$ for $i\in [M^0]$. 
For each $k \geq k_0$, it is seen that $\g^0 := \{\g_i^0: i \in [M^0]\}$ forms a partition of index set $\{1,\cdots, d(k)\}$. 
With some reordering of components, we can rewrite $\bbeta_{k}^0$ as $(\eta_1^0 \bone_{|\g_1^0|}, \ldots, \eta_{M^0}^0 \bone_{|\g_{M^0}^0|})^T$, where $\bone_m \in \mathbb R^m$ is a vector of ones.
Hereafter, we slightly abuse the notation and directly write
\begin{align}
    \bbeta_{k}^0= (\eta_1^0 \bone_{|\g_1^0|}, \ldots, \eta_{M^0}^0 \bone_{|\g_{M^0}^0|})^T.
    \label{equ::beta-k-reorder}
\end{align}

To facilitate the derivations below, denote by $\g := \{\g_i: i \in [|\g|]\}$ a generic partition of the index set $\{1, \ldots, d(k)\}$, and define $\bB_k = (\bB_{k,i,j})\in \mathbb{R}^{d(k) \times |\g|}$, where $\bB_{k,i,j}=1$ if $i \in \g_j$ and $\bB_{k,i,j}=0$ otherwise. 
Then based on the oracle grouping $\g^0$ of indices in $\bbeta_{k}^0$, we can derive $\bB_k^0$ and have $\bbeta_{k}^0=\bB_k^0\bfeta^0$.
More importantly, by definition, it is seen that $\bX_{k}\bB_k^0 \in \mathbb R^{n_0 \times M^0}$ is invariant across all $k \geq k_0$, and we thus denote it as $\bD_{}$. Indeed, matrix $\bD$ records the homogeneity in inference function values across all $n_0$ untreated nodes, and right-multiplying $\bB_k^0$ merges groups in $\bX_k$ with the same inference function values in parition $S_k$ definded in Section \ref{sec::matching-estimator}. The regression model \eqref{equation::untreated-nodes-model} can be rewritten as
\begin{align}
    \by_{obs} = 
    \bD_{}\bfeta^0 + \bveps_0. 
    \label{equation::simplified-v-ground-truth}
\end{align}
Observe that the representation in \eqref{equation::simplified-v-ground-truth} is identical across all $k \geq k_0$  and serves as the ground truth. Since $\bfeta^0$ has distinct components, model \eqref{equ::beta-k-reorder} has taken into full account of the underlying homogeneity structure in the true interference function values. In light of representation $\bbeta_{k}^0$ in \eqref{equ::beta-k-reorder}, we see that parameter $M_0$ is the intrinsic dimensionality of $\bbeta_{k}^0$ and plays the same role as the sparsity parameter in high-dimensional sparse learning problems. We also note that $M^0$ is allowed to slowly diverge with sample size $n$.

To quantify the difficulty of grouping pursuit, we further define a signal strength measure $\xi_{\min}:= \min\{ |\eta_{\ell_1}^0 - \eta_{\ell_2}^0|: 1 \leq \ell_1 < \ell_2 \leq M^0\}$, and denote by $s_{\min}$ the smallest group size  with $s_{\min}:=\min_{i \in [M^0]}\{\bone_{n_0}^T\bD_{}\be_i\}$.
Then the oracle-assisted least squares estimate for $\bfeta^0$ is
$$
    \hat\bfeta^{0, ols} : =(\hat\eta_1^{ols}, \ldots, \hat\eta_{M^0}^{ols})^T=(\bD_{}^T \bD_{})^{-1} \bD_{}^T \by_{obs}.
$$ 

Since the components of $\bfeta^0$ are bounded as assumed in Definition \ref{definition::potential-outcome-model}, we can solve the SFL problem in \eqref{equation::fused-lasso} with the implicit constraint that $\|\bX_{k}\bbeta_{k}\|^2_2 = O(n_0)$. Given $k$ and the corresponding partition $\g^0$ as defined above, let 
$   \hat\bbeta^{ols} = (\hat\eta_1^{ols} \bone_{|\g_1^0|}, \ldots, \hat\eta_{M^0}^{ols} \bone_{|\g_{M^0}^0|})^T \in \mathbb R^{d(k)}.
$
It is seen that $\hat\bbeta^{ols}$ is the oracle-assisted OLS estimator of $\bbeta_{k}^0$. 

For each fixed $k \geq k_0$,
solving the regularization problem \eqref{equation::fused-lasso} yields an estimated regression coefficient vector $ \hat\bbeta^{grp}\in \mathbb R^{d(k)}$ as well as a partition  $\hat\g$ of the index set $\{1, \ldots, d(k)\}$ recording the estimated homogeneity structure in $\hat\bbeta^{grp}$. 
Denote by $M=|\hat{\g}|$ and $\hat\g := \{\hat\g_i:i\in [M]\}$. Similar to $\bbeta_{k}^0$, the solution $\hat\bbeta^{grp}$  can also be written as
$$
    \hat\bbeta^{grp} = \hat\bB\hat\bfeta^{grp}=(\hat\eta_1^{grp} \bone_{|\hat\g_1|}, \ldots, \hat\eta_{M}^{grp} \bone_{|\hat\g_M|})^T,
$$
where $\hat\bfeta^{grp} = (\hat\eta_1^{grp}, \ldots, \hat\eta_{M}^{grp})^T$ is the vector recording the distinct values in $\hat\bbeta^{grp}$, and $\hat\bB$ is a matrix defined analogous to $\bB_k$ based on the estimated group partition $\hat\g$. 
Note that since the estimate $\hat \g$ may not be perfect, its cardinality may be different from that of $\g^0$, resulting in different dimensionalities of $\hat\bfeta^{grp}$ and $\bfeta^0$. Nevertheless, the estimate $\hat\bbeta^{grp}$ and the target $\bbeta_{k}^0$ have the same dimensionality for any given $k$. 

After obtaining $\hat\bbeta^{grp}$ for an input $k$, the corresponding predicted interference function values of untreated nodes can be written as $\hat\bD_{\hat\g}\hat\bfeta$, where $\hat\bD_{\hat\g}=\bX_{k}\hat\bB$.  
Motivated by such representation, we next introduce a key technical assumption, referred to as the restricted eigenvalue condition in the literature, that is frequently imposed to facilitate the theoretical analysis of the regularized estimator obtained from \eqref{equation::fused-lasso}.
For any partition $\g$ of the index set $\{1, \ldots, d(k)\}$, let us define $\bD_{\g}=\bX_{k} \bB_k$.
Then for any $\bfeta  \in \mathbb R^{|\g|}$ satisfying $\|\bD_{\g} \bfeta \|^2_2 = O(n_0)$, the restricted eigenvalue $c_{\min}(\g)$ is defined as the smallest eigenvalue of 
\begin{align*}
    \bD_{\g}^T
    \Big\{
        \frac{\bI_{n_0} \|\by_{obs}- \bD_{\g}\bfeta \|^2_2 - (\by_{obs}- \bD_{\g}\bfeta )(\by_{obs}- \bD_{\g}\bfeta )^T}{4n_0^2 \big(\sqrt{(2n_0)^{-1}\|\by_{obs}- \bD_{\g}\bfeta \|^2_2} \big)^3}
    \Big\}
    \bD_{\g}.
\end{align*}
Denote by $\mathcal{C}(\bX_k)$ the column space spanned by $\bX_k$.
One can verify the positivity of $c_{\min}(\g)$ when $\by_{obs} \notin \mathcal{C}(\bX_k)$ for $k \geq k_0$.
We refer to the theorem below and its proof for more discussions on the restricted eigenvalue condition.

\begin{theorem}
\label{theorem::lasso-solution} 
    Under model \eqref{def::potential-outcome-model}, for each given $k \geq k_0$ such that $d(k)\prec n_0$ and any partition $\g$ of the index set $\{1, \ldots, d(k)\}$,
    when $M^0 < M^* \leq \min(\sqrt{n_0}, d(k))$ for some $M^*$ and 
    \begin{align}    
    \label{assumption::restricted-eigen-value-condi}
        \min_{|\g| \leq (M^*)^2, \bfeta_{\scriptsize\g} \in \mathbb R^{|\scriptsize\g|}} c_{\min}(\g) > \lambda_1 \lambda_2^{-1} (2M^*+1),
    \end{align}
    it holds that as $n_0, d(k) \rightarrow \infty$,    
    $   \Prob(\hat\g \neq \g^0) \leq \Prob(\hat\bbeta^{grp} \neq \hat\bbeta^{ols}) \rightarrow 0$, 
    provided that $A)$ $\lambda_2 \in (0, 2\xi_{\min}/3)$ and $s_{\min}(\xi_{\min}-3\lambda_2/2)^2/(4 \sigma_0^2) - 2 \log M^0 \rightarrow \infty$, and $B)$ $ - \log d(k) + (n_0^2 \lambda_1^2)/ (2 \max_{i \in [d(k)]} \|\bx_i\|_2^2) \rightarrow \infty$ with  $\bx_i$ the $i$th column of $\bX_{k}$.
\end{theorem}

Theorem \ref{theorem::lasso-solution} above establishes the grouping consistency of the covariates and the asymptotic equivalence of $\hat\bbeta^{grp}$ and $\hat\bbeta^{ols}$. Note that here we do \textit{not} impose any additional structure assumptions on the interference function and the interference network. Condition \eqref{assumption::restricted-eigen-value-condi} above is similar to that in \cite{shen2010grouping} and \cite{belloni2011square}. 
Such restricted eigenvalue condition is frequently employed to theoretically analyze high-dimensional regularized estimators \citep{Hebiri2011TheSA,dalalyan2012fused}.
In particular, for model \eqref{def::potential-outcome-model}, when the partitioning of $n_0$ untreated nodes based on neighborhood size $k$ yields balanced sets with size of order $n_0/d(k)$, it holds that $c_{\min}(\g) \geq c/d(k)$ with probability tending to one for some $c>0$.
To appreciate the restricted eigenvalue condition and conditions A)--B) in Theorem \ref{theorem::lasso-solution}, let us consider a simple scenario when $M^0$ and $M^*$ are independent of $n_0$ and $d(k)$.
Then the conditions reduce to $\lambda_1 \sqrt{n_0d(k)/\log d(k)} \rightarrow \infty$ and $d(k) c \lambda_1 < \lambda_2 \leq 2\xi_{\min}/3 - \delta_{n_0}$ for some constant $c>0$ and sequence $\delta_{n_0}>0$ with $\sqrt{n_0/d(k)}\delta_{n_0} \rightarrow \infty$, where the signal strength $\xi_{\min}$ needs to satisfy that $\sqrt{n_0/d(k)}\xi_{\min} \rightarrow \infty$. Hence, as $d(k)$ diverges, the constraint on $\lambda_1/\lambda_2$ tightens and a stronger signal strength is required for consistent estimation.

To infer the ADET $\tau$, we define $\hat f_{i,k}^{grp}$ as the estimated interference function value on node $i$ by matching with $\hat\bbeta^{grp}$ based on $\hat\g$ according to Remark \ref{remark::matching-procedure}. 
Then following an analogous procedure as in Section \ref{section::OLS-estimation-of-interference-function}, we let $\tilde{\bv},\tilde{\bu} \in \mathbb{R}^{M}$ be vectors such that $(\sum_{i=1}^n Z_i)^{-1}\sum_{i=1}^nZ_i \hat f_{i,k}^{grp} = \tilde{\bv}^T \hat\bfeta^{grp}$ and $(\sum_{i=1}^n Z_i)^{-1}\sum_{i=1}^n \{Z_i \hat f_{i,k}^{grp} - (1-Z_i)p_i \hat f_{i,k}^{grp}/(1-p_i)\} = \tilde{\bu}^T \hat\bfeta^{grp}$.
In view of Remark \ref{remark::matching-procedure}, the components of $\tilde{\bv}$ satisfy that $\tilde v_l=\tilde c_l/\sum_{i=1}^n Z_i$ for $l \in [M]$, where $\tilde c_l$ is the total number of treated nodes with feature in $\{g_{k,\ell}:\ell \in \hat\g_l\}$.
Similarly, the components of $\tilde{\bu}$ satisfy that $\tilde u_l=\sum_{i\in \tilde E_{k,l}} \{Z_i-(1-Z_i)p_i/(1-p_i)\}/\sum_{i=1}^n Z_i$ for $l \in [M]$, where $\tilde E_{k,l} := \{j \in [n]: \gamma_0(G_j^{\bz}(k)) = g_{k,\ell}, \ell \in \hat\g_l\}$ is the collection of all nodes with feature in $\{g_{k,\ell}:\ell \in \hat\g_l\}$. We can then compute $\hat \tau^{OR}$ and $\hat \tau^{DR}$ through \eqref{equation::def-OR} and \eqref{equation::def-DR}, denoted as $\hat \tau^{OR}_{\rm sfl}$ and $\hat \tau^{DR}_{\rm sfl}$ for clarity.
Combining the property of the OLS estimation and the results in Theorem \ref{theorem::lasso-solution}, we now introduce the inference of the ADET $\tau$ under the SFL framework.

\begin{theorem}
    \label{theorem::SFL-infer-tau-asy-cp}
    Under model \eqref{def::potential-outcome-model}, given the potential outcomes, interference network $G$, treatments $\bZ$, mapping $\gamma_0(\cdot)$, and any upper bound $k \geq k_0$ of neighborhood size, let
    $$
        {\rm CI}_{k,{\rm sfl}} := \Big[\hat\tau_{\rm sfl} - \boldsymbol{\Phi}^{-1}(1-\alpha/2) w_{k,{\rm sfl}},~ \hat\tau_{\rm sfl} + \boldsymbol{\Phi}^{-1}(1-\alpha/2) w_{k,{\rm sfl}} \Big]
    $$
    for each $\alpha \in (0,1)$, and define ${\rm CI}_{k,{\rm sfl}}^{OR}$ and ${\rm CI}_{k,{\rm sfl}}^{DR}$ by substituting $\hat\tau_{\rm sfl}$ above with $\hat\tau^{OR}_{\rm sfl}$ and $\hat\tau^{DR}_{\rm sfl}$, respectively, and replacing $w_{k,{\rm sfl}}$ with
    \begin{align*}
      w_{k,{\rm sfl}}^{OR} = \Big[ \{ \tilde{\bv}^T (\bD_{\hat\g}^T \bD_{\hat\g})^{-1} \tilde{\bv} + n_1^{-1} \} \by_{obs}^T (\bI_{n_0} -\bH_{\hat\g}) \by_{obs} / (n_0 - |\hat\g|) \Big]^{1/2},
    \end{align*}
    and
    \begin{align*}
      w_{k,{\rm sfl}}^{DR} = \Big[ \Big\{ \tilde{\bu}^T (\bD_{\hat\g}^T \bD_{\hat\g})^{-1} \tilde{\bu} + n_1^{-1} + \sum_{i=1}^n\frac{(1-Z_i) p_i^2}{n_1^2 (1-p_i)^2}  \Big\} \by_{obs}^T (\bI_{n_0} -\bH_{\hat\g}) \by_{obs} / (n_0 - |\hat\g|) \Big]^{1/2},
    \end{align*}
    respectively, where $\bD_{\hat\g} \in \mathbb R^{n_0 \times |\hat\g|}$ is the design matrix obtained according to $\hat\g$ and $\bH_{\hat\g}$ is the projection matrix onto the column space of $\bD_{\hat\g}$.
     Then under all the conditions of Theorem \ref{theorem::lasso-solution}, it holds that $\Prob(\tau \in {\rm CI}_{k,{\rm sfl}}^{OR}) \rightarrow 1 - \alpha$ and $\Prob(\tau \in {\rm CI}_{k,{\rm sfl}}^{DR}) \rightarrow 1 - \alpha$ as $n_0 \rightarrow \infty$. 
\end{theorem}

From Theorem \ref{theorem::SFL-infer-tau-asy-cp} above, it is seen that the asymptotic results when $n_0 \rightarrow \infty$ implicitly imply that the size of $G$ with given treatments goes to infinity.
The SFL-based inference procedure in Theorem \ref{theorem::SFL-infer-tau-asy-cp} accommodates potentially high-dimensional settings and addresses the homogeneity among the interference function values. In particular, the SFL-based procedure provides valid inference for $\tau$ with tighter CIs compared to those from the OLS-based procedure in Section \ref{section::OLS-estimation-of-interference-function}. 

\begin{remark}
    In general, our method does not require specifying $\gamma_0(\cdot)$ since the subgraph $G_i^{\bz}(k)$ serves as a sufficient statistic for $\gamma_0(G_i^{\bz}(k))$.
    Although the SFL method addresses the overfitting issue associated with unnecessarily fine partitioning of nodes, choosing subgraphs $G_i^{\bz}(k)$ as the node feature may violate Assumption \ref{assumption::Balanced-features} in practice. Thus, we assume correctly specified $\gamma_0(\cdot)$; see
    \citet{savje2021average} for discussions on the misspecified mappings.
\end{remark}

\section{Inference on the neighborhood size}
\label{section::inference-on-the-neighborhood-size}

We now develop an inference procedure for the true neighborhood size $k_0$. 
To this end, we focus on the untreated nodes with $z_i=0$ and write model \eqref{equation::untreated-nodes-model} with true $k_0$ as
\begin{align}
    \by_{obs} = \bX_{k_0}\bbeta_{k_0}^0 + \sigma_0 \bu, \text{ where } \bu \sim \mathcal{N}(0, \bI_{n_0}).
    \label{equation::control-grp-outcome}
\end{align}
In additional to its own interest, a natural by-product of such study yields a conservative upper bound on $k_0$ which can be incorporated in the inference method suggested in Section \ref{sec::inference-for-the-average-treatment-effect} for inferring $\tau$.
Let us first introduce a regularity condition below for identifying $k_0$.

\begin{assumption}[Identifiability]
\label{assumption::identifiability-condition-k0}
    Given $\gamma_0(\cdot)$, assume that $k_0$ in Assumption \ref{assumption::NI} is the smallest value of $k$ such that $\bX_{k} \bbeta_{k} = \bX_{k_0} \bbeta_{k_0}^0$ for some $\bbeta_{k}\in\mathbb{R}^{d(k)}$.
\end{assumption}

Our proposal is based on the \textit{repro samples} idea proposed in \cite{wang2022finite}. 
Without any prior knowledge of $k_0$, we first construct a data-driven candidate set for $k_0$, which is expected to contain the true neighborhood size with overwhelming probability, to narrow down the search region in the parameter space.
Inspired by the \textit{repro samples} approach, we simulate artificial ${\bf u}_1^*, {\bf u}_2^*, \cdots, {\bf u}_B^* \overset{i.i.d.}{\sim} \mathcal{N}(0, \bI_{n_0})$ as the \textit{repro} copies of the error term $\bu$ in model \eqref{equation::control-grp-outcome} and estimate $k_0$ by solving
\begin{align}\label{eq:obj-func}
    (\hat k_{b,\lambda}, \hat{\bbeta}_{b, \hat k},\hat \sigma_b) = \argminA_{k,\scriptsize{\bbeta}_{k}, \sigma} \left\{ \|\by_{obs}- \bX_{k}\bbeta_{k}-\sigma \bu_b^* \|^2_2  + \lambda k \right\},
\end{align}
where $b = 1, \cdots, B$. We then obtain the candidate set
$
    \mathcal S_B :=\{(\hat k_{b,\lambda}, \hat{\bbeta}_{b, \hat k},\hat \sigma_b): b=1,\cdots, B\}.
$
When $B$ is large enough, some repro copies would fall within a small neighborhood of the unobserved realization of error term in \eqref{equation::control-grp-outcome} associated with $\by_{obs}$, and event $\{\text{some } \hat k_{b,\lambda} =k_0\}$ is very likely to happen.
The algorithm of constructing $\mathcal S_B$ is summarized in Section \ref{sec.supp.infer-k0} of the Supplementary Material.
Alternatively, if an conservative upper bound $K$ on parameter $k_0$ is available, one can also choose the candidate set as $\mathcal S_B=[K]$.

We proceed with constructing the confidence set for $k_0$ with asymptotic coverage at least $1 - \alpha + o(1)$ for each given significance level $\alpha \in (0,1)$. To this end, we go one step further by applying the \textit{conditional repro samples} method \citep{wang2022finite}.
Specifically, for each given tuple $(k, \bbeta_{k},\sigma_0)$, we can create artificial repro samples for the untreated nodes through 
$
    \by^* = \bX_{k} \bbeta_{k} + \sigma_0 \bu^*,
    \label{equation::repro-control-data}
$
where $\bu^* \overset{i.i.d.}{\sim} N(0, \bI_{n_0})$.
Given the set of candidate values, since we are only interested in the inference on $k_0$, we instead adopt the sufficient statistics idea and consider a generating procedure of $\by^*$ that is free of $(\bbeta_{k},\sigma_0)$.
This will greatly reduce the computational cost in the subsequent inference procedure.
Standard calculations yield that 
\begin{align}
    \label{equation::y_repro_decomposition}
    \by^* =&~ \bA_k(\mathbf \by^*) + b_k(\mathbf \by^*)\frac{(\bI_{n_0} -\bH_k)\bu^*}{\|(\bI_{n_0} -\bH_k)\bu^*\|_2},
\end{align}
where $\bA_k(\mathbf \by^*) = \bH_k\mathbf \by^*$ and $b_k(\mathbf \by^*) = \|(\bI_{n_0} -\bH_k)\by^*\|_2$.
With shorthand notation $\bW_k(\by^*)=\{\bA_k(\mathbf \by^*), b_k(\mathbf \by^*)\}$, conditional on $\bW_k(\by^*)$ the distribution of $\by^*$ is independent of parameters $(\bbeta_{k}, \sigma_0)$. We are now ready to define the \textit{nuclear mapping function} \citep{wang2022finite} that is free of $(\bbeta_{k}, \sigma_0)$ and provides a desired confidence set for $k_0$.
For each given $\bY \in \mathbb R^{n_0}$ and given potential candidate set $\mathcal S_B$, one can estimate $k_0$ via
\begin{align}
\label{def::nuclear-mapping-function}
    \hat {k}(\bY) 
    = 
    \underset{k' \in \mathcal S_B}{\rm argmin}\
    \big\{
    \min_{\bbeta_{k'} \in \mathbb R^{d(k')}}
    \{\|\bY- \bX_{k'}\bbeta_{k'}\|^2_2 + \lambda' k'\}
    \big\}.
\end{align}
The output $\hat {k}(\bY)$ above is referred to as a \textit{nuclear statistic} calculated from the \textit{nuclear mapping function}.
By construction, conditional on $\bW_k(\bY)$ the distributions of $\bY$ and $\hat {k}(\bY)$ are independent of the unknown parameters $(\bbeta_{k}, \sigma)$.
Then from Theorem 3 of \cite{wang2022finite}, we see that if there is a Borel set $\mathcal{B}_{\alpha}(k,\bw)$ such that
\begin{align}
    \Prob\{\hat {k}(\bY) \in \mathcal{B}_{\alpha}(k,\bw) | \bW_k(\bY)=\bw\} \geq 1 - \alpha,
    \label{equation::nuclear-mapping-borel-set}
\end{align}
there exists a confidence set $\Gamma_{\alpha}(\bY)$ for $k_0$ such that $\Prob\{k_0 \in \Gamma_{\alpha}(\bY)\} \geq 1 - \alpha$,
with form 
\begin{align}\nonumber
    \Gamma_{\alpha}(\bY)= & \big\{0 \leq k \leq n: \text{ there exist some } \bu^* \sim N(0, \bI_{n_0}) \text{~and~} (\bbeta_{k}, \sigma) \text{ such that } \\ 
    & \bY = \bX_{k} \bbeta_{k} + \sigma \bu^* \text{ and } \hat {k}(\bY) \in \mathcal{B}_{\alpha}\big(k,\bW_k(\bY)\big) \big\}.
    \label{equation::def-confidence-set}
\end{align}

Hence, to construct the confidence set for $k_0$, we need only to find a valid Borel set $\mathcal{B}_{\alpha}(k,\bw)$ such that \eqref{equation::nuclear-mapping-borel-set} above holds.
Based on  \eqref{equation::y_repro_decomposition}, we can generate $\by^*$'s that are free of  $(\bbeta_{k}, \sigma)$, and thus overcome the impact of the nuisance parameters. 
This allows us to construct the Borel set for $k_0$ alone. 
Denote by $
    p(k|\bw) = \Prob\{ \hat {k}(\bY) = k | \bW_k(\bY) = \bw \} 
$ the conditional probability mass function and let 
\begin{align} 
    \label{def::cumulative-prob-function}
    \mathcal{F}(k|\bw)
    =
    \sum_{k':p(k'|\bw) \leq p(k|\bw)} p(k'|\bw).
\end{align}
Such construction helps us derive $\mathcal{B}_{\alpha}(k,\bw)$ as outlined in the proposition below.

\begin{proposition}
    \label{proposition::construct-borel-set}
    Define the Borel set as
        $\mathcal{B}_{\alpha}(k,\bw)
        =
        \big\{ 0 \leq k' \leq n_0: \mathcal{F}(k'|\bW_k(\bY)=\bw) \geq \alpha \big\}$. 
    Then \eqref{equation::nuclear-mapping-borel-set} holds. 
    Moreover, we have 
        $\Prob \big\{\hat {k}(\bY) \in \mathcal{B}_{\alpha}\big(k,\bW_k(\bY)\big) \big\}  \geq 1 - \alpha$.
\end{proposition}

With a candidate set $\mathcal S_B$, we can derive the confidence set for $k_0$ as
\begin{align}
    \bar\Gamma_{\alpha}(\by_{obs})
    :=
    \Gamma_{\alpha}(\by_{obs}) \cap \mathcal S_B
    =
    \big\{k \in \mathcal S_B: \hat {k}(\by_{obs}) \in \mathcal{B}_{\alpha}\big(k,\bW_k(\by_{obs})\big) \big\}.
    \label{equation::simp-form-considence-set}
\end{align}
To demonstrate the validity of the confidence set, define the separation measure \citep{shen2013constrained,wang2022finite} between the true model and various candidate models with under-specified $k$ as
$$
    C_{\min} = \min_{0 \leq k< k_0} \bigg\{\frac{\| \bX_{k_0} \bbeta_{k_0}^0 - \bX_{k} \bbeta_{k}\|_2^2}{n_0(k_{0}-k)}\bigg\}.
$$
From Assumption \ref{assumption::identifiability-condition-k0}, it holds that $C_{\min}>0$.
The theoretical guarantees on the coverage probability of $\bar\Gamma_{\alpha}(\by_{obs})$ are given in the two theorems below, examining both scenarios when $n_0$ is finite or the number of Monte Carlo copies $B$ is limited.

\begin{theorem}
\label{thm::confidence-cover-1}
    Assume that $n_0 - d(k_0) > 4$. Then for each $\delta > 0$, there exists a constant $\gamma_{\delta}>0$ such that when $\lambda \in \Big[\gamma_{\delta} \sigma^2_0 / (\sqrt{1+\frac{2}{3}\gamma^{\frac{1}{4}}_{\delta}} - 1), n_0 \gamma^{1/4}_{\delta} \frac{C_{\min}}{6}\Big]$, for any finite $n_0$, significance level $\alpha \in (0,1)$, and any $\delta>0$, we have 
    $
        \Prob(k_0 \in \bar\Gamma_{\alpha}(\by_{obs})) \geq 1 - \alpha - \delta - o(e^{-c_1B})
    $
    for some $c_1>0$, where $B \rightarrow \infty$ and $\bar\Gamma_{\alpha}(\by_{obs})$ is the confidence set given by \eqref{equation::simp-form-considence-set}.
\end{theorem}

\begin{theorem}
\label{thm::confidence-cover-2}
     Assume that there exist some constants $m_1, m_2>0$ such that
     $
        \frac{\lambda}{n_0} \in \Big[ (\sigma^2_0 + \frac{2\sigma^2_0}{n_0}) + m_1, \, \min \big \{ 0.015 C_{\min} - \frac{3\sigma^2_0 \ln2 }{n_0}, 0.015 C_{\min} - \frac{\sigma^2_0 d(k_0) }{n_0}\big\} - m_2\Big]
    $ 
    and $\frac{\log (2k_0)}{n_0 - d(k_0) - 1} < \log(\frac{5}{3})$.
    Then for any finite $B$ and each 
    $\alpha \in (0,1)$, we have
    $
        \Prob(k_0 \in \bar\Gamma_{\alpha}(\by_{obs})) \geq 1 - \alpha - o(e^{-c_2n_0})
    $
    for some $c_2>0$, where $n_0 \rightarrow \infty$ and $\bar\Gamma_{\alpha}(\by_{obs})$ is the confidence set given by \eqref{equation::simp-form-considence-set}.
\end{theorem}

Note that $\mathcal{B}_{\alpha}\big(k,\bW_k(\by_{obs})\big)$ in \eqref{equation::simp-form-considence-set} depends on $\mathcal{F}(k|\bw)$, which is unknown in practice because $p(k|\bw)$ is unknown. In view of \eqref{equation::y_repro_decomposition},  we can consistently estimate $p(k|\bw)$ through the Monte Carlo method by generating repro samples $\by^*$  of $\by_{obs}$ using \eqref{equation::y_repro_decomposition} by setting $\{\bA_k(\mathbf \by^*), b_k(\mathbf \by^*)\} = \{\bA_k(\mathbf \by_{obs}), b_k(\mathbf \by_{obs})\}$ and sampling $\bu^* \sim N(0, \bI_{n_0})$.
The detailed algorithm for constructing the confidence set $\bar\Gamma_{\alpha}(\by_{obs})$ is summarized in Section \ref{sec.supp.infer-k0} of the Supplementary Material.

From Theorems \ref{thm::confidence-cover-1} and \ref{thm::confidence-cover-2} above, we see that $\bar\Gamma_{\alpha}(\by_{obs})$ captures $k_0$ with probability at least $1-\alpha-o(1)$ whenever the number of treated nodes $n_0$ or the the number of Monte Carlo copies  goes to infinity,
under certain conditions on the penalty level $\lambda$.
They provide practical guidance on the choice of $\lambda$. 
For instance, considering fixed $n_0$, we can first define a grid of points in range $[\nu_1, n_0 \nu_0]$ for some small $\nu_0,\nu_1>0$ 
and then search for the best $\lambda$ among the grids using some information criterion \citep{FanTang2013,tang2016fused} when constructing $\mathcal S_B$. The tuning parameter $\lambda'$ used in calculating $\hat{k}(\bY)$ can be selected using similar grid search method. 

We now discuss how the confidence set constructed above can assist us in making an informed choice of the input neighborhood size $k$ for the prior-assisted inference procedure in Section \ref{section::the-general-prior-assisted-inference-framework}. 
Denote by $k^*$ and $k_\alpha^*$ the maximum values in $\mathcal S_B$ and $\bar\Gamma_{\alpha}(\by_{obs})$, respectively. 
Then as long as either $n_0 \rightarrow \infty$ or $B \rightarrow \infty$, for each $\alpha \in (0,1)$, it holds that 
$
    \Prob(k_0 \leq k_\alpha^*) 
    \geq \Prob(k_0 \leq k_\alpha^*, k_0 \in \bar\Gamma_{\alpha}(\by_{obs})) 
    = \Prob(k_0 \in \bar\Gamma_{\alpha}(\by_{obs})) 
    \geq 1 - \alpha - o(1).
$
Replacing $\mathcal S_B$ with $[K]$ in \eqref{equation::simp-form-considence-set} to compute $\bar\Gamma_{\alpha}(\by_{obs})$ for a given upper bound $K$ of $k_0$ also yields
$
    \Prob(k_0 \leq k_\alpha^*) 
    \geq  
    1 - \alpha - o(1)
$, justifying its validity in terms of the asymptotic coverage.

\section{Simulation studies}
\label{section::numerical-experiments}

In this section, we empirically evaluate the performance of the HNCI inference framework for both the ADET and the true neighborhood size. We fix network size $n=1000$ and generate heterogeneous direct treatment effect for each node where $\{\tau_i\} \sim \mathrm{Uniform}[0.6,0.8]$.
For the interference network edges $\{G_{i,j}\}$, we independently draw $\{\xi_i\} \sim \mathrm{Uniform}[0,1]$ and set $G_{i,j} = G_{j,i} \sim \mathrm{Bernoulli}(g(\xi_i,\xi_j))$ for $i<j$.
We consider two graphon functions $g_1(\xi_i,\xi_j) = 0.02$ (Erd{\H{o}}s--R{\'e}nyi model) and $g_2(\xi_i,\xi_j) = (\ell-0.3) \mathbbm{1}\{\xi_i,\xi_j \in ((\ell-1)/6,\ell/6)\} / 40 + 0.3/40$ (stochastic block model with a blockwise constant structure), with $\mathbbm{1}{\{\cdot\}}$ denoting the indicator function.
We then generate the treatment assignments based on the propensity score of each node where $\{p_i\} \sim \mathrm{Uniform}[0.03,0.06]$.
Two mappings are adopted for the matching procedure in neighborhood interference: $\gamma_1(G_i^{\bz}(k)) = \{\lfloor M_{i,\ell}/4 \rfloor\}_{\ell \leq k}$, where $M_{i,\ell}$ is the number of treated depth-$\ell$ neighbors of node $i$ in $G_i^{\sbz}(k)$; and $\gamma_2(G_i^{\bz}(k)) = \{\lfloor \rho_{i,\ell}/0.05 \rfloor\}_{\ell \leq k}$, where $\rho_{i,\ell}$ is the proportion of treated nodes among depth-$\ell$ neighbors of node $i$ in $G_i^{\sbz}(k)$.
When inferring the ADET in Section \ref{sec::infer-adet}, we fix the true neighborhood size $k_0=2$, denote by $\{T_{i,1}, T_{i,2}\}$ the features of node $i$ based on $k_0$-hop, and set the interference function $f(T_{i,1}, T_{i,2}) = 5 T_{i,1} / \max\{T_{i,1}\} + 2.5 T_{i,2} / \max\{T_{i,2}\}$.
Then we generate the potential outcomes using model \eqref{def::potential-outcome-model} with $\{\epsilon_i\} \sim \mathrm{N}(0,\,\text{std}=0.5)$.
By construction, we have four data generating mechanisms of \textit{setting 1}: $g_1(\xi_i,\xi_j)$ and $\gamma_1(G_i^{\bz}(k))$; \textit{setting 2}: $g_1(\xi_i,\xi_j)$ and $\gamma_2(G_i^{\bz}(k))$; \textit{setting 3}: $g_2(\xi_i,\xi_j)$ and $\gamma_1(G_i^{\bz}(k))$; and \textit{setting 4}: $g_2(\xi_i,\xi_j)$ and $\gamma_2(G_i^{\bz}(k))$.

\subsection{Performance of inferring the ADET}
\label{sec::infer-adet}

For each of the four data generating mechanisms introduced above, we first generate the interference network, the treatment assignments, and the direct treatment effects.
We then generate $1000$ replications of potential outcomes $\{Y_i\}$ and infer the ADET using four methods: the OR estimator with OLS and SFL methods, respectively; the DR estimator with OLS and SFL methods, respectively. In specific, when applying the SFL method, we choose $\lambda_1=\lambda_2= c_0 / \sqrt{n}$ with $c_0=1/30$.
We construct $1000$ CIs using each method and perform the entire procedure $100$ times (i.e., repetitions).

\begin{figure}[h]
    \centering
    \includegraphics[width=\textwidth]{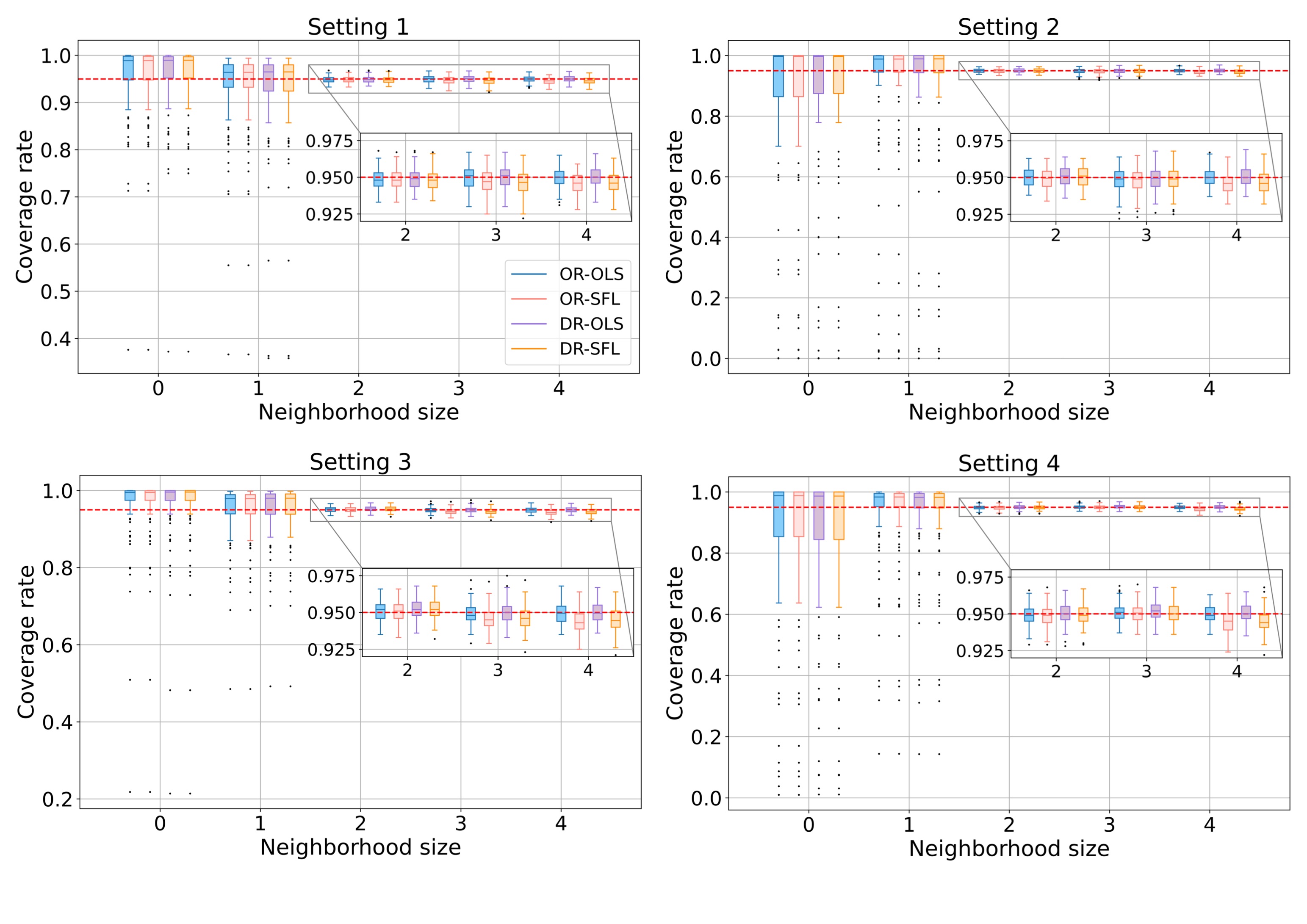}
    \vspace{-4.5em}
    \caption{Empirical coverage probabilities of different methods across $100$ repetitions.
    }
    \label{fig::er.sbm_g1.2_cp}
\end{figure}

As discussed in Section \ref{section::the-general-prior-assisted-inference-framework}, inferring the ADET requires only a conservative upper bound on $k_0$. 
Here, we vary $k \in \{0,1,2,3,4\}$, including underfitting scenarios with $k<k_0=2$.
When $k > 2$, the nodes are partitioned into unnecessarily fine groups.
The empirical coverage probabilities of different methods are shown in Figure \ref{fig::er.sbm_g1.2_cp}.

Across the four settings, all methods maintain empirical coverage probabilities close to the nominal level when $k \geq k_0$.
However, when $k < k_0$, Figure \ref{fig::er.sbm_g1.2_cp} demonstrates considerable variation in the coverage rates across the $100$ repetitions, ranging from $0$ to $1$ depending on the realization of the interference network and treatment assignments.
This aligns with findings in Section \ref{section::discussion-on-k0s-role-in-inferring-tau} under the simplified model \eqref{def::potential-outcome-model-1-simplified}.
Ignoring interference (i.e., $k=0$) or considering only direct neighbors (i.e., $k=1$) affects inference validity, highlighting the need to estimate $k_0$ or set a conservative upper bound when inferring the ADET.

\begin{table}[h] 
    \caption {Average confidence interval widths for different methods.} 
    \vspace{0.5em}
    \centering
    \resizebox{\textwidth}{!}{
    \begin{tabular}{ll|ccccc||ccccc}
    \toprule
    \multirow{2}{*}{} & \multirow{2}{*}{Method} & \multicolumn{5}{c||}{Mapping $\gamma_1$}  & \multicolumn{5}{c}{Mapping $\gamma_2$} \\
    \cmidrule{3-12}
      & & $k=0$ & $k=1$ & $k=2$ & $k=3$ & $k=4$ & $k=0$ & $k=1$ & $k=2$ & $k=3$ & $k=4$ \\
    \midrule
    \multirow{4}{*}{Graphon 1} & OR - OLS & 0.5107 & 0.3808 & 0.2988 & 0.3035 & 0.3038 & 0.8378 & 0.6613 & 0.3090 & 0.3095 & 0.3128 \\
                               & OR - SFL & 0.5107 & 0.3808 & 0.2987 & 0.2993 & 0.2993 & 0.8378 & 0.6612 & 0.3086 & 0.3086 & 0.3073 \\
                               & DR - OLS & 0.5110 & 0.3813 & 0.2992 & 0.3039 & 0.3041 & 0.8410 & 0.6635 & 0.3103 & 0.3108 & 0.3141 \\
                               & DR - SFL & 0.5110 & 0.3813 & 0.2990 & 0.2996 & 0.2996 & 0.8410 & 0.6634 & 0.3100 & 0.3100 & 0.3087 \\
    \midrule
    \multirow{4}{*}{Graphon 2} & OR - OLS & 0.5987 & 0.4369 & 0.3120 & 0.3164 & 0.3188 & 0.7274 & 0.5680 & 0.3080 & 0.3086 & 0.3138 \\
                               & OR - SFL & 0.5987 & 0.4369 & 0.3118 & 0.3117 & 0.3111 & 0.7274 & 0.5679 & 0.3077 & 0.3077 & 0.3059 \\
                               & DR - OLS & 0.6017 & 0.4394 & 0.3138 & 0.3182 & 0.3206 & 0.7304 & 0.5702 & 0.3093 & 0.3099 & 0.3151 \\
                               & DR - SFL & 0.6017 & 0.4394 & 0.3136 & 0.3135 & 0.3129 & 0.7304 & 0.5701 & 0.3091 & 0.3090 & 0.3072 \\
    \bottomrule
    \end{tabular}
    }
\label{tab::average-mean-CI-length-across-100-iterations}
\end{table}

Regarding the widths of CIs under different input $k$, we take the average over $1000$ replications within each single simulation repetition.
As a result, under each setting, we obtain the average CI widths for different methods and choices of $k$'s, respectively. The results are presented in Table \ref{tab::average-mean-CI-length-across-100-iterations}.
When applying the same estimator, the SFL method in general produces shorter CIs than those of the OLS method, especially for larger values of $k$. 
When $k=0$, there is no difference between the OLS and SFL methods since all nodes belong to the same group.
For the OLS method, the average CI width decreases as $k$ increases from $0$ to $2$, but then increases as $k$ continues to grow, due to conservative variance estimation.
This does not necessarily hold for the SFL method as it groups the coefficients to address the overfitting issue.
Table \ref{tab::average-mean-CI-length-across-100-iterations} again underscores the importance of selecting a $k$ that is no smaller than $k_0$. 
Combining this with Figure \ref{fig::er.sbm_g1.2_cp}, when $k < k_0$ the inference of the ADET is unreliable while the CIs are wide.

Additional simulation results are presented in Section \ref{section::appendix-simulations} of the Supplementary Material, where we examine scenarios with misspecified propensity scores and settings \textit{without exact matching}; that is, the interference function values of nodes within the same group are \textit{approximately} centered around a common value. 
The results illustrate the robust empirical performance of our HNCI inference procedure under both misspecified settings.

\subsection{Performance of inferring the neighborhood size}

We now evaluate the inference methods for $k_0$ suggested in Section \ref{section::inference-on-the-neighborhood-size}. 
Following the convention, we denote $\mathcal{S}_B$ as the candidate set calculated applying \eqref{eq:obj-func}, and $\bar\Gamma_{\alpha}(\by_{obs})$ as the confidence set from \eqref{equation::simp-form-considence-set} with $\alpha=0.05$.
We apply the same settings as introduced above with network size $n = 1000$, considering different interference functions for different $k_0$ values. 
When $k_0 = 0$, indicating no interference among the nodes, the interference function in \eqref{def::potential-outcome-model} is zero. 
Recall that features of node $i$ based on $1$-hop and $2$-hop are represented by $\{T_{i,1}\}$ and $\{T_{i,1}, T_{i,2}\}$, respectively. 
For $k_0 = 1$, we use the interference function $f(T_{i,1}) = 1.5 T_{i,1} / \max\{T_{i,1}\}$. For $k_0 = 2$, we set $f(T_{i,1}, T_{i,2}) = 10 T_{i,1} / \max\{T_{i,1}\} + 1.2 T_{i,2} / \max\{T_{i,2}\}$.
We apply two approaches to infer $k_0$ and compare the coverage probability and the average cardinality of the output confidence set: 1) {\tt Conf1}: $\bar\Gamma_{\alpha}(\by_{obs})$ with $\mathcal S_B=[K]$, where $K$ is some upper bound on potential $k_0$; 2) {\tt Conf2}: $\bar\Gamma_{\alpha}(\by_{obs})$ with $\mathcal S_B$ the candidate set calculated from \eqref{eq:obj-func}. 

We set the number of Monte Carlo copies $B=200$ when computing $\mathcal S_B$, and $J=100$ when computing $\bar\Gamma_{\alpha}(\by_{obs})$.
We use the Bayesian information criterion (BIC) to choose the tuning parameters $\lambda$ and $\lambda'$.
To investigate the performance of different methods, we repeat the simulation $500$ times under each setting. 
The inference results are provided in Table \ref{tab::k0-table}.

\begin{table}[h] 
    \caption {Inference results on $k_0$ by different methods across $500$ repetitions shown as pairs representing (coverage probability, average cardinality of confidence set).} 
    \vspace{0.5em}
    \centering
    \resizebox{\textwidth}{!}{
    \begin{tabular}{ll|ccc||ccc}
    \toprule
    \multirow{2}{*}{} & \multirow{2}{*}{Method} & \multicolumn{3}{c||}{Mapping $\gamma_1$}  & \multicolumn{3}{c}{Mapping $\gamma_2$} \\
    \cmidrule{3-8}
      & & $k_0=0$ & $k_0=1$ & $k_0=2$    
      & $k_0=0$ & $k_0=1$ & $k_0=2$  \\
    \midrule
    \multirow{2}{*}{Graphon 1} & {\tt Conf1} & (1.00, 5.00) & (1.00, 4.02) & (1.00, 4.03) & (1.00, 5.00) & (1.00, 4.83) & (1.00, 3.64) \\
                               & {\tt Conf2} & (1.00, 1.00) & (0.99, 1.00) & (1.00, 1.00) & (1.00, 1.00) & (1.00, 1.00) & (0.99, 1.00)  \\
    \midrule
    \multirow{2}{*}{Graphon 2} & {\tt Conf1} & (1.00, 5.00) & (1.00, 4.03) & (1.00, 4.05) & (1.00, 5.13) & (1.00, 5.07) & (1.00, 3.91) \\
                               & {\tt Conf2} & (1.00, 1.00) & (0.98, 1.00) & (1.00, 1.00) & (1.00, 1.00) & (1.00, 1.00) & (1.00, 1.00)  \\
    \bottomrule
    \end{tabular}
    }
\label{tab::k0-table}
\end{table}

As shown in Table \ref{tab::k0-table}, the coverage probabilities of {\tt Conf1} and {\tt Conf2} are close to $1$ under various settings.
Due to the discrete nature of $k_0$, although the methods are conservative in terms of coverage probability, {\tt Conf2} yields a confidence set with an average size of $1$, which precisely captures the true value over 98\% of the time under the four settings.
As for {\tt Conf1}, the average cardinality of the output confidence set generally decreases as $k_0$ increases, particularly when using the mapping $\gamma_2(\cdot)$.
Specifically, under setting $2$ with graphon $1$ and mapping $\gamma_2(\cdot)$, when $k_0=2$ we observe that some confidence sets exclude $0$ and $1$ across $500$ repetitions, indicating that exploiting larger values of neighborhood size is necessary for the downstream analysis. This can serve as a practical guidance for selecting the neighborhood size.
As noted in \cite{wang2022finite}, the discrete nature of the confidence set can often preclude an exact confidence level of $1-\alpha$, leading to its ``conservativeness."
Using the candidate set $\mathcal S_B$ can help obtain a smaller confidence set when necessary.

\section{Real data application}
\label{section::teenage-friends-and-lifestyle-study-analysis}

We further apply the suggested HNCI inference framework to the teenage friends and lifestyle study conducted at a secondary school in Glasgow between 1995 and 1997 \citep{michell1997girls}. 
Specifically, we are interested in investigating and making inference on the \textit{network} causal effect of romantic relationships on the substance use during the last wave of the study.
Such study collected the friendship network among 160 teenagers.
The adjacency matrix is symmetrized to form the interference network, where $G_{i,j} = G_{j,i} = 1$ if either student $i$ or $j$ considers the other a friend. 
We consider the binary variable ``romantic" as the treatment assignments $\bz$, which indicates whether the student is in a romantic relation ($0$ for no/untreated and $1$ for yes/treated).
The outcomes of interest are alcohol consumption ($1$-$5$ scale), tobacco use ($1$-$3$ scale), and cannabis use ($1$-$4$ scale), with higher values indicating more frequent consumption. 
Visualizations and additional details of the data can be found in Section \ref{sec.supp.realdata} of the Supplementary Material.

We now specify the HNCI implementation for this network causal inference application.
With the interference network $G$ and treatments $\bz$, let us consider the mapping based on the number of treated neighbors for the matching procedure in neighborhood interference, where $\gamma_0(G_i^{\bz}(k)) = \{\lfloor M_{i,\ell}/2 \rfloor\}_{\ell \leq k}$ and $M_{i,\ell}$ is as defined in Section \ref{section::numerical-experiments}.
Observe that our inference procedure is flexible and does \textit{not} assume a specific form for the interference function. 
When inferring the ADET using the DR estimator, we set the propensity score for each node as the proportion of treated nodes in $G$.
Since the maximum node depth in the interference network is $2$, it is natural to set $\mathcal S_B=\{0,1,2\}$ when computing the confidence set $\bar\Gamma_{\alpha}(\by_{obs})$ for $k_0$. 
Across all three substance use domains, the resulting 95\% confidence set is $\{1,2\}$, motivating us to set $k=2$ as a conservative upper bound on $k_0$ for inferring the ADET of romantic relationships, as suggested at the end of Section \ref{section::inference-on-the-neighborhood-size}.
The results are summarized in Table \ref{tab::glasgow-w3-infer-tau}.
The inference results based on the OR estimator are consistent with those from the DR estimator, which could be attributed to the good approximation of our potential outcome model to the real data.
As expected, the SFL-based methods provide shorter CIs, highlighting the advantage of SFL in addressing the potential overfitting issues.

\begin{table}[h] 
    \caption {Inference results: 95\% CIs for the ADET by different methods. } 
    \vspace{0.5em}
    \centering
    \resizebox{0.9\linewidth}{!}{
    \begin{tabular}{lcccc}\toprule
    Method & OR-OLS & OR-SFL & DR-OLS & DR-SFL \\ \midrule 
    Alcohol & $(-0.3745, 0.5619)$ & $(-0.3336, 0.5059)$ & $(-0.3745, 0.5619)$ & $(-0.3336, 0.5059)$  \\ 
    Tobacco & $(-0.0958, 0.5084)$ & \textbf{(0.0082, 0.4872)} & $(-0.0958, 0.5084)$ & \textbf{(0.0082, 0.4872)}  \\ 
    Cannabis & $(-0.0497, 0.8168)$ & $(-0.0432, 0.7340)$ & $(-0.0497, 0.8168)$ & $(-0.0432, 0.7340)$  \\ \bottomrule
    \end{tabular}
    }
\label{tab::glasgow-w3-infer-tau}
\end{table}

Most existing studies have focused on the association between the romantic relationship and substance use. 
Previous research suggests that being in a romantic relationship or engaging in romantic activities is associated with the initiation and continuation of substance use among adolescents \citep{furman2009romantic}.
Our study addresses the challenge of inferring the network causal effects of romantic relationships on substance use under the network interference.
In practice, the SFL-based methods are preferred, particularly in scenarios with potential overfitting.
As illustrated in Table \ref{tab::glasgow-w3-infer-tau}, while the methods suggest no ADET of romantic relationships on the frequency of alcohol consumption or cannabis use, the SFL-based methods indicate that the ADET of romantic relationship on tobacco use is significantly greater than zero ($\alpha=0.05$), which is missed by the OLS-based methods.

\section{Discussions}
\label{section::discussion}

We have investigated the problem of high-dimensional causal inference under network interference and suggested a new method of HNCI by exploiting the ideas of neighborhood adaptive estimation and repro samples. It enables flexible, tuning-free inference of the ADET based on the square-root fused clipped Lasso (SFL) and a conservative upper bound on the neighborhood size. We have theoretically justified CIs on the ADET and confidence set for the neighborhood size.
Our current work assumes an exact matching of the values for the interference function. It would be interesting to consider the approximate matching when the interference function values are close but not necessarily identical. To allow for more flexible network feature engineering and pattern generation, it would be beneficial to incorporate the idea of graph neural network (GNN). 
These problems are beyond the scope of the current paper and will be interesting topics for future research.

\bibliographystyle{apalike}
\bibliography{references}

\begin{thebibliography}{}

\bibitem[Awan et~al., 2020]{awan2020almost}
Awan, U., Morucci, M., Orlandi, V., Roy, S., Rudin, C., and Volfovsky, A.
  (2020).
\newblock Almost-matching-exactly for treatment effect estimation under network
  interference.
\newblock In {\em International Conference on Artificial Intelligence and
  Statistics}, pages 3252--3262.

\bibitem[Baird et~al., 2018]{baird2018optimal}
Baird, S., Bohren, J.~A., McIntosh, C., and {\"O}zler, B. (2018).
\newblock Optimal design of experiments in the presence of interference.
\newblock {\em Review of Economics and Statistics}, 100(5):844--860.

\bibitem[Belloni et~al., 2011]{belloni2011square}
Belloni, A., Chernozhukov, V., and Wang, L. (2011).
\newblock Square-root lasso: pivotal recovery of sparse signals via conic
  programming.
\newblock {\em Biometrika}, 98(4):791--806.

\bibitem[Belloni et~al., 2022]{BelloniFangVolfovsky2022}
Belloni, A., Fang, F., and Volfovsky, A. (2022).
\newblock Neighborhood adaptive estimators for causal inference under network
  interference.
\newblock {\em arXiv preprint arXiv:2212.03683}.

\bibitem[Boyd et~al., 2011]{boyd2011distributed}
Boyd, S., Parikh, N., Chu, E., Peleato, B., Eckstein, J., et~al. (2011).
\newblock Distributed optimization and statistical learning via the alternating
  direction method of multipliers.
\newblock {\em Foundations and Trends{\textregistered} in Machine learning},
  3(1):1--122.

\bibitem[Crump et~al., 2009]{crump2009dealing}
Crump, R.~K., Hotz, V.~J., Imbens, G.~W., and Mitnik, O.~A. (2009).
\newblock Dealing with limited overlap in estimation of average treatment
  effects.
\newblock {\em Biometrika}, 96(1):187--199.

\bibitem[Dalalyan and Chen, 2012]{dalalyan2012fused}
Dalalyan, A. and Chen, Y. (2012).
\newblock Fused sparsity and robust estimation for linear models with unknown
  variance.
\newblock {\em Advances in Neural Information Processing Systems}, 25.

\bibitem[D’Amour et~al., 2021]{d2021overlap}
D’Amour, A., Ding, P., Feller, A., Lei, L., and Sekhon, J. (2021).
\newblock Overlap in observational studies with high-dimensional covariates.
\newblock {\em Journal of Econometrics}, 221(2):644--654.

\bibitem[Eckles et~al., 2017]{eckles2017design}
Eckles, D., Karrer, B., and Ugander, J. (2017).
\newblock Design and analysis of experiments in networks: reducing bias from
  interference.
\newblock {\em Journal of Causal Inference}, 5(1):20150021.

\bibitem[Erd{\H{o}}s and R{\'e}nyi, 1959]{erdds1959random}
Erd{\H{o}}s, P. and R{\'e}nyi, A. (1959).
\newblock On random graphs.
\newblock {\em Publicationes Mathematicae Debrecen}, 6:290--297.

\bibitem[Fan and Tang, 2013]{FanTang2013}
Fan, Y. and Tang, C.~Y. (2013).
\newblock Tuning parameter selection in high dimensional penalized likelihood.
\newblock {\em Journal of the Royal Statistical Society Series B},
  75(3):531--552.

\bibitem[Forastiere et~al., 2021]{forastiere2021identification}
Forastiere, L., Airoldi, E.~M., and Mealli, F. (2021).
\newblock Identification and estimation of treatment and interference effects
  in observational studies on networks.
\newblock {\em Journal of the American Statistical Association},
  116(534):901--918.

\bibitem[Furman et~al., 2009]{furman2009romantic}
Furman, W., Low, S., and Ho, M.~J. (2009).
\newblock Romantic experience and psychosocial adjustment in middle
  adolescence.
\newblock {\em Journal of Clinical Child \& Adolescent Psychology},
  38(1):75--90.

\bibitem[Gao and Ding, 2023]{gao2023causal}
Gao, M. and Ding, P. (2023).
\newblock Causal inference in network experiments: regression-based analysis
  and design-based properties.
\newblock {\em arXiv preprint arXiv:2309.07476}.

\bibitem[Hebiri and van~de Geer, 2011]{Hebiri2011TheSA}
Hebiri, M. and van~de Geer, S.~A. (2011).
\newblock The smooth-lasso and other $\ell$1$+\ell$2-penalized methods.
\newblock {\em Electronic Journal of Statistics}, 5:1184--1226.

\bibitem[Jagadeesan et~al., 2020]{Jagadeesan2020Designs}
Jagadeesan, R., Pillai, N.~S., and Volfovsky, A. (2020).
\newblock {Designs for estimating the treatment effect in networks with
  interference}.
\newblock {\em The Annals of Statistics}, 48(2):679--712.

\bibitem[Leung, 2022]{leung2022causal}
Leung, M.~P. (2022).
\newblock Causal inference under approximate neighborhood interference.
\newblock {\em Econometrica}, 90(1):267--293.

\bibitem[Li and Wager, 2022]{li2022random}
Li, S. and Wager, S. (2022).
\newblock Random graph asymptotics for treatment effect estimation under
  network interference.
\newblock {\em The Annals of Statistics}, 50(4):2334--2358.

\bibitem[Liu et~al., 2016]{liu2016inverse}
Liu, L., Hudgens, M.~G., and Becker-Dreps, S. (2016).
\newblock On inverse probability-weighted estimators in the presence of
  interference.
\newblock {\em Biometrika}, 103(4):829--842.

\bibitem[Michell and Amos, 1997]{michell1997girls}
Michell, L. and Amos, A. (1997).
\newblock Girls, pecking order and smoking.
\newblock {\em Social Science \& Medicine}, 44(12):1861--1869.

\bibitem[Puelz et~al., 2022]{puelz2022graph}
Puelz, D., Basse, G., Feller, A., and Toulis, P. (2022).
\newblock A graph-theoretic approach to randomization tests of causal effects
  under general interference.
\newblock {\em Journal of the Royal Statistical Society Series B},
  84(1):174--204.

\bibitem[Robins et~al., 1994]{robins1994estimation}
Robins, J.~M., Rotnitzky, A., and Zhao, L.~P. (1994).
\newblock Estimation of regression coefficients when some regressors are not
  always observed.
\newblock {\em Journal of the American statistical Association},
  89(427):846--866.

\bibitem[Rubin, 1980]{rubin1980randomization}
Rubin, D.~B. (1980).
\newblock Randomization analysis of experimental data: the {F}isher
  randomization test comment.
\newblock {\em Journal of the American statistical association},
  75(371):591--593.

\bibitem[S{\"a}vje et~al., 2021]{savje2021average}
S{\"a}vje, F., Aronow, P., and Hudgens, M. (2021).
\newblock Average treatment effects in the presence of unknown interference.
\newblock {\em The Annals of statistics}, 49(2):673--701.

\bibitem[Shen and Huang, 2010]{shen2010grouping}
Shen, X. and Huang, H.-C. (2010).
\newblock Grouping pursuit through a regularization solution surface.
\newblock {\em Journal of the American Statistical Association},
  105(490):727--739.

\bibitem[Shen et~al., 2013]{shen2013constrained}
Shen, X., Pan, W., Zhu, Y., and Zhou, H. (2013).
\newblock On constrained and regularized high-dimensional regression.
\newblock {\em Annals of the Institute of Statistical Mathematics},
  65(5):807--832.

\bibitem[Sussman and Airoldi, 2017]{sussman2017elements}
Sussman, D.~L. and Airoldi, E.~M. (2017).
\newblock Elements of estimation theory for causal effects in the presence of
  network interference.
\newblock {\em arXiv preprint arXiv:1702.03578}.

\bibitem[Tan, 2006]{tan2006distributional}
Tan, Z. (2006).
\newblock A distributional approach for causal inference using propensity
  scores.
\newblock {\em Journal of the American Statistical Association},
  101(476):1619--1637.

\bibitem[Tang and Song, 2016]{tang2016fused}
Tang, L. and Song, P.~X. (2016).
\newblock Fused lasso approach in regression coefficients clustering--learning
  parameter heterogeneity in data integration.
\newblock {\em Journal of Machine Learning Research}, 17(113):1--23.

\bibitem[Thi and Dinh, 1997]{LeThi1997SolvingAC}
Thi, H. A.~L. and Dinh, T.~P. (1997).
\newblock Solving a class of linearly constrained indefinite quadratic problems
  by d.c. algorithms.
\newblock {\em Journal of Global Optimization}, 11:253--285.

\bibitem[Wang et~al., 2022]{wang2022finite}
Wang, P., Xie, M.-G., and Zhang, L. (2022).
\newblock Finite- and large-sample inference for model and coefficients in
  high-dimensional linear regression with repro samples.
\newblock {\em arXiv preprint arXiv:2209.09299}.

\bibitem[Yang et~al., 2012]{yang2012feature}
Yang, S., Yuan, L., Lai, Y.-C., Shen, X., Wonka, P., and Ye, J. (2012).
\newblock Feature grouping and selection over an undirected graph.
\newblock In {\em Proceedings of the 18th ACM SIGKDD International Conference
  on Knowledge Discovery and Data Mining}, pages 922--930.

\bibitem[Yohai and Maronna, 1979]{yohai1979asymptotic}
Yohai, V.~J. and Maronna, R.~A. (1979).
\newblock Asymptotic behavior of ${M}$-estimators for the linear model.
\newblock {\em The Annals of Statistics}, 7:258--268.

\end{thebibliography}


\newpage
\appendix
\setcounter{page}{1}
\setcounter{section}{0}
\renewcommand{\theequation}{A.\arabic{equation}}
\setcounter{equation}{0}

\begin{center}{\bf \Large  Supplementary Material to ``HNCI: High-Dimensional Network Causal Inference''}

\bigskip

Wenqin Du, Rundong Ding, Yingying Fan and Jinchi Lv
\end{center}

\noindent This Supplementary Material contains additional details on inferring the ADET $\tau$ with the OLS method, the summirized algorithms for inferring the true neighborhood size $k_0$, the proofs of Theorems \ref{theorem::OLS-infer-tau-asy-cp}--\ref{thm::confidence-cover-2}, Propositions \ref{proposition::infer-tau-model1-overfit-underfit}--\ref{proposition::construct-borel-set}, and some technical lemmas, as well as some additional technical details and results of numerical studies. Unless stated otherwise, all the notation follows the same definitions as in the main body of the paper. The code for reproducing the simulations and real data analysis is available at \url{https://github.com/WenqinDu/HNCI}.

\section{Additional details on inferring $\tau$ with OLS method}
\label{sec.supp.ols}

In this section, we provide additional details and discussions on inferring $\tau$ with the OLS method introduced in Section \ref{section::OLS-estimation-of-interference-function}.
Under the potential outcome model \eqref{def::potential-outcome-model}, the OR and DR estimators in \eqref{equation::def-OR}--\eqref{equation::def-DR} can be expressed as
\begin{align} \label{equation::OR-diff}
    \hat \tau^{OR} - \tau 
    =&~ 
    \frac{1}{\sum_{i=1}^n Z_i}\sum_{i=1}^nZ_i(f_i - \hat f_{i,k}) + \frac{1}{\sum_{i=1}^n Z_i}\sum_{i=1}^nZ_i\epsilon_{i,1},\\ \nonumber
    \hat \tau^{DR} - \tau 
    =&~ 
    \frac{1}{\sum_{i=1}^n Z_i}\sum_{i=1}^n\Big\{Z_i(f_i - \hat f_{i,k}) - \frac{(1-Z_i)(f_i-\hat f_{i,k}) p_i }{1-p_i}\Big\} + \\ \label{equation::DR-diff}
    &~ \frac{1}{\sum_{i=1}^n Z_i}\sum_{i=1}^n \Big\{Z_i\epsilon_{i,1} - \frac{(1-Z_i) p_i \epsilon_{i,0}}{1-p_i}\Big\},
\end{align}
where $f_i = f\big(\gamma_0(G_i^{\bz}(k_0))\big)$, and error terms $\epsilon_{i,0}$ and $\epsilon_{i,1}$ are independent normal random variables with mean zero and variance $\sigma_0^2$, associated with untreated nodes and treated nodes, respectively.
In view of \eqref{equation::OR-diff}--\eqref{equation::DR-diff}, the inference of the ADET $\tau$ relies on valid inference of a linear combination of interference function values and consistent estimation of $\sigma_0^2$.
Under our model setting and applying the OLS-based inference method introduced in Section \ref{section::OLS-estimation-of-interference-function}, one can derive that $\hat \tau^{OR} \equiv \hat \tau^{DR}$ when all $p_i$'s are identical.

To conduct inference on ADET $\tau$ using $\hat \tau^{OR}$, observe that $\{f_i\}$ in the first term on the right-hand side of \eqref{equation::OR-diff} are estimated from the untreated nodes, whereas the second term therein involves only error terms of the treated nodes. 
The inference of $\tau$ follows directly from the independence between these two components. 
Recall that
\begin{align*}    
  w_k^{OR} =&~ \big[\{ \bv^T (\bX_{k}^T\bX_{k})^{-1} \bv + n_1^{-1} \} \by_{obs}^T (\bI_{n_0} -\bH_k) \by_{obs} / (n_0 - d(k)) \big]^{1/2}.
\end{align*}
The first term $\bv^T (\bX_{k}^T\bX_{k})^{-1} \bv$ above arises from the uncertainty of inferring the interference function values using untreated nodes, while the second term $n_1^{-1}$ quantifies the uncertainty due to the model error terms on treated nodes as in \eqref{equation::OR-diff}.
In contrast, when employing the DR estimator, the second term on the right-hand side of \eqref{equation::DR-diff} involves $\bepsilon_0$ of the untreated nodes. Hence, to provide valid inference result using $\hat \tau^{DR}$, we need to carefully assess the dependence between different components. Technical details are provided in the proof of Theorem \ref{theorem::OLS-infer-tau-asy-cp} in Section \ref{sec.supp.A5}.

\section{Algorithms for inferring neighborhood size $k_0$}
\label{sec.supp.infer-k0}

In this section, we summarize the generating process for the the candidate set $\mathcal S_B$ and the confidence set $\bar\Gamma_{\alpha}(\by_{obs})$ as introduced in Section \ref{section::inference-on-the-neighborhood-size} in Algorithms \ref{algorithm::candidate-set}--\ref{algorithm::confidence-set}. As mentioned in \cite{wang2022finite}, utilizing a candidate set as the input of the downstream analysis is particularly effective when the target parameter for inference is discrete.
The generating process for the candidate set can serve as an intermediate step in our inference procedure.
We present the theoretical guarantees on the validity of Algorithm \ref{algorithm::candidate-set} in accurately recovering $k_0$ in Lemmas \ref{thm::candidate-cover-1}--\ref{thm::candidate-cover-2}; see Sections \ref{sec.supp.A12} and \ref{sec.supp.A13}.
Moreover, theoretical guarantees on the validity of Algorithm \ref{algorithm::confidence-set} in terms of the asymptotic coverage are given by Theorems \ref{thm::confidence-cover-1}--\ref{thm::confidence-cover-2} in Section \ref{section::inference-on-the-neighborhood-size}.

\begin{algorithm}[H]
\caption{Candidate set construction for $k_0$}
\label{algorithm::candidate-set}
    \KwIn{The potential outcomes of untreated nodes $\by_{obs} \in \mathbb R^{n_0}$, interference network $G$, treatments $\bZ$, mapping $\gamma_0(\cdot)$, and the number of Monte Carlo copies $B$.}
    
    Simulate a large number $B$ copies of ${\bf u}^* \sim N(0, \bI_{n_0})$. Denote the $B$ copies as $\bu^*_b$, $b = 1, \cdots, B$\;

    Compute $\hat{k}_{b,\lambda}$ as the solution to \eqref{eq:obj-func} for $b = 1, \ldots, B$ and a grid of values for $\lambda$. For each $b$, use certain selection criterion to choose a subset $\Lambda_b$ of all values of $\lambda$\;
    
    \KwOut{The candidate set $\mathcal S_B = \left\{\hat{k}_{b, \lambda}: \lambda \in \Lambda_b, b=1, \dots,  B \right\}$.}
\end{algorithm}

\vspace{1.2em}

\begin{algorithm}[H]
\caption{Confidence set construction for $k_0$}
\label{algorithm::confidence-set}
    \KwIn{The potential outcomes of untreated nodes ${\by}_{obs} \in \mathbb R^{n_0}$, interference network $G$, treatments $\bZ$, mapping $\gamma_0(\cdot)$, candidate set $\mathcal S_B$, significance level $\alpha \in (0,1)$, and the number of Monte Carlo copies $J$.}
    \For{$k \in \mathcal S_B$}{
        \For{$j \in {1, \dots, J}$}{
            Calculate $\bw_{obs} = (\bA_{k,obs}, b_{k,obs})= (\bH_k\by_{obs}, \|(\bI_{n_0} -\bH_k) \by_{obs}\|_2)$\;
            
            Generate $\bu_j^* \sim N(0,\bI_{n_0})$ and compute $\by_j^* = \bA_{k,obs} + b_{k,obs}\frac{(\bI_{n_0} - \bH_k) \bu_j^*}{\|(\bI_{n_0}-\bH_k) \bu_j^*\|_2}$\;

            Compute 
            $\hat {k}(\by^*_j) = \underset{k' \in \mathcal S_B}{\rm argmin}\
            \big\{
            \min_{\bbeta_{k'} \in \mathbb R^{d(k')}}
            (\|\by^*_j- \bX_{k'}\bbeta_{k'}\|^2_2 + \lambda' k')
            \big\}$
            for $\lambda'$ chosen from a grid of values under certain selection criterion\;
        }
        Estimate the conditional probability $\hat p(k'|\bw_{obs}) = \frac{1}{J}\sum_{j=1}^{J} \mathbbm{1}(\hat {k}(\by^*_j) = k')$ and compute $\hat{\mathcal{F}}(k'|\bw_{obs})$ via \eqref{def::cumulative-prob-function} using $\hat p(k'|\bw_{obs})$ for $ k' \in \mathcal S_B$\;

        Calculate 
        $
            \hat k(\by_{obs}) = \underset{k' \in \mathcal S_B}{\rm argmin}\
            \big\{
            \min_{\bbeta_{k'} \in \mathbb R^{d(k')}}
            (\|\by_{obs}- \bX_{k'}\bbeta_{k'}\|^2_2 + \lambda' k')\big\} ;
        $
        
    }

    \KwOut{The confidence set $\bar\Gamma_{\alpha}(\by_{obs})$ with significance level $\alpha$ \\
    \hspace{4cm}
    $
        \bar\Gamma_{\alpha}(\by_{obs})
        =
        \big\{k \in \mathcal S_B: \hat{\mathcal{F}}(\hat {k}(\by_{obs})|\bw_{obs}) \geq \alpha \big\}.
    $
    }
\end{algorithm}

\section{Proofs of Theorems \ref{theorem::OLS-infer-tau-asy-cp}--\ref{thm::confidence-cover-2} and Propositions \ref{proposition::infer-tau-model1-overfit-underfit}--\ref{proposition::construct-borel-set}} \label{sec.supp.A}

\subsection{Proof of Theorem \ref{theorem::OLS-infer-tau-asy-cp}} \label{sec.supp.A5}

We first prove that applying model \eqref{equation::untreated-nodes-model}, for each $k \geq k_0$, we have $\Expected(\hat f_{i,k}) = f_i$. 
Recall the definition of $\bbeta_{k}^0$ in \eqref{def::interaction-function-of-ture-k0} for any $k \geq k_0$.
Under model \eqref{equation::untreated-nodes-model} where $\by_{obs} = \bX_{k}\bbeta_{k}^0 + \bveps_0$, the oracle values of the interference function for the untreated nodes can be written as $\bX_{k}\bbeta_{k}^0$. By construction, we have 
$$
    \Expected(\hat \bbeta_{k}) = (\bX_{k}^T\bX_{k})^{-1} \bX_{k}^T \bX_{k} \bbeta_{k}^0 = \bbeta_{k}^0.
$$ 
Then the estimated values of the interference function satisfy $\Expected(\bX_{k}\hat \bbeta_{k}) = \bX_{k}\bbeta_{k}^0$, which entails $\Expected(\hat f_{i,k}) = f_i$ for $i \in \{j \in [n]:z_j=0\}$.
Therefore, $\{\hat f_{i,k}\}$ are unbiased for the untreated nodes. 
Combining Remark \ref{remark::matching-procedure} and Assumption~\ref{assumption::Balanced-features} on balanced features, the unbiasedness holds for the treated nodes as well.

The asymptotically normal approximation for the ordinary least squares (OLS) estimator is a direct consequence of the results from \cite{yohai1979asymptotic}.
For the DR estimator, the asymptotic independence of the two terms on the right-hand side of \eqref{equation::DR-diff} is characterized in the lemma below, which completes the proof of Theorem \ref{theorem::OLS-infer-tau-asy-cp}. 

\begin{lemma}
\label{lemma::DR-asym-independent-f-and-error}
    Under model \eqref{equation::untreated-nodes-model}, for any $k \geq 0$, given the interference network $G$, treatments $\bZ$, and mapping $\gamma_0(\cdot)$, components
    $\big(\sum_{i=1}^n Z_i\big)^{-1}\sum_{i=1}^n\big\{Z_i(f_i - \hat f_{i,k}) - ((1-Z_i)(f_i-\hat f_{i,k}) p_i ) / (1-p_i)\big\}$
    and
    $\big(\sum_{i=1}^n Z_i\big)^{-1}\sum_{i=1}^n \big\{Z_i\epsilon_{i,1} - ((1-Z_i) p_i \epsilon_{i,0}) / (1-p_i)\big\}$
    in \eqref{equation::DR-diff} are asymptotically independent.
\end{lemma}

\subsection{Proof of Theorem \ref{theorem::lasso-solution}} \label{sec.supp.A6}

With the shorthand notation $\by = \by_{obs}$, $\bX = \bX_{k}$, $\bB=\bB_{k}$, $d = d(k)$, and $\bbeta^0 = \bbeta_{k}^0$, we can define the objective function of the square-root fused clipped Lasso (SFL) as 
\begin{align}
\label{equation::def-obj-s}
    S(\bbeta) = \sqrt{(2n_0)^{-1}\|\by- \bX\bbeta\|^2_2} + \lambda_1 J(\bbeta),
\end{align}
where
\begin{align*}
    J(\bbeta)
    =
    \sum_{1 \leq j<j' \leq d} \min\Big\{\big|\beta_{j}-\beta_{j'}\big|,\lambda_2\Big\}
\end{align*}
denotes the fused clipped Lasso penalty.
It can be naturally decomposed into $S(\bbeta) = S_1(\bbeta) - S_2(\bbeta)$ with 
\begin{align*}
    &~ S_1(\bbeta) = \sqrt{(2n_0)^{-1}\|\by- \bX\bbeta\|^2_2} + \lambda_1 \sum_{1\leq j<j' \leq d}|\beta_j - \beta_{j'}|, \\
    &~ S_2(\bbeta) = \lambda_1 \sum_{1\leq j<j' \leq d} h_2(\beta_j - \beta_{j'}), \ h_2(x) = (|x|+\lambda_2)_+.
\end{align*}
Here, $(|x|+\lambda_2)_+$ represents the positive part of $|x|+\lambda_2$. As noted in \cite{shen2010grouping}, $S_2(\bbeta)$ corrects the estimation bias due to the use of convex penalty $\lambda_1 \sum_{1\leq j<j' \leq d}|\beta_j - \beta_{j'}|$ for the nonconvex problem \eqref{equation::def-obj-s}.
Then we can see that the solution $\hat\bbeta^{grp}$ based on Algorithm 2 in \cite{shen2010grouping} satisfies that for some partition $(\g_1, \ldots, \g_M)$ of $\{1,\cdots, d\}$  with $M \leq \min(\sqrt{n_0}, d)$, 
\begin{align}
\label{equation::kkt-1}
    &~ -\frac{(\sum_{j \in \g_m} \bx_j^T)(\by- \bX\bbeta)}{2 \sqrt{(2n_0)^{-1}\|\by- \bX\bbeta\|^2_2}} + n_0 \lambda_1 \sum_{j \in \g_m} \triangle_j(\bbeta) = 0, \\
\label{equation::kkt-2}
    &~ \Bigg| \frac{\bx_j^T(\by- \bX\bbeta)}{2 \sqrt{(2n_0)^{-1}\|\by- \bX\bbeta\|^2_2}} - n_0 \lambda_1 \triangle_j(\bbeta) \Bigg| \leq n_0 \lambda_1 (|\g_m|-1) \text{~for~} j \in \g_m \text{~with~} |\g_m| \geq 2,
\end{align}
where $\triangle_j(\bbeta) = \sum_{j': j' \neq j} \{ \text{sign}(\beta_j-\beta_{j'}) - \triangledown_{\beta_j} h_2(\beta_j-\beta_{j'}) \}$ for each $j \in [d]$.

We now define an event $\mathcal{F}$ below and aim to show that $\hat\bbeta^{ols}$ is the unique solution satisfying \eqref{equation::kkt-1} and \eqref{equation::kkt-2} on $\mathcal{F}$, which indicates that it is the unique minimizer of $S(\bbeta)$ on event $\mathcal{F}$. Then we will show that such event $\mathcal F$ holds with high probability.

Specifically, let us define 
\begin{align*}
    \mathcal{F} :=&~ \big\{ \min_{1\leq k<\ell \leq M^0} |\hat\eta_k^{ols}-\hat\eta_\ell^{ols}| > 3 \lambda_2/2 \big\} 
    \bigcap 
    \big\{ \sqrt{(2n_0)^{-1}\|\by- \bX\hat\bbeta^{ols}\|^2_2} \geq \sigma_0/2 \big\} \\
    &~ \bigcap_{m:|\g_m^0| \geq 2} \big\{ \max_{j:j \in \g_m^0} \Big| \bx_j^T(\by - \bX\hat\bbeta^{ols})/\sigma_0 \Big| \leq n_0 \lambda_1 (|\g_m^0|-1) \big\}, 
\end{align*}
where we recall that $\g^0$ is the partition corresponding to $\hat{\bbeta}^{ols}$.

\textit{Step 1}. 
When $\min_{1\leq k<\ell \leq M^0} |\hat\eta_k^{ols}-\hat\eta_\ell^{ols}| > 3 \lambda_2/2$, it holds that
\begin{align}
    \triangle_j(\hat\bbeta^{ols}) = 0
    \label{equ::ols-hlper-0}
\end{align}
for $j\in [d]$, which implies 
\begin{align}
    \sum_{j \in \g_m^0} \triangle_j(\bbeta)=0
    \label{equ::ols-hlper-1}
\end{align}
for each $m \in [M^0]$.
By the first-order equation of least squares, we have
\begin{align}
    (\sum_{j \in \g_m^0} \bx_j^T)(\by- \bX\hat\bbeta^{ols}) = 0
    \label{equ::ols-hlper-2}
\end{align}
for each $m \in [M^0]$.
Using \eqref{equ::ols-hlper-1} and \eqref{equ::ols-hlper-2}, we see that $\hat\bbeta^{ols}$ directly satisfies \eqref{equation::kkt-1} with partition $\g = \g^0$ on event $\mathcal{F}$. 
Combining \eqref{equ::ols-hlper-0} with the inequalities on event $\mathcal F$ that $\max_{j:j \in \g_m^0} \Big| \bx_j^T(\by - \bX\hat\bbeta^{ols})/\sigma_0 \Big| \leq n_0 \lambda_1 (|\g_m^0|-1)$ for any $|\g_m^0| \geq 2$ and $\sqrt{(2n_0)^{-1}\|\by- \bX\hat\bbeta^{ols}\|^2_2} \\\geq \sigma_0/2$, it follows immediately that $\hat\bbeta^{ols}$ satisfies \eqref{equation::kkt-2} on event $\mathcal{F}$ with partition $\g = \g^0$.

Therefore, we see that $\hat\bbeta^{ols}$ is a local minimizer of $S(\bbeta)$ on event $\mathcal{F}$.
We next show that $S(\bbeta)$ is close to a strictly convex function, and $\hat\bbeta^{ols}$ is the unique minimizer of that convex function on event $\mathcal{F}$ under our assumptions.

\textit{Step 2}. We define 
$$h(x)=|x|\mathbbm{1}(|x|<\lambda_2) + \lambda_2 \mathbbm{1}(|x| \geq \lambda_2).$$
Following the same procedure as in \cite{shen2010grouping}, we can modify the penalty term in \eqref{equation::def-obj-s} so that it is smooth.
Specifically, denote by 
\begin{align}
    \tilde{h}(x) = 
    \begin{cases} 
          h(x), & \text{if } |x| \leq \lambda_2/2 \text{ or } |x| \geq 3\lambda_2/2; \\
          -\frac{1}{2 \lambda_2} (x - \lambda_2)^2 + \frac{1}{2} (x - \lambda_2) + \frac{7}{8} \lambda_2, & \text{if } |x - \lambda_2| < \lambda_2/2; \\
          -\frac{1}{2 \lambda_2 } (x + \lambda_2)^2 - \frac{1}{2} (x + \lambda_2) + \frac{7}{8} \lambda_2, & \text{if } |x + \lambda_2| < \lambda_2/2.
          \label{equ::def-new-h}
    \end{cases}
\end{align}
Accordingly, we will consider 
\begin{align}
    \tilde S(\bbeta) = \sqrt{(2n_0)^{-1}\|\by- \bX\bbeta\|^2_2} + \lambda_1 \sum_{1 \leq j<j' \leq d} \tilde{h}(\beta_j - \beta_{j'}).
    \label{equ::def-new-s}
\end{align}

Note that the difference between \eqref{equation::def-obj-s} and \eqref{equ::def-new-s} arises solely from substituting $h(x)$ with $\tilde{h}(x)$. In light of the construction in \eqref{equ::def-new-h}, the gradients of $h(x)$ and $\tilde{h}(x)$ are identical whenever $x<\lambda_2/2$ or $x>3\lambda_2/2$. 
On event $\mathcal{F}$, it is seen that $|\hat\beta_j^{ols} - \hat\beta_{j'}^{ols}|$ is either zero or greater than $3\lambda_2/2$.
This implies that $\tilde S(\bbeta)$ and $S(\bbeta)$ share the same subgradient on event $\mathcal{F}$.
Since we have shown that $\hat\bbeta^{ols}$ is a local minimizer of $S(\bbeta)$ on event $\mathcal{F}$, it is thus also a local minimizer of $\tilde S(\bbeta)$ on event $\mathcal{F}$.

We proceed to show that $\hat\bbeta^{ols}$ is the unique minimizer of $\tilde S(\bbeta)$ on event $\mathcal{F}$. Observe that for any partition $\g = \{\g_1,\cdots, \g_M\}$ of $\{1,\cdots, d\}$ with $M \leq \min(\sqrt{n_0}, d)$, $\tilde S(\bbeta)$ is a function of $\bfeta $ 
with $\bbeta = (\beta_1, \ldots, \beta_d)^T = (\eta_1 \bone_{|\g_1|}, \ldots, \eta_{M} \bone_{|\g_{M}|})^T$.
As derived in \cite{shen2010grouping}, the Hessian matrix of the second component in $\tilde S(\bbeta)$ is given by 
\begin{align*}
    \lambda_1 \lambda_2^{-1} \big\{ \bone_{|\g|}\bone_{|\g|}^T - (|\g|+1) \bI_{|\g|} \big\}.
\end{align*}
Under the square-root fused clipped Lasso setting, the Hessian matrix of the first component in $\tilde S(\bbeta)$ can be written as 
\begin{align*}
    \bH_{sq} =&~ \bD_{\g}^T
    \Big\{
        \frac{\bI_{n_0} \|\by- \bX\bbeta\|^2_2 - (\by- \bX\bbeta)(\by- \bX\bbeta)^T}{4n_0^2 \big(\sqrt{(2n_0)^{-1}\|\by- \bX\bbeta\|^2_2} \big)^3}
    \Big\}
    \bD_{\g} \\
    =&~ \bD_{\g}^T
    \Big\{
        \frac{\bI_{n_0} \|\by- \bD_{\g}\bfeta \|^2_2 - (\by- \bD_{\g}\bfeta )(\by- \bD_{\g}\bfeta )^T}{4n_0^2 \big(\sqrt{(2n_0)^{-1}\|\by- \bD_{\g}\bfeta \|^2_2} \big)^3}
    \Big\}
    \bD_{\g}.
\end{align*}

We next show the positive definiteness of matrix $\bH_{sq}$, where we will use the fact that $\by \notin \mathcal{C}(\bX)$ with $\mathcal{C}(\bX)$ standing for the column space of matrix $\bX$. Recall that we can decompose $\bD_{\g}$ as $\bX \bB$, where $\bB \in \mathbb{R}^{d \times |\g|}$ is a full column rank matrix. 
Same as in $\bX$, the columns of $\bB$ are orthogonal to each other and consist of entries that are either $0$ or $1$.

To prove that matrix $\bH_{sq}$ is positive definite, we only need to show that for any $\ba \in \mathbb{R}^d$ and any $\bbeta \in \mathbb{R}^d$, 
\begin{align*}
    h_1(\ba, \bbeta) :=
    \ba^T \bX^T \{ \bI_{n_0} \|\by- \bX\bbeta\|^2_2 - (\by- \bX\bbeta)(\by- \bX\bbeta)^T \} \bX \ba > 0.
\end{align*}
We proceed in what follows. Since $\by \notin \mathcal{C}(\bX)$, there does not exist $c_0 \neq 0$ such that $\bX \ba = c_0 (\by- \bX\bbeta)$.
Consequently, by the Cauchy--Shwarz inequality  we have that 
$$h_1(\ba, \bbeta) = \|\by- \bX\bbeta\|^2_2  \|\bX \ba\|^2_2 - \{(\by- \bX\bbeta)^T (\bX \ba)\}^2 > 0.$$ 
In view of the definition of $c_{\min}(\g)$ in Section \ref{section::estimation-of-interference-function-using-fused-lasso-approach} and noting that $\bX\bbeta$ can be written as $\bD_{\g}\bfeta$,
the above result leads to $c_{\min}(\g)>0$.
We now carefully characterize the order of $c_{\min}(\g)$ when $\|\bX\bbeta\|^2_2 = O(n_0)$. A useful observation is that 
\begin{align}\nonumber
    \frac{h_1(\ba, \bbeta)}{\|\by- \bX\bbeta\|_2^2 \|\bX \ba\|_2^2} & =
    \frac{\ba^T \bX^T}{\|\bX \ba\|_2} \Big\{ \bI_{n_0} - \frac{(\by- \bX\bbeta)(\by- \bX\bbeta)^T}{\|\by- \bX\bbeta\|^2_2} \Big\} \frac{\bX \ba}{\|\bX \ba\|_2}\\ \label{equation::h1_normalized}
    &= 1 - \cos^2(\theta),
\end{align}
where $\theta$ is the angle between vectors $\by- \bX\bbeta$ and $\bX \ba$.

By definition, $\cos^2(\theta)$ is maximized at $\ba^* = (\bX^T \bX)^{-1}\bX^T (\by- \bX\bbeta)$.
Combining this with the fact that $\by = \bX\bbeta^0 + \bepsilon$ for $\bepsilon \sim \mathcal{N}(\bzero, \sigma_0^2 \bI_{n_0})$,  we can deduce that
\begin{align} \nonumber
    \langle \by- \bX\bbeta, \bX \ba^* \rangle
    =&~
    (\by- \bX\bbeta)^T \bX (\bX^T \bX)^{-1}\bX^T \bX (\bbeta^0 - \bbeta) + (\by- \bX\bbeta)^T \bX (\bX^T \bX)^{-1}\bX^T \bepsilon \\ \nonumber
    =&~ (\by- \bX\bbeta)^T \bX (\bbeta^0 - \bbeta) + (\by- \bX\bbeta)^T P_X \bepsilon \\ \nonumber
    =&~ \{\bX (\bbeta^0-\bbeta) + \bepsilon \}^T\{\bX (\bbeta^0 - \bbeta) + P_X \bepsilon\} \\ \label{equation::bound-inner-prod}
    =&~ \|\bX (\bbeta^0 - \bbeta) + P_X \bepsilon\|_2^2,
\end{align}
where $P_X$ denotes the projection matrix onto the column space of $\bX$. Correspondingly, we define $P_{X^{\perp}}$ as the projection matrix onto the complement space of $\mathcal{C}(\bX)$.

In view of \eqref{equation::bound-inner-prod}, we can show that the angle $\theta$ between vectors $\by- \bX\bbeta$ and $\bX \ba$ for any $\ba \in \mathbb{R}^d$ satisfies
\allowdisplaybreaks
\begin{align} \nonumber
    \cos^2(\theta)~
    \leq&~
    \Big\{\frac{\langle \by- \bX\bbeta, \bX \ba^* \rangle}{\|\by- \bX\bbeta\|_2 \|\bX \ba^*} \|_2\Big\}^2 \\ \nonumber
   =&~ \Big\{ \frac{\|\bX (\bbeta^0 - \bbeta) + P_X \bepsilon\|_2^2}{\|\bX (\bbeta^0 - \bbeta) + \bepsilon\|_2 \|\bX (\bX^T \bX)^{-1}\bX^T (\by- \bX\bbeta)\|_2}\Big\}^2 \\ \nonumber
    =&~ \Big\{ \frac{\|\bX (\bbeta^0 - \bbeta) + P_X \bepsilon\|_2}{\|\bX (\bbeta^0 - \bbeta) + \bepsilon\|_2}\Big\}^2
    \\ \label{equation::bound-cos}
    =&~ \frac{\|\bX (\bbeta^0 - \bbeta) + P_X \bepsilon\|_2^2}{\|\bX (\bbeta^0 - \bbeta) + P_X \bepsilon\|_2^2 + \|P_{X^{\perp}} \bepsilon\|_2^2}.
\end{align}

Recall that the columns of $\bX$ are orthogonal to each other and consist of entries that are either $0$ or $1$.
By assumption, we have $d\prec n_0$. 
It follows directly that $\|P_{X^{\perp}} \bepsilon\|_2^2/(n_0-d) \overset{p}{\rightarrow} \sigma_0^2$ as $n_0 \rightarrow \infty$.
As a result, for $n_0 \rightarrow \infty$, there exist some $l_1>l_2>0$ such that
\begin{align}
    \label{equ::posi-def-helper-1}
    \Prob \big\{ n_0l_1 \geq \|P_{X^{\perp}} \bepsilon\|_2^2 \geq n_0l_2 \big\} \rightarrow 1.
\end{align}
Moreover, by the triangle inequality, it holds that $\|\bX (\bbeta^0 - \bbeta) + P_X \bepsilon\|_2 \leq \|\bX (\bbeta^0 - \bbeta)\|_2 + \|P_X \bepsilon\|_2$. Hence, for all $\bbeta$ satisfying $\|\bX\bbeta\|^2_2 = O(n_0)$, we can derive that 
\begin{align}
    \label{equ::posi-def-helper-2}
    \Prob \big\{ \|\bX (\bbeta^0 - \bbeta) + P_X \bepsilon\|_2^2 \leq n_0l_3 \big\} \rightarrow 1
\end{align}
for $n_0 \rightarrow \infty$ and some $l_3>0$ since $\bbeta^0$ is bounded as introduced in Definition \ref{definition::potential-outcome-model}.
Combining \eqref{equation::h1_normalized} with \eqref{equation::bound-cos}--\eqref{equ::posi-def-helper-2} yields that for any vector $\ba \in \mathbb{R}^d$, as $n_0 \rightarrow \infty$  we have
\begin{align}
\label{equation::bound-h1_normalized}
    \Prob \Big\{\frac{h_1(\ba, \bbeta)}{\|\by- \bX\bbeta\|_2^2 \|\bX \ba\|_2^2} \geq c_1\Big\} \rightarrow 1
\end{align}
for some $c_1 \in (0,1)$.

For all $\bbeta$ satisfying $\|\bX\bbeta\|^2_2 = O(n_0)$, let us define 
\begin{align*}
    c_2 := \max_{\bbeta: \, \|\bX\bbeta\|^2_2 = O(n_0)} \sqrt{(2n_0)^{-1}\|\by- \bX\bbeta\|^2_2}.
\end{align*}
Recall that
\begin{align*}
    \bb^T \bH_{sq} \bb 
    =&~ 
    \frac{1}{2n_0 \sqrt{(2n_0)^{-1}\|\by- \bX\bbeta\|^2_2}}
    \bb^T \bB^T \bX^T \Big\{ \bI_{n_0} - \frac{(\by- \bX\bbeta)(\by- \bX\bbeta)^T}{\|\by- \bX\bbeta\|^2_2} \Big\} \bX \bB \bb.
\end{align*}
Then it follows from \eqref{equation::bound-h1_normalized} that for any $\bb \in \mathbb{R}^{|\g|}$ with $\|\bb\|_2 = 1$, we have 
\begin{align*}
    \Prob \Big\{\bb^T \bH_{sq} \bb 
    \geq \frac{c_1 \|\bX \bB \bb\|_2^2}{2n_0 \sqrt{(2n_0)^{-1}\|\by- \bX\bbeta\|^2_2}}\Big\} \rightarrow 1,
\end{align*}
and thus 
\begin{align}
    \Prob \Big\{\bb^T \bH_{sq} \bb 
    \geq \frac{c_1 \|\bX \bB \bb\|_2^2}{2 c_2 n_0} \Big\} \rightarrow 1
    \label{equ::general-sq-min-eigen-val}
\end{align}
as $n_0 \rightarrow \infty$.
Combining \eqref{equ::general-sq-min-eigen-val} with the fact that $\|\bX \bB \bb\|_2^2 \geq s_{\min} (\min_{m \in [|\g|]}|\g_m|)$, it holds that
$$
    \Prob\Big\{c_{\min}(\g) \geq c_1 s_{\min} (\min_{m \in [|\g|]}|\g_m|) / (2c_2n_0)\Big\} \rightarrow 1.
$$ 
In general, if $c_1 s_{\min} (\min_{m \in [|\g|]}|\g_m|) / (2c_2n_0) > \lambda_1\lambda_2^{-1}(2M^*+1)$ for all $\g$ satisfying $|\g|\leq(M^*)^2$, then the restricted eigenvalue condition in Theorem \ref{theorem::lasso-solution} is satisfied with probability tending to one as $n_0 \rightarrow  \infty$.

Recall that $\hat\bbeta^{ols}$ is a local minimizer of $\tilde S(\bbeta)$ on event $\mathcal{F}$ and the Hessian matrix of $\tilde S(\bbeta)$ can be written as
\begin{align*}
    \bH_{sq} + \lambda_1 \lambda_2^{-1} \big\{ \bone_{|\g|}\bone_{|\g|}^T - (|\g|+1) \bI_{|\g|} \big\},
\end{align*}
where the smallest eigenvalue of $\bH_{sq}$ is $c_{\min}$ and the smallest eigenvalue of $\lambda_1 \lambda_2^{-1} \big\{ \bone_{|\g|}\bone_{|\g|}^T - (|\g|+1) \bI_{|\g|}$ is $-\lambda_1 \lambda_2^{-1} (|\g|+1)$.
By the assumption in \eqref{assumption::restricted-eigen-value-condi}, it holds that $c_{\min}(\g) > \lambda_1 \lambda_2^{-1} (|\g|+1)$. 
Hence, by Weyl's inequality, the Hessian matrix of $\tilde S(\bbeta)$ is positive definite. 
This implies that $\tilde S(\bbeta)$ is strictly convex in \(\boldsymbol{\beta}\), and consequently $\hat\bbeta^{ols}$ is the unique minimizer of $\tilde S(\bbeta)$ on event $\mathcal{F}$.

\textit{Step 3}. 
Given that $\hat\bbeta^{ols}$ is the unique minimizer of $\tilde S(\bbeta)$, the proof of the claim that it is also the unique minimizer of $S(\bbeta)$ on event $\mathcal{F}$ is identical to that of Theorem 3 in \cite{shen2010grouping}, for which we omit the details but provide a high-level outline here. 
The main idea is to show that $S(\bbeta)=\tilde S(\bbeta)$ over the set $E := \{\bbeta:||\eta_k-\eta_{\ell}|-\lambda_2|>\lambda_2/2, 1 \leq k < \ell \leq |\g|, \bfeta  \in \mathbb R^{|\g|}\}$, while $S(\bbeta)$ has no local minimizer in the complement set $E^c$ on event  $\mathcal{F}$ under assumption \eqref{assumption::restricted-eigen-value-condi}.
Combining this with the result in Step 1, it follows that $\hat\bbeta^{ols}$ is a local minimizer of $S(\bbeta)$ in set $E$ on event $\mathcal{F}$.
By construction, we have $S(\bbeta)=\tilde S(\bbeta)$ over set $E$. 
Consequently, $\hat\bbeta^{ols}$ is also a local minimizer of $\tilde S(\bbeta)$ in set $E$ on event $\mathcal{F}$ and its uniqueness follows directly from Step 2. Therefore, $\hat\bbeta^{ols}$ is the unique minimizer of $S(\bbeta)$ in set $E$ on event $\mathcal{F}$, and thus the unique minimizer of $S(\bbeta)$ on event $\mathcal{F}$.

\textit{Step 4}. 
By the Step 3 result that $\hat\bbeta^{ols}$ is the unique minimizer of $S(\bbeta)$ on event $\mathcal{F}$, we can obtain immediately that 
$$\hat\bbeta^{grp} = \hat\bbeta^{ols}$$ on event $\mathcal{F}$. This entails that 
$$\Prob(\hat\g \neq \g^0) \leq \Prob(\hat\bbeta^{grp} \neq \hat\bbeta^{ols}),$$ which is further upper bounded by $\Prob(\mathcal{F}^c)$.

We now turn to examining $\Prob(\mathcal{F}^c)$. 
Recall that the definition of $\hat\bbeta^{ols}$ is based on the oracle grouping $\g^0$.
For each $1 \leq k < \ell \leq M^0$, it holds that  
\begin{align}
    \hat\eta_k^{ols}-\hat\eta_\ell^{ols} \sim N\big(\eta_k^0 -\eta_\ell^0, \Var(\hat\eta_k^{ols}-\hat\eta_\ell^{ols})\big),
    \label{equ::con-helper-1}
\end{align}
where 
\begin{align}
    \Var(\hat\eta_k^{ols}-\hat\eta_\ell^{ols}) \leq 2 \max \{\text{diag}(\bD_{}^T \bD_{})^{-1}\} \sigma_0^2 = 2s_{\min}^{-1}\sigma_0^2
    \label{equ::con-helper-2}
\end{align}
with $s_{\min}:=\min_{i \in [M^0]}\{\bone_{n_0}^T\bD_{}\be_i\}$ the smallest group size of nodes with the same value of the interference function under the oracle grouping $\g^0$. 

Combining \eqref{equ::con-helper-1}--\eqref{equ::con-helper-2} and the definition $\xi_{\min} = \min\{ |\eta_{k}^0 - \eta_{\ell}^0|: 1 \leq k < \ell \leq M^0\}$, it holds that
\begin{align} \nonumber
    \Prob\big\{\min_{1\leq k<\ell \leq M^0} |\hat\eta_k^{ols}-\hat\eta_\ell^{ols}| \leq \frac{3 \lambda_2}{2}\big\}
    &\leq \sum_{1\leq k<\ell \leq M^0} \Prob\Bigg\{ \frac{|\hat\eta_k^{ols}-\hat\eta_\ell^{ols}| - |\eta_k^0 -\eta_\ell^0|}{\sqrt{\Var(\hat\eta_k^{ols}-\hat\eta_\ell^{ols})}} \leq \frac{3 \lambda_2/2 - |\eta_k^0 -\eta_\ell^0|}{\sqrt{\Var(\hat\eta_k^{ols}-\hat\eta_\ell^{ols})}} \Bigg\}\\ \nonumber
    &\leq \sum_{1\leq k<\ell \leq M^0} \boldsymbol{\Phi} \Bigg(\frac{3 \lambda_2/2 - \xi_{\min}}{\sqrt{\Var(\hat\eta_k^{ols}-\hat\eta_\ell^{ols})}} \Bigg)\\ \label{con-helper-3}
    &\leq \frac{M^0(M^0-1)}{2} \boldsymbol{\Phi}\Big(\frac{-\sqrt{s_{\min}} (\xi_{\min}-3\lambda_2/2)}{\sqrt{2} \sigma_0}\Big).
\end{align}
Using \eqref{con-helper-3}, condition A) in Theorem \ref{theorem::lasso-solution}, and the fact that 
\begin{align}
\label{equation::inequ-1}
    \boldsymbol{\Phi}(-|x|) \leq \sqrt{2/\pi}/|x|\exp(-x^2/2)
\end{align}
with $\boldsymbol{\Phi}(\cdot)$ the cumulative distribution function (CDF) of the standard normal distribution, as $n_0, d \rightarrow \infty$, we have that $\Prob\big\{\min_{1\leq k<\ell \leq M^0} |\hat\eta_k^{ols}-\hat\eta_\ell^{ols}| \leq 3 \lambda_2/2\big\} \rightarrow 0$.
Moreover, by assumption, it holds that $M^0 < \sqrt{n_0}$. 
It follows from the least-squares property that the average of squared errors concentrates around $\sigma_0^2$ that 
\begin{align*}
    \Prob\big\{\sqrt{(2n_0)^{-1}\|\by- \bX\hat\bbeta^{ols}\|^2_2} < \sigma_0/2\big\}
    &= \Prob\big\{\|\by- \bX\hat\bbeta^{ols}\|^2_2 / n_0  < \sigma_0^2/2 \big\} \\
    &= \Prob\big\{\|\by- \bD_{}\hat\bfeta^{ols}\|^2_2 / n_0  < \sigma_0^2/2 \big\} \\
    &\rightarrow  0
\end{align*}
as $n_0 \rightarrow \infty$.

Furthermore, it holds that 
\begin{align*}
    \bx_j^T(\by - \bX\hat\bbeta^{ols})/\sigma_0
    \sim
    N(0, \| \{\bI_{n_0} - \bD_{} (\bD_{}^T \bD_{})^{-1} \bD_{}^T \} \bx_j \|_2^2),
\end{align*}
where $\| \{\bI_{n_0} - \bD_{} (\bD_{}^T \bD_{})^{-1} \bD_{}^T \} \bx_j \|_2^2 \leq \| \bx_j \|_2^2$.
With an application of \eqref{equation::inequ-1} and condition B) in Theorem \ref{theorem::lasso-solution}, as $n_0, d \rightarrow \infty$, we can obtain that 
\begin{align*}
    & \Prob\Big[ \bigcap_{m:|\g_m^0| \geq 2} \big\{ \max_{j:j \in \g_m^0} \Big| \bx_j^T(\by - \bX\hat\bbeta^{ols})/\sigma_0 \Big| > n_0 \lambda_1 (|\g_m^0|-1) \big\} \Big] \\
    &= \sum_{m=1}^{M^0} \sum_{j \in \g_m^0} \Prob\Big[ \big\{ \Big| \bx_j^T(\by - \bX\hat\bbeta^{ols})/\sigma_0 \Big| > n_0 \lambda_1 (|\g_m^0|-1) \big\} \Big] \\
    &\leq d \cdot \boldsymbol{\Phi}\Big( \frac{-n_0 \lambda_1}{\max_{i \in [d]} \|\bx_i\|_2} \Big) \rightarrow 0.
\end{align*}
Therefore, it holds that $\Prob(\mathcal{F}^c)\rightarrow 0$ as $n_0,d \rightarrow \infty$, which concludes the proof of Theorem \ref{theorem::lasso-solution}.

\begin{remark}
    In Step 2, consider a special case for \eqref{equ::general-sq-min-eigen-val} when the partition of $n_0$ untreated nodes based on neighborhood size $k$ has group sizes of the same order, i.e., $n_0/d$, then it holds that for any $\bb \in \mathbb{R}^{|\g|}$ with $\|\bb\|_2=1$, as $n_0 \rightarrow \infty$ we have
    $$
        \Prob \Big\{\bb^T \bH_{sq} \bb \geq c_1 (\min_{m \in [|\g|]}|\g_m|) / (2c_2d) \Big\} \rightarrow 1,
    $$ 
    indicating that $c_{\min}(\g) \geq c_1 (\min_{m \in [|\g|]}|\g_m|) (2c_2d)^{-1}$ with probability tending to one. 
\end{remark}

\subsection{Proof of Theorem \ref{theorem::SFL-infer-tau-asy-cp}} \label{sec.supp.A8}

Given $\hat\g$, we can match $\hat f_{i,k}$ with $\hat\bbeta^{grp}$ based on the interference network, denoted as $\hat f_{i,k}^{grp}$.
We will show that
\begin{align}    
\label{equation::asy-normal-fgrp-f}
    \sigma_0^{-1} (\tilde{\bv}^T (\bD_{\hat\g}^T \bD_{\hat\g})^{-1} \tilde{\bv})^{-1/2} 
    \frac{1}{\sum_{i=1}^n Z_i}\sum_{i=1}^nZ_i (\hat f_{i,k}^{grp} - f_i) \xrightarrow{d} N(0, 1).
\end{align}
Then, combining \eqref{equation::OR-diff} and \eqref{equation::asy-normal-fgrp-f}, and noting that the error terms in \eqref{def::potential-outcome-model} of the treated nodes are independent of the inference procedure of the values of interference function based on the untreated nodes, we can obtain that 
\begin{align*} 
    \sigma_0^{-1} (\tilde{\bv}^T (\bD_{\hat\g}^T \bD_{\hat\g})^{-1} \tilde{\bv} + n_1^{-1})^{-1/2} 
    (\hat \tau^{OR}_{\rm sfl} - \tau)
    \xrightarrow{d}
    N(0, 1)
\end{align*}
as $n_0 \rightarrow \infty$. Similarly, applying Lemma \ref{lemma::DR-asym-independent-f-and-error} for the DR estimator, we have that 
\begin{align*} 
    \sigma_0^{-1} \Big\{\tilde{\bu}^T (\bD_{\hat\g}^T \bD_{\hat\g})^{-1} \tilde{\bu} + n_1^{-1} + \sum_{i=1}^n\frac{(1-Z_i) p_i^2}{n_1^2 (1-p_i)^2 } \Big\}^{-1/2} 
    (\hat \tau^{DR}_{\rm sfl} - \tau)
    \xrightarrow{d}
    N(0, 1)
\end{align*}
as $n_0 \rightarrow \infty$. Finally, it follows from the asymptotic properties of the OLS regression that a consistent estimator of $\sigma_0^2$ takes the form $$\by_{obs}^T (\bI_{n_0} -\bH_{\hat\g}) \by_{obs} / (n_0 - |\hat\g|).$$ 
Combining the above results yields the desired conclusions of Theorem \ref{theorem::SFL-infer-tau-asy-cp}. 

We proceed to prove \eqref{equation::asy-normal-fgrp-f}. It follows from Theorem \ref{theorem::lasso-solution} and the boundedness of the interference function that 
\begin{align*} 
    & \frac{1}{\sum_{i=1}^n Z_i}\sum_{i=1}^nZ_i(f_i - \hat f_{i,k}^{grp})
    -
    \frac{\mathbbm{1}\{\hat\g = \g^0\}}{\sum_{i=1}^n Z_i}\sum_{i=1}^nZ_i(f_i - \hat f_{i,k}^{grp})   \\ 
    &= 
    \frac{\mathbbm{1}\{\hat\g \neq \g^0\}}{\sum_{i=1}^n Z_i}\sum_{i=1}^nZ_i(f_i - \hat f_{i,k}^{grp}) 
    \xrightarrow{p} 0.
\end{align*}
Hence, to characterize the asymptotic behavior of $\frac{1}{\sum_{i=1}^n Z_i}\sum_{i=1}^nZ_i(f_i - \hat f_{i,k}^{grp})$, we need only to focus on the event when $\hat\g = \g^0$.
By rewriting $\frac{1}{\sum_{i=1}^n Z_i}\sum_{i=1}^nZ_i \hat f_{i,k}^{grp}$ as $\bv^T \hat \bbeta^{grp}$, we can deduce that 
\begin{align}
\label{equation::infer-tau-explain-2}
\bv^T \hat \bbeta^{grp} - \bv^T \hat \bbeta^{ols} \xrightarrow{p} 0
\end{align}
and 
\begin{align} 
 \label{equation::infer-tau-explain-3}
    \frac{\mathbbm{1}\{\hat\g = \g^0\}}{\sum_{i=1}^n Z_i}\sum_{i=1}^nZ_i(f_i - \hat f_{i,k}^{grp}) - \frac{\mathbbm{1}\{\hat\g = \g^0\}}{\sum_{i=1}^n Z_i}\sum_{i=1}^nZ_i(f_i - \hat f_{i,k}^{ols})  
    \xrightarrow{p} 0,
\end{align}
where $\hat f_{i,k}^{ols}$ is based on matching $\hat f_{i,k}$ with $\hat\bbeta^{ols}$ under $\g^0$.

Note that $\hat\bfeta^{grp} = (\bD_{\hat\g}^T \bD_{\hat\g})^{-1} \bD_{\hat\g}^T \by_{obs}$, where $\bD_{\hat\g}$ is the design matrix based on the estimated grouping $\hat\g$ of covariates.
By the property of the least-squares regression, when $\hat\g = \g^0$ it holds that  $$\hat\bfeta^{grp}=\hat\bfeta^{ols} := (\hat\eta_1^{ols}, \ldots, \hat\eta_{M^0}^{ols})^T$$ and 
$$\sigma_0^{-1} (\bD_{\hat\g}^T \bD_{\hat\g})^{1/2} (\hat\bfeta^{ols} - \bfeta^0) \xrightarrow{d} \mathcal{N}(\bzero, \bI_{M^0}).$$
From the definition, we have the weighted average $\bv^T \hat \bbeta^{ols} = \tilde{\bv}^T \hat\bfeta^{ols}$ for some $\tilde{\bv} \in \mathbb{R}^{M^0}$, which gives 
$$\sigma_0^{-1} (\tilde{\bv}^T (\bD_{\hat\g}^T \bD_{\hat\g})^{-1} \tilde{\bv})^{-1/2} (\bv^T \hat \bbeta^{ols} - \tilde{\bv}^T \bfeta^0) \xrightarrow{d} N(0, 1)$$
with $\tilde{\bv}^T \bfeta^0 = \bv^T \bbeta^0$.
Then by \eqref{equation::infer-tau-explain-2}--\eqref{equation::infer-tau-explain-3} and an application of Slutsky's theorem, we can obtain that 
$$\sigma_0^{-1} (\tilde{\bv}^T (\bD_{\hat\g}^T \bD_{\hat\g})^{-1} \tilde{\bv})^{-1/2} (\bv^T \hat \bbeta^{grp} - \tilde{\bv}^T \bfeta^0) \xrightarrow{d} N(0, 1),$$
which finishes the proof of \eqref{equation::asy-normal-fgrp-f}.
This completes the proof of Theorem \ref{theorem::SFL-infer-tau-asy-cp}.

\subsection{Proof of Theorem \ref{thm::confidence-cover-1}} \label{sec.supp.A12}

Let us first introduce two lemmas below to facilitate the proof.

\begin{lemma}[\cite{wang2022finite}]
    \label{thm::wang2022finite-3}
    Given $\bW_k(\bY)=\bw$, if the conditional distribution of $\hat {k}(\bY)$ is free of $(\bbeta_{k}, \sigma)$ and there exists a Borel set $\mathcal{B}_{\alpha}(k,\bw)$ satisfying \eqref{equation::nuclear-mapping-borel-set}, it holds that $\Prob\{k_0 \in \Gamma_{\alpha}(\bY)\} \geq 1 - \alpha$ with $\Gamma_{\alpha}(\bY)$ as defined in \eqref{equation::def-confidence-set}.
\end{lemma}

\begin{lemma}
\label{thm::candidate-cover-1}
Assume that $n_0 - d(k_0) > 4$. For each $\delta > 0$, there exists a constant $\gamma_{\delta}>0$ such that when $\lambda \in \Big[\gamma_{\delta} \sigma^2_0 / (\sqrt{1+\frac{2}{3}\gamma^{\frac{1}{4}}_{\delta}} - 1), n_0 \gamma^{1/4}_{\delta} \frac{C_{\min}}{6}\Big] $, the finite-sample probability bound that the true model is not included in the model candidate set  $\mathcal S_B$, calculated from Algorithm \ref{algorithm::candidate-set}, is given by 
\begin{align}
\label{equation::confidence-set-k0-as-B-inf}
    \Prob(k_0 \notin \mathcal S_B) \leq (1 - \frac{\gamma^{n_0-1}_{\delta}}{n_0-1})^{B} + \delta.
\end{align}
\end{lemma}

In view of Lemma \ref{thm::candidate-cover-1} above, for a fixed number of treated nodes and feasible $\lambda$, we have $$\Prob(k_0 \notin \mathcal S_B) \rightarrow \delta$$ as $B \rightarrow \infty$, where $\delta>0$ is arbitrarily small. 
The proof of Lemma \ref{thm::candidate-cover-1} is given in Section \ref{sec.supp.A9}.
Combining Proposition \ref{proposition::construct-borel-set}, Lemma \ref{thm::wang2022finite-3}, and \eqref{equation::simp-form-considence-set}, it holds for any finite $n_0$ and $d(k_0)$, significance level $\alpha \in (0,1)$, and any $\delta>0$ that 
\begin{align}
\label{equation:ttt}
    \Prob(k_0 \notin \bar\Gamma_{\alpha}(\by_{obs})) 
    \leq&~ \Prob(k_0 \notin \Gamma_{\alpha}(\by_{obs})) + \Prob(k_0 \notin \mathcal S_B) \nonumber\\
    \leq&~ \alpha + \Prob(k_0 \notin \mathcal S_B).
\end{align}
Further, from Lemma \ref{thm::candidate-cover-1}, we can obtain that 
$$\Prob(k_0 \notin \mathcal S_B) \leq \delta + o(e^{-c_1B})$$ for some $c_1 < -\log\big(1- \gamma^{n_0-1}_\delta/(n_0-1)\big)$. Combing the results above concludes the proof of Theorem \ref{thm::confidence-cover-1}.

\subsection{Proof of Theorem \ref{thm::confidence-cover-2}} \label{sec.supp.A13}

Similar to the proof of Theorem \ref{thm::confidence-cover-1} in Section \ref{sec.supp.A12}, let us first introduce a key lemma below, whose proof is given in Section \ref{sec.supp.A10}.

\begin{lemma}
\label{thm::candidate-cover-2}
Assume that $$\frac{\lambda}{n_0} \in \left[ \sigma^2_0(1 + \frac{2}{n_0}) + m_1, \min \left \{ 0.015 C_{\min} - \frac{3\sigma^2_0 \ln2 }{n_0}, 0.015 C_{\min} - \frac{\sigma^2_0 d(k_0) }{n_0}\right\} - m_2\right]$$  for positive constants $m_1, m_2>0$. Then the finite-sample probability bound that the true model is not included in the confidence set $\mathcal S_B$, obtained from Algorithm \ref{algorithm::candidate-set}, is given by 
    \begin{align}
        &\Prob(k_0 \notin \mathcal S_B) \notag\\
        &\leq 4 \exp \left\{ - \frac{n_0 m_1}{3 \sigma^2_0}   \right\} +  4 \exp \left\{\frac{2}{3}  - \frac{n_0 m_2}{18 \sigma^2_0}\right\} \nonumber\\
        &\quad+ \exp \left\{ -n_0 B \Big[ \log(\frac{5}{3})\frac{n_0 - d(k_0) - 1}{n_0} -\frac{\log (2k_0)}{n_0}  \Big]\right\}.
    \end{align}   
\end{lemma}

By resorting to Lemma \ref{thm::candidate-cover-2} above,  we can show that 
$$\Prob(k_0 \notin \mathcal S_B) \rightarrow 0$$ for any $B$ as $n_0 \rightarrow \infty$, provided that  $\frac{\log (2k_0)}{n_0 - d(k_0) - 1} < \log(\frac{5}{3})$ and $\lambda$ is in the feasible region. Thus, Algorithm \ref{algorithm::candidate-set} will capture $k_0$ with overwhelming probability whenever the number of treated nodes $n_0$ or the the number of Monte Carlo copies $B$ goes to infinity, under the above conditions on the penalty level $\lambda$.
In light of \eqref{equation:ttt}, the conclusion follows directly from Lemma \ref{thm::candidate-cover-2} by setting
\begin{align*}
    0<c_2< \min\left\{\frac{m_1}{3 \sigma^2_0},~ \frac{m_2}{18 \sigma^2_0},~ B \Big\{ \log(\frac{5}{3})\frac{n_0 - d(k_0) - 1}{n_0} -\frac{\log (2k_0)}{n_0}  \Big\} \right\}.
\end{align*}
This completes the proof of Theorem \ref{thm::confidence-cover-2}.

\subsection{Proof of Proposition \ref{proposition::infer-tau-model1-overfit-underfit}} \label{sec.supp.A1}

For simplicity, let us denote the true model as
\begin{align}
\label{quation::true-model} 
    \wt\by_{obs} =
    \begin{bmatrix}
        \bz & \wt\bX_{k_0}
    \end{bmatrix}
    \begin{bmatrix}
        \tau \\
        \bbeta_{k_0}^0
    \end{bmatrix}
    +
    \bepsilon_{},
\end{align}
where $\bbeta_{k_0}^0 \in \mathbb{R}^{d(k_0)}$.
Accordingly, for neighborhood size $k \geq k_0$, the outcome model takes the form
\begin{align*}
    \wt\by_{obs} =
    \begin{bmatrix}
        \bz & \wt\bX_{k}
    \end{bmatrix}
    \begin{bmatrix}
        \tau \\
        \bbeta_{k}^0
    \end{bmatrix}
    +
    \bepsilon,
\end{align*}
where $\bbeta_k \in \mathbb{R}^{d(k)}$ and $\mathcal{C}(\bX_{k_0}) \subseteq \mathcal{C}(\bX_k)$.
Recall that $\bD_k = \big[ \bz ~ \wt\bX_k \big]$.
Since it holds that $\mathcal{C}(\bX_{k_0}) \subseteq \mathcal{C}(\bX_k)$, there exists a matrix $\bM^* \in \mathbb{R}^{d(k) \times d(k_0)}$ such that
\begin{align} 
\nonumber
    \Expected(
    \begin{bmatrix}
        \hat \tau_k \\
        \hat \bbeta_{k}
    \end{bmatrix}) 
    =&~ 
    \Expected\{(\bD_k^T \bD_k)^{-1} \bD_k^T 
    \wt\by_{obs} \}
    = 
    (\bD_k^T \bD_k)^{-1} \bD_k^T \bD_{k_0} 
    \begin{bmatrix}
        \tau \\
        \bbeta_{k_0}^0
    \end{bmatrix}\\ \label{equation::overfit-decomp}
    =&~ 
    (\bD_k^T \bD_k)^{-1} \bD_k^T \bD_k
    \begin{bmatrix}
        1 & \bzero_{d(k_0)}^T\\
        \bzero_{d(k)} &  \bM^*
    \end{bmatrix}
    \begin{bmatrix}
        \tau \\
        \bbeta_{k_0}^0
    \end{bmatrix}.
\end{align}
Hence, we have $\Expected(\hat\tau_k)=\tau$ when $k \geq k_0$.
Observe that the decomposition in \eqref{equation::overfit-decomp} above does not hold for $k < k_0$ in general.

Further, under the setting when $k \geq k_0$, the asymptotically normal approximation for the ordinary least squares (OLS) estimator results from \cite{yohai1979asymptotic}.
By the least-squares property, the width of the asymptotic level $(1-\alpha)$ CI for $\tau$ is given by 
$
    \hat \tau(k) \pm \boldsymbol{\Phi}^{-1}(1-\alpha/2) (\hat\sigma^2_k)^{-1/2}
$,
where
\begin{align}
\label{equ::CI-def}
    \hat\sigma^2_k := \wt\by_{obs}^T (\bI_n -  \bD_k(\bD_k^T\bD_k)^{-1}\bD_k^T) \wt\by_{obs} / (n-d(k)-1) \be_1^T(\bD_k^T\bD_k)^{-1}\be_1.
\end{align}

We now proceed to show that for any $k_1 > k_2 \geq k_0$, when $\bD_{k_1}$ and $\bD_{k_2}$ are of full column rank, it holds that  $\Expected(\hat\sigma^2_{k_1}) \geq \Expected(\hat\sigma^2_{k_2})$.
Without loss of generality, we prove that $\Expected(\hat\sigma^2_{k})\geq\Expected(\hat\sigma^2_{k_0})$ for any $k\geq k_0$.
Since the mean squared error estimator is unbiased for $\sigma_0^2$, we have 
\begin{align}
\label{equ::std-helper-1}
    \Expected\left\{\frac{\wt\by_{obs}^T (\bI_n -  \bD_k(\bD_k^T\bD_k)^{-1}\bD_k^T) \wt\by_{obs}}{n-d(k)-1}\right\}
    =
    \Expected\left\{\frac{\wt\by_{obs}^T (\bI_n -  \bD_{k_0}(\bD_{k_0}^T\bD_{k_0})^{-1}\bD_{k_0}^T) \wt\by_{obs}}{n-d(k_0)-1}\right\}
\end{align}
Combining \eqref{equ::CI-def} with \eqref{equ::std-helper-1}, it remains to show that 
$$\be_1^T(\bD_k^T\bD_k)^{-1}\be_1 \geq \be_1^T(\bD_{k_0}^T\bD_{k_0})^{-1}\be_1.$$ 

Switching the order of the covariates, we can \textit{reformulate} the regression problem as 
\begin{align*}
    \wt\by_{obs} =
    \begin{bmatrix}
        \wt\bX_{k} & \bz
    \end{bmatrix}
    \begin{bmatrix}
        \bbeta_{k} \\
        \tau
    \end{bmatrix}
    +
    \bepsilon.
\end{align*}
Since the rank of matrix 
$
    \bD^*_k :=
    \begin{bmatrix}
        \wt\bX_{k} & \bz
    \end{bmatrix}
$
is $d(k) + 1$, we will consider the Gram--Schmidt orthogonalization of columns in $\bD^*_k$; that is, there exists a matrix $\bM \in \mathbb{R}^{(d(k)+1) \times (d(k)+1)}$ of rank $d(k)+1$ to be defined later such that 
$$(\bD^*_k\bM)^T(\bD^*_k\bM) = \bI_{d(k)+1}.$$
Consequently, it holds that 
$$\bM \bM^T = \{(\bD^*_k)^T \bD^*_k\}^{-1}.$$
Thus, it suffices to calculate the last diagonal entry of $\bM \bM^T$, which, by construction, is equal to $\be_1^T(\bD_k^T\bD_k)^{-1}\be_1$.

Applying the Gram--Schmidt orthogonalization of columns in $\bD^*_k$, we can deduce that 
\begin{align*}
    \bM = 
    \begin{bmatrix}
        1/\sqrt{\sum_{i=1}^n \bD^*_{i,1}} & 0 & \ldots & 0 & c^* \bz^T \bD^*_{\cdot,1}/\sqrt{\sum_{i=1}^n \bD^*_{i,1}} \\
        0 & 1/\sqrt{\sum_{i=2}^n \bD^*_{i,1}} & \ldots & 0 & c^* \bz^T \bD^*_{\cdot,2}/\sqrt{\sum_{i=1}^n \bD^*_{i,2}}\\
        \vdots & \vdots & \ddots & 0 & \vdots \\
        0 & 0 & \ldots & 1/\sqrt{\sum_{i=1}^n \bD^*_{i,d(k)}} & c^* \bz^T \bD^*_{\cdot,d(k)} /\sqrt{\sum_{i=1}^n \bD^*_{i,d(k)}} \\
        0 & 0 & \ldots & 0 & c^* 
    \end{bmatrix},
\end{align*}
where $c^* = \big\|\bz - \sum_{i=1}^{d(k)} \frac{\bz^T \bD^*_{\cdot,i}}{\sum_{j=1}^n \bD^*_{j,i}} \bD^*_{\cdot,i} \big\|_2^{-1}$,  $\bD^*_{\cdot,i}$ is the $i$th column of $\bD^*_k$, and $\bD^*_{i,j}$ is the $(i,j)$th entry of $\bD^*_k$. Hence, it follows that 
\begin{align}
\label{equation::Gram-Schmidt}
    \be_1^T(\bD_k^T\bD_k)^{-1}\be_1
    =
    \big\|\bz - \sum_{i=1}^{d(k)} \frac{\bz^T \bD^*_{\cdot,i}}{\sum_{j=1}^n \bD^*_{j,i}} \bD^*_{\cdot,i} \big\|_2^{-1}.
\end{align}
Observe that $\mathcal{C}(\bD^*_{k_0}) \subseteq  \mathcal{C}(\bD^*_k)$ when $k \geq k_0$. Therefore, combining \eqref{equation::Gram-Schmidt} with the projection theorem, we can obtain that 
$$\be_1^T(\bD_k^T\bD_k)^{-1}\be_1 \geq \be_1^T(\bD_{k_0}^T\bD_{k_0})^{-1}\be_1,$$ 
which concludes the proof of Proposition \ref{proposition::infer-tau-model1-overfit-underfit}.

\subsection{Proof of Proposition \ref{proposition::construct-borel-set}} \label{sec.supp.A11}

By definition, it holds that 
\begin{align*}
    \mathcal{B}_{\alpha}(k,\bw)
    =
    \big\{ 0 \leq k' \leq n_0: \mathcal{F}(k'|\bW_k(\bY)=\bw) \geq \alpha \big\},
\end{align*}
which is free of $(\bbeta_{k}, \sigma_0)$.
Recall that $\hat {k}(\bY)$ is also independent of $(\bbeta_{k}, \sigma_0)$. Hence, it follows that 
\begin{align*}
    \Prob\{\hat {k}(\bY) \in \mathcal{B}_{\alpha}(k,\bw) | \bW_k(\bY)=\bw\} 
    =&~ \Prob \big\{ \mathcal{F} \big(\hat {k}(\bY) | \bW_k(\bY)=\bw \big) \geq \alpha \big\}  \\
    =&~ \sum_{k' \in \mathcal{B}_{\alpha}(k,\bw)} p(k'|\bw) \\
    =&~ 1 - \sum_{k' \notin \mathcal{B}_{\alpha}(k,\bw)} p(k'|\bw) \\
    \geq&~ 1 - \alpha.
\end{align*}
Moreover, we have that 
\begin{align*}
    \Prob \big\{\hat {k}(\bY) \in \mathcal{B}_{\alpha}\big(k,\bW_k(\bY)\big) \big\}  
    =
    \Expected \Big[ \Prob\big\{\hat {k}(\bY) \in \mathcal{B}_{\alpha}(k,\bw) | \bW_k(\bY)\big\} \Big]
    \geq 1 - \alpha,
\end{align*} 
which concludes the proof of Proposition \ref{proposition::construct-borel-set}.

\section{Proofs of some key lemmas} \label{sec.supp.B}

To facilitate the technical analysis, let us introduce some necessary notation. We simplify the Euclidean norm $\|\cdot\|_2$ as $\|\cdot\|$ whenever there is no ambiguity.
Denote by $$\rho(\bv_1,\bv_2) = \cos^{2}(\bv_1,\bv_2) = \frac{\| \bH_{\bv_2} \bv_1\|^2}{\| \bv_1 \|^2}$$ the square of the cosine of the angle between two vectors. For each given $k$, let 
$$\rho_{k^{\perp}}(\bv_1,\bv_2) = \rho\{ (\bI - \bH_{k})\bv_1 , (\bI - \bH_{k})\bv_2\}$$ be the square of the cosine of the angle between $(\bI - \bH_{k})\bv_1$ and $(\bI - \bH_{k})\bv_2$. Further, denote by $ d(k) = |\hat E_{k}|$ the number of patterns for a given neighborhood $k$. Hereafter, we use the shorthand notation $\bbeta_{k_0}^0 =\bbeta_{k_0}$.

\subsection{Proof of Lemma \ref{lemma::DR-asym-independent-f-and-error}} \label{sec.supp.A4}

Recall that $\{\hat f_{i,k}\}_{i \in [n]}$ are estimated from the untreated nodes and $\{\epsilon_{i,1}\}_{i \in [n]}$ are the error terms for the treated nodes.
We need only to show the asymptotic independence of 
\begin{align}
\label{equ::asy-independence-component-1}
    \big(\sum_{i=1}^n Z_i\big)^{-1}\sum_{i=1}^n \big\{Z_i(f_i - \hat f_{i,k}) - ((1-Z_i)(f_i-\hat f_{i,k}) p_i ) / (1-p_i)\big\}
\end{align}
and 
\begin{align}
\label{equ::asy-independence-component-2}
    \big(\sum_{i=1}^n Z_i\big)^{-1}\sum_{i=1}^n ((1-Z_i) p_i \epsilon_{i,0}) / (1-p_i)
\end{align}
with $f_i = f\big(\gamma_0(G_i^{\bz}(k_0))\big)$.

According to the matching procedure of the interference functions between the treated and untreated nodes, there exists some $\bv_1 \in \mathbb R^{d(k)}$ such that
\begin{align*}
    \big(\sum_{i=1}^n Z_i\big)^{-1}\sum_{i=1}^n Z_i \hat f_{i,k}
    =
    \bv_1^T \hat \bbeta_{k},
\end{align*}
where $\bone_{d(k)}^T\bv_1 = 1$,  
$
\hat \bbeta_{k} = (\bX_{k}^T\bX_{k})^{-1} \bX_{k}^T \by_{obs}
$
is the OLS estimate, and $d(k)$ is the number of distinct features of untreated nodes based on the $k$-hop neighborhood as defined in Section \ref{sec::matching-estimator}.
Moreover, there exists some $\bv_2 \in \mathbb R^{n_0}$ such that $\bone_{n_0}^T\bv_2 = 1$ and $\bv_1^T = \bv_2^T \bX_{k}$, which entails that
\begin{align}
\label{equation::asy-ind-term-1}
    \big(\sum_{i=1}^n Z_i\big)^{-1}\sum_{i=1}^n Z_i \hat f_{i,k}
    =
    \bv_2^T \bX_{k} \hat \bbeta_{k}.
\end{align}
Similarly, for the untreated nodes, we have that 
\begin{align}
\label{equation::asy-ind-term-2}
    \sum_{i=1}^n (1-Z_i) \hat f_{i,k}
    =
    \bone_{n_0}^T \bX_{k} \hat \bbeta_{k}
\end{align}
and $\sum_{i=1}^n (1-Z_i) \epsilon_{i,0} = \bone_{n_0}^T \bepsilon_0$.

Denote by $\delta_i=p_i/(1-p_i)$ and let $\bdelta$ represent the vector of $\delta_i$'s where $Z_i=0$.
Combining \eqref{equation::asy-ind-term-1} and \eqref{equation::asy-ind-term-2}, for any $a_1,a_2\in\mathbb{R}$, we can deduce that
\begin{align} \nonumber
    &~ a_1 \frac{\sum_{i=1}^n\big\{Z_i(f_i - \hat f_{i,k}) - (1-Z_i)(f_i-\hat f_{i,k}) p_i / (1-p_i) \big\}}{\sum_{i=1}^n Z_i}
    +
    a_2 \frac{\sum_{i=1}^n (1-Z_i) p_i \epsilon_{i,0}}{\sum_{i=1}^n Z_i (1-p_i)}\\ \nonumber
    =&~ a_1 \bv_2^T \bX_{k} (\bbeta_{k}^0 - \hat \bbeta_{k}) 
    - a_1 \bdelta^T \bX_{k} (\bbeta_{k}^0 - \hat \bbeta_{k}) / n_1 
    + a_2 \bdelta^T \bepsilon_0 / n_1 \\ \nonumber
    =&~ \big\{- a_1 \bv_2^T \bX_{k}(\bX_{k}^T \bX_{k})^{-1} \bX_{k} + a_1 \bdelta^T \bX_{k}(\bX_{k}^T \bX_{k})^{-1} \bX_{k} / n_1 + a_2 \bdelta^T / n_1 \big\} \bepsilon_0,
\end{align}
where $\bveps_0 \sim \mathcal{N}(0, \sigma_0^2 \bI_{n_0})$.
This establishes the joint normality between \eqref{equ::asy-independence-component-1} and \eqref{equ::asy-independence-component-2}. Moreover, the covariance of the two components is given by 
\begin{align} \label{equation::asy-inde-cov}
    &~ \Cov\Bigg( 
    \frac{\sum_{i=1}^n\big\{Z_i(f_i - \hat f_{i,k}) - (1-Z_i)(f_i-\hat f_{i,k}) p_i / (1-p_i) \big\}}{\sum_{i=1}^n Z_i}
    ,
    \frac{\sum_{i=1}^n (1-Z_i) p_i \epsilon_{i,0}}{\sum_{i=1}^n Z_i (1-p_i)}
    \Bigg)\\ \nonumber
    =&~ \frac{-1}{n_1} \Cov(\bv_2^T \bX_{k} \hat \bbeta_{k}, \bdelta^T \bepsilon_0) 
    +
    \frac{1}{n_1^2} \Cov(\bdelta^T \bX_{k} \hat \bbeta_{k}, \bdelta^T \bepsilon_0) \\ \nonumber
    =&~
    \frac{-1}{n_1} \Cov(\bv_2^T \bX_{k} (\bX_{k}^T \bX_{k})^{-1} \bX_{k} \bepsilon_0, \bdelta^T \bepsilon_0)
    + \\ \nonumber
    &~
    \frac{1}{n_1^2} \Cov(\bdelta^T \bX_{k} (\bX_{k}^T \bX_{k})^{-1} \bX_{k} \bepsilon_0, \bdelta^T \bepsilon_0) \\ \nonumber
    =&~
    \frac{-1}{n_1} \bv_2^T \bX_{k} (\bX_{k}^T \bX_{k})^{-1} \bX_{k} \bdelta \sigma_0^2 + \frac{1}{n_1^2} \bdelta^T \bX_{k} (\bX_{k}^T \bX_{k})^{-1} \bX_{k}  \bdelta  \sigma_0^2 \\ \label{equation::asy-inde-order}
    \asymp&~ \sigma_0^2 / n.
\end{align}
The asymptotic order in \eqref{equation::asy-inde-order} above holds because of the boundedness of $\delta_i$'s induced by Definition \ref{definition::potential-outcome-model} and the fact that $\bone_{n_0}^T\bv_2 = 1$.
Therefore, given treatments $\bZ$, the covariance between the two components in \eqref{equation::asy-inde-cov} goes to zero as the number of nodes increases.
This completes the proof of Lemma \ref{lemma::DR-asym-independent-f-and-error}.

\subsection{Proof of Lemma \ref{thm::candidate-cover-1}} \label{sec.supp.A9}

The proof follows similar arguments as those in \cite{wang2022finite}, based on Lemmas \ref{lem::angle}--\ref{lem::finite_penalize_complete} below. 
The proofs of Lemmas \ref{lem::angle}--\ref{lem::finite_penalize_complete} are presented in Sections \ref{sec.supp.B1}--\ref{sec.supp.B3}, respectively.

\begin{lemma}
    \label{lem::angle}
     For any $-1 \leq \gamma_1,\gamma_2 \leq 1$, if $\bU^{*} \thicksim \bU \thicksim N(0,\bI)$ it holds that 
    \begin{align*}
       \mathbb{P}_{\bU^{*}}\{ \rho_{k^{\perp}}(\bU^{*},\bX_{k_0} \bbeta_{k_0}^0) <\gamma^{2}_1\} & = \mathbb{P}_{\bU}\{ \rho_{k^{\perp}}(\bU, \bX_{k_0}\bbeta_{k_0}^0 ) 
       < \gamma^{2}_1 \} \\
       &> 1 - 2 \{\arccos(\gamma_1)\}^{n_0 - d(k) - 1}
    \end{align*}
    and 
    \begin{align*}
        \mathbb{P}_{(\bU^{*},\bU)}
        \{ \rho(\bU^{*},\bU) > 1- \gamma^{2}_2\} > \frac{\gamma^{n-2}_2 \arcsin(\gamma_2)}{n_0-1}.
    \end{align*}
    Moreover, $\rho_{k^{\perp}}(\bU^{*},\bX_{k_0} \bbeta_{k_0}^0)$ and $\rho(\bU^{*},\bU)$ are independent. 
\end{lemma}

Lemma \ref{lem::angle} above controls the angle between $\bU$ and $\bU^*$, and the angle between $\bU, \bU^*$ and $\bX_{k_0}\bbeta_{k_0}^0$. Furthermore, we will introduce Lemma \ref{lem::finite_penalize} below, which bounds the probability that the neighborhood size obtained through the repro samples differs from the true neighborhood size. 
Specifically, it provides the probability bound of obtaining the true $k_0$ when the repro sample $\bu^*$ falls within close proximity of $\bu^{rel}$. 

\begin{lemma}
    \label{lem::finite_penalize}
Assume that $n_0-d(k_0)>4$. 
Let $\bU^*$ be a random repro sample of $\bU$ such that $\bU^*, \bU \sim N(0, \bI_{n_0})$, and
$$\hat{k}_{\bU^*} = \argminA_{k}\left\{\min_{\bbeta_k, \sigma} \|\bY - \bX_{k} \bbeta_{k} -\sigma \bU^* \|^2 + \lambda k\right\}.$$
Then for any $0<\gamma_2 < 0.005$ with $C_{\min} >52{\sqrt{\gamma_2}}\left( \frac{\log k_0}{2n_0}+ \gamma_2\right) \sigma_0^2$ and $$ \lambda \in \left[\frac{\gamma_2 \sigma^2_0}{\sqrt{1 + \frac{2}{3}\gamma^{\frac{1}{4}}_2} - 1}, n_0\gamma_2^{1/4}\frac{C_{\min}}{6}\right],$$
it holds that 
\begin{align}\label{eq:lemma2}
    & \Prob_{(\bU, \bU^*)}\left\{\hat k_{\bU^*} \neq k_0, \rho(\bU^*, \bU) > 1- \gamma_2^2\right\}\notag \\
    & \leq   2  \exp \left\{ n_0(-\frac{ C_{\min}(1 - \gamma^2_1)^2}{26 \gamma_2 \sigma^2_0} + 2 \gamma_2+ \frac{\log k_0}{n_0}
  ) \right\} +  4 \exp\left \{ - \frac{(1 - \gamma^2_1) \lambda}{4 \gamma_2 \sigma^2_0} + 2n_0 \gamma_2\right\} \\
     & \quad+  2(\arccos \tilde\gamma_1)^{n_0-1} \sum_{k<k_0} \frac{1}{(\arccos \tilde{\gamma}_1)^{d(k)}}  + k_0 (\frac{n_0-1}{2})^{\frac{d(k_0)}{2}} \gamma^{n_0-2}_2,\notag
\end{align}
where $\gamma_1 = \sqrt{1-\gamma_2^{1/4}}$ and $\tilde\gamma_1 = (1-\sqrt{\gamma_2})\sqrt{1-\gamma_2^{1/4}}- \sqrt{\frac{2-2\sqrt{1-\gamma^2_2}}{\gamma_2}}$.

\end{lemma}

We next present Lemma \ref{lem::finite_penalize_complete} below, which provides an upper bound for the probability that the true neighborhood size does not lie in the repro sample set constructed by Algorithm \ref{algorithm::candidate-set}.

\begin{lemma}
    \label{lem::finite_penalize_complete}
    Assume that $n_0-d(k_0)>4$. Then for any $0<\gamma_2<0.005$ with $C_{\min} >52{\sqrt{\gamma_2}}\left( \frac{  \log k_0}{2n_0}+ \gamma_2\right) \sigma_0^2$, $ \lambda \in \bigg[\frac{\gamma_2 \sigma^2_0}{\sqrt{1 + \frac{2}{3}\gamma^{\frac{1}{4}}_2} - 1}, \allowbreak 
         n_0\gamma_2^{1/4}\frac{C_{\min}}{6}\bigg]$, and $\tilde\gamma_1 = (1-\sqrt{\gamma_2})\sqrt{1-\gamma_2^{1/4}}- \sqrt{\big(2-2\sqrt{1-\gamma^2_2}\big)/\gamma_2}$,
    a finite-sample probability bound that the true model is not covered by the model candidates set $\mathcal S_B$, obtained by Algorithm~\ref{algorithm::candidate-set} with objective function \eqref{eq:obj-func}, is given by 
    \begin{align}
        \label{eq:prob_bound_penalize_finite}
          & \Prob_{({\cal U}^d, \bY)}(k_0 \notin \mathcal S_B )   \nonumber\\
         & \leq 2  \exp \left\{ n_0(-\frac{ C_{\min}}{26 \sqrt{\gamma_2} \sigma^2_0} + 2 \gamma_2+ \frac{\log k_0}{n_0}
          ) \right\} +  4 \exp\left \{ - \frac{ \lambda}{4 \gamma^{\frac{3}{4}}_2 \sigma^2_0} + 2n_0 \gamma_2\right\} \nonumber\\
         &\quad +  2(\arccos \tilde\gamma_1)^{n_0-1} \sum_{k<k_0} \frac{1}{(\arccos \tilde{\gamma}_1)^{d(k)}}  + k_0 (\frac{n_0-1}{2})^{\frac{d(k_0)}{2}} \gamma^{n_0-2}_2 + \left(1-\frac{\gamma_2^{n_0-1}}{n_0-1}\right)^B.
    \end{align}
\end{lemma}

Finally, the conclusion of Lemma \ref{thm::candidate-cover-1} can be obtained by directly applying the result in Lemma \ref{lem::finite_penalize_complete} above. To be specific, the first four terms in \eqref{eq:prob_bound_penalize_finite} converge to $0$ as $\gamma_2$ goes to $0$.  Hence, for any $\delta >0, $ there exists some $\gamma_\delta>0$ such that when $\gamma_2=\gamma_\delta$, the sum of the first four terms of \eqref{eq:prob_bound_penalize_finite} is smaller than $\delta$, which entails the probability bound in Lemma \ref{thm::candidate-cover-1}. This completes the proof of Lemma \ref{thm::candidate-cover-1}.

\subsection{Proof of Lemma \ref{thm::candidate-cover-2}} \label{sec.supp.A10}

The proof of Lemma \ref{thm::candidate-cover-2} is similar to that of Lemma \ref{thm::candidate-cover-1} in Section \ref{sec.supp.A9}.
Let us first introduce a key Lemma \ref{lem::symptotic_bound_penalize0} below, which provides an additional probability bound on the difference between the neighborhood size obtained through the repro samples and the true neighborhood size under the finite-sample setting. 
The proof of Lemma \ref{lem::symptotic_bound_penalize0} is given in Section \ref{sec.supp.B4}.

\begin{lemma}
    \label{lem::symptotic_bound_penalize0}
    For any finite $n_0$, if $$\frac{\lambda}{n_0} \in \left[ \sigma^2_0(1 + \frac{2}{n_0}) + m_1, \min \left \{ \frac{(1-\gamma^2_1)^2C_{\min}}{6} - \frac{3\sigma^2_0 \log 2 }{n_0}, \frac{(1-\gamma^2_1)^2C_{\min}}{6} - \frac{\sigma^2_0 d(k_0) }{n_0}\right\} - m_2\right],$$ 
    a finite-sample probability bound that the true model is not covered by the model candidate set $\mathcal S_B$, obtained by Algorithm~\ref{algorithm::candidate-set} with objective function \eqref{eq:obj-func}, is given by 
    \begin{align}
    \label{eq:prob_bound_penalize_asymptotic}
         & \Prob_{({\cal U}^d, \*Y)}(k_0 \notin \mathcal S_B ) \nonumber\\
         & \leq 4 \exp \left\{ - \frac{n_0 m_1}{3 \sigma^2_0}   \right\} +  4 \exp \left\{\frac{2}{3}  - \frac{n_0 m_2}{18 \sigma^2_0}\right\} + \left[2 k_0\{\arccos (\gamma_1)\}^{n-d(k_0)-1 }\right]^B, 
    \end{align}
    where $\cos(0.3\pi) < \gamma_1<1$ is any real number. 
\end{lemma}

Then we are ready to see that Lemma \ref{thm::candidate-cover-2} follows directly from Lemma \ref{lem::symptotic_bound_penalize0} by setting $\gamma^2_1 = 0.7$, which concludes the proof of Lemma \ref{thm::candidate-cover-2}.

\subsection{Proof of Lemma \ref{lem::angle}} \label{sec.supp.B1}

Denote by 
$$\bI - \bH_k = \sum_{i=1}^{n_0-d(k)} D_i D^{T}_i$$ the eigen-decomposition of matrix $\bI - \bH_k$, where $D_i$'s are eigenvectors of the projection matrix $\bI - \bH_k$. Let us define $V_i = D^{T}_i \bU$ and $w_i = D^{T}_{i} \bX_{k_0} \bbeta 
_{k_0}^0$ for $i = 1,..., n_0 - d(k)$. It follows that $V_1,...,V_{n_0-d(k)}$ are independent and identically distributed (i.i.d.) $N(0,1)$ and
\begin{align}\label{eq:010}
    \mathbb{P}_{\bU^{*}} \{ \rho_{k^{\perp}}(\bU^{*},\bX_{k_0} \bbeta_{k_0}^0) < \gamma^{2}_1 \} & = 
    \mathbb{P}_{\bU} \{ \rho_{k^{\perp}}(\bU,\bX_{k_0} \bbeta_{k_0}^0) < \gamma^{2}_1\}  \nonumber \\
   & = \mathbb{P}_{\bU}\{ \frac{\sum_{i=1}^{n_0-d(k)} w_i V_i}{\sqrt{\sum_{i=1}^{n_0-d(k)}w^{2}_i} \sqrt{\sum_{i=1}^{n_0-d(k)} V^{2}_{i}} } < \gamma_1\} \nonumber \\
    & = \mathbb{P}_{\bU} \{ |\cos(\varphi)| < \gamma_1\},
\end{align}
where $\varphi = \varphi(\bU)$ (or $\pi$ - $\varphi$) represents the angle between $(V_1,...,V_{n_0-d(k)})$ and $(w_1,...,w_{n_0-d(k)})$ for $0 \leq \varphi \leq \pi$.
We further transform the usual coordinates of $V_1,...,V_{n_0-d(k)}$ into the spherical coordinates, with $\varphi$ being the first angle coordinate. Then by the Jacobian of the spherical transformation, the density function of $\varphi$ is given by 
$$
f(\varphi) = \sin^{n_0-d(k)-2}(\varphi) / c
$$
with $0 \leq \varphi \leq \pi$, where $c = \int_{0}^{\pi} \sin^{n_0-d(k)-2}(\varphi)d\varphi  = 2 \int_{0}^{\frac{\pi}{2}} \sin^{n_0-d(k)-2}(\varphi)d\varphi$  is the normalizing constant.

By the basic inequalities, it holds that 
$$
\frac{2}{\pi}\varphi < \sin(\varphi) < \min \{\varphi,1 \} = \varphi \textbf{1}_{(0 < \varphi < 1)} + \textbf{1}_{(1 \leq \varphi <\frac{\pi}{2})}
$$
for $0<\varphi <\frac{\pi}{2}$. 
Then it follows from the definition of $c$ that
$$
\frac{\pi}{n_0-d(k)-1} < c < 2(\frac{1}{n_0-d(k)-1} + (\frac{\pi}{2}  - 1)) < 4. 
$$
Hence, using the above two displayed results, we can deduce that 
\begin{align}\label{eq:011}
\mathbb{P}_{\bU}\left\{|\cos \varphi| <\gamma_1 \right\}  
=&~ \frac{2}{c}\int_{\arccos (\gamma_1)}^{\pi/2} \sin^{n_0-d(k)-2}(s) ds \nonumber \\
=&~ 1 - \frac{2}{c} \int_0^{{\arccos (\gamma_1)}}\sin^{n_0-d(k)-2}(s) d s \nonumber \\
>&~ 1 - \frac{ 2(n_0-d(k)-1)\int_0^{{\arccos (\gamma_1)}}s^{n_0-d(k)-2} ds}{\pi} \nonumber \\
=&~ 1-  \frac{2\{\arccos (\gamma_1)\}^{n_0-d(k)-1}}{\pi} \nonumber \\
>&~ 1-  2\{\arccos (\gamma_1)\}^{n_0-d(k)-1}.
\end{align}
Combining \eqref{eq:010} and \eqref{eq:011} proves the first desired result in the lemma.

Now we prove the second desired result in the lemma. Conditional on $\bU^* = \bu^*$, using similar arguments as above but replacing $n_0-d(k)$ with $n_0$, we can show that 
\begin{align}\label{eq:u_epsi} 
& \mathbb{P}_{ \bU}\left\{{\|(\bu^*)^\top \bU\|}\big/{(\|\bu^*\|\| \bU\|)} > \sqrt{1 - \gamma_2^2} \bigg| \bu^* \right\}
=\mathbb{P}_{ \bU}\left\{  |\cos(\psi)|  > \sqrt{1 - \gamma_2^2}  
\bigg| \bu^* \right\} \nonumber \\
 & \quad= \frac{2}{c_1}\int_0^{\arcsin \gamma_2} \sin^{n_0-2}(s) ds   \\
 & \quad>  \frac{2}{c_1}\int_0^{\arcsin \gamma_2} (\frac{s \gamma_2 }{\arcsin \gamma_2})^{n_0-2} ds> \frac{\gamma_2^{n_0-2}\arcsin \gamma_2}{n_0-1}, \nonumber 
\end{align}
where $\psi = \psi(\bu^*, \bu)$ (or $\pi - \psi$) represents the angle between $\bu$ and $\bu^*$, and the normalizing constant is given by 
$$c_1 = \int_0^{\pi} \sin^{n_0-2} (\psi) d\psi = 2 \int_0^{\frac \pi 2} \sin^{n_0-2} (\psi) d\psi.$$ 
Similarly, we can obtain that conditional on $\bU=\bu$, 
\begin{align}\label{eq:u_epsi-1}
& \mathbb{P}_{ \bU^*}\left\{{\|\bU^*{}^\top \bu\|}\big/{(\|\bU^*\|\| \bu\|)} > \sqrt{1 - \gamma_2^2} \bigg| \bu \right\}
=\mathbb{P}_{ \bU^*}\left\{  |\cos(\psi)| > \sqrt{1 - \gamma_2^2}   \bigg| \bu \right\} \nonumber\\
& \quad = \frac{2}{c_1}\int_0^{\arcsin \gamma_2} \sin^{n_0-2}(s) ds \nonumber\\
&\quad > \frac{\gamma_2^{n_0-2}\arcsin \gamma_2}{n_0-1}.  
\end{align}

Since \eqref{eq:u_epsi} and \eqref{eq:u_epsi-1} do not involve $\bu^*$ or $\bu$, we can deduce that 
\begin{align*}
\mathbb{P}_{(\bU^*, \bU)}\big\{ \rho(\bU^*, \bU)> 1- \gamma_2^2 \big\} & =    \mathbb{P}_{\bU^{*}}\left\{ \rho(\bU^*, \bU)> 1- \gamma_2^2\bigg| \bU \right\} \\ 
& = \mathbb{P}_{ \bU}\left\{ \rho(\bU^*, \bU)> 1- \gamma_2^2 \bigg| \bU^{*} \right\} \\
& > \frac{\gamma_2^{n-2}\arcsin \gamma_2}{n_0-1}.
\end{align*} 
In particular, we see from the fact of 
$$\mathbb{P}_{(\bU^*, \bU)}\big\{ \rho(\bU^*, \bU)> 1- \gamma_2^2 \big\} = \mathbb{P}_{ \bU}\left\{ \rho(\bU^*, \bU)> 1- \gamma_2^2 \bigg| \bU^{*} \right\}$$
shown above that the conditional distribution of $\rho(\bU^*, \bU)$ given $\bU^*$ is identical to its marginal distribution and does not depend on $\bU^*$. Thus, we have that 
\begin{align}\nonumber
    &\mathbb{P}_{(\bU^*, \bU)}\big\{ \rho(\bU^*, \bU)> 1- \gamma_2^2 \big\} \mathbb{P}_{\bU^*}\big\{ \bU^* = \bu^* \big\}\\ \label{equ::independent}
    & = \mathbb{P}_{(\bU^*, \bU)}\big\{ \rho(\bU^*, \bU)> 1- \gamma_2^2, \bU^* = \bu^*\big\}.
\end{align}
It follows immediately from \eqref{equ::independent} above that $\rho(\bU^*, \bU)$ and $\bU^*$ are independent. Consequently, $\rho(\bU^*,\bU)$ and $\rho_{k^\bot}(\bU^*, \bX_{k_0} \bbeta_{k_0}^0)$ are independent since the only source of randomness in $\rho_{k^\bot}(\bU^*, \bX_{k_0} \bbeta_{k_0}^0)$ is from $\bU^*$, and $\rho(\bU^*,\bU)$ and $\bU^*$ are independent as shown above. This completes the proof of Lemma \ref{lem::angle}.

\subsection{Proof of Lemma \ref{lem::finite_penalize}} \label{sec.supp.B2}

To prove Lemma \ref{lem::finite_penalize}, we aim to bound the probability $\Prob_{(\bU, \bU^*)}\left\{\hat k_{\bU^*} \neq k_0, \rho(\bU^*, \bU) > 1- \gamma_2^2\right\}$. We first introduce Lemma \ref{lemma:u_d} to split this probability into three parts as in \eqref{equ::bound-helper-1} and bound each part separately. To handle the first part, we define $D(k,\bu^*)$ as below and rewrite event $\{\hat k_{\bU^*} \neq k_0\}$ as the union of events characterized by $D(k,\bu^*)$ and $D(k_0,\bu^*)$. We then bound each of these components individually. Finally, we apply Lemmas \ref{lem::angle} and  \ref{lem:epsilon_bound} to bound the target probability $\Prob_{(\bU, \bU^*)}\left\{\hat k_{\bU^*} \neq k_0, \rho(\bU^*, \bU) > 1- \gamma_2^2\right\}.$

Let us first introduce a useful lemma below, which decomposes the projection matrix on the space spanned by 
$\begin{pmatrix}
    \bX_{k} , \bu^{*}
\end{pmatrix}$ 
into two projection matrices. 
\begin{lemma}[\cite{wang2022finite}]
    \label{lem_proj}
    For any $k$ and $\bu^{*}$, it holds that 
    $$
    \bI - \bH_{k,\bu^{*}} = \bI - \bH_{k}- \bO_{k^{\perp}\bu^{*}},
    $$ where $\bH_{k,\bu^{*}} = \begin{pmatrix}
        \bX_{k} & \bu^{*}
    \end{pmatrix} \begin{pmatrix}
        \bX^{T}_{k} \bX_{k} & \bX^{T}_{k} \bu^{*} \\
        (\bu^{*})^{T}\bX_{k} & (\bu^{*})^{T}\bu^{*}
    \end{pmatrix}^{-1} \begin{pmatrix}
        \bX^{T}_{k} \\
        (\bu^{*})^{T}
    \end{pmatrix}$ is the projection matrix on the space spanned by $\begin{pmatrix}
        \bX_{k} , \bu^{*}
    \end{pmatrix}$ and $\bO_{k^{\perp}\bu^{*}}$ is the projection matrix on the space spanned by $(\bI - \bH_{k}) \bu^{*}$.
\end{lemma} 

To invoke Lemma \ref{lem_proj} above, we let 
\begin{equation}\label{eq:002}
    D(k, \bu^*)=  \|(\bI - \bH_{k,\bu^*})\bY\|^2 + 2 \lambda k =  \|(\bI-\bH_k-  \bO_{ k^{\bot} \bu^*})\bY\|^2 + 2 \lambda k
\end{equation}
for each $ k< k_0$. 
Denote by 
\begin{align}
    \hat{k}_{\bu^*} = \argminA_{k} D(k,\bu^*).
\end{align}
If there exists some $k$ such that $D(k,\bu^*) - D(k_0,\bu^*) <0$, it holds that 
$$ \hat{k}_{\bu^*} \neq k_0.$$ 
On the other hand, if $ \hat{k}_{\bu^*} \neq k_0$, we have that 
$$D(\hat{k}_{\bu^*},\bu^*) - D(k_0,\bu^*) <0.$$ 
Hence, it follows that 
\begin{align}
\label{equ:: union expression}
    \bigcup_{k} \{D(k,\bu^*) - D(k_0,\bu^*) <0\} = \{ \hat{k}_{\bu^*} \neq k_0\}.
\end{align}

To further bound the above probability, we introduce the technical lemma below.

\begin{lemma}
    \label{lemma:u_d}
Let 
    \begin{align}
    \label{eq:def_E}
        E(\gamma_1, \gamma_2)= \left\{ \max_{ k < k_0} \rho_{k^\bot}(\bU^*, \bX_{k_0} \bbeta_{k_0}^0) < \gamma_1^2, \, \rho(\bU^*, \bU) > 1- \gamma_2^2  \right\}
    \end{align}
    for any $0<\gamma_1, \gamma_2 <1$, and  $\rho(\bu, k) = \frac{\|\bH_{k} \bu\|^2}{\|\bu\|^2}$. Then it holds that 
    \begin{align}
    \label{eq:def_subset_E}
        \tilde{E}(\gamma_1,\gamma_2) =& \left\{\rho(\bU^*, \bU) > 1- \gamma_2^2, \, \max_{k<k_0} \rho_{k^\bot}(\bU, \bX_{k_0} \bbeta_{k_0}^0) < \tilde\gamma_1^2, \, \max_{k<k_0} \rho(\bU, k) < 1- \gamma_2  \right\}\\ \nonumber
         \subset& E(\gamma_1, \gamma_2),
    \end{align}
where  $\tilde\gamma_1=(1-\sqrt{\gamma_2})\gamma_1 - \sqrt{\frac{2-2\sqrt{1-\gamma^2_2}}{\gamma_2}}$. 
\end{lemma}    
The proof of Lemma \ref{lemma:u_d} above is presented in Section \ref{sec.supp.B5}.
By applying Lemma \ref{lemma:u_d}, the desired result in Lemma \ref{lem::finite_penalize} can be upper bounded as
\begin{align} \nonumber
     & \Prob_{(\bU^*, \bU)}(\hat k_{\bU^*} \neq k_0, \rho(\bU^*, \bU) > 1- \gamma_2^2) \\ \nonumber
     &\leq  \Prob_{(\bU^*, \bU)}\{\hat k_{\bU^*} \neq k_0, \, (\bU^*, \bU) \in E(\gamma_1, \gamma_2)\} + \Prob\left(\max_{k < k_0} \rho_{k^\bot}(\bU, \bX_{k_0} \bbeta_{k_0}^0) \geq \tilde\gamma_1^2\right)\\ \label{equ::bound-helper-1}
     &\hspace{1.4em} + \Prob\left(\max_{k <k_0} \rho(\bU, k) \geq 1- \gamma_2 \right).
\end{align}
Let us first bound the first component on the right-hand side of \eqref{equ::bound-helper-1}. The desired result reduces to 
\begin{align} 
 & \mathbb P\left\{ \{ \hat{k}_{\bu^*} \neq k_0\},(\bU^*, \bU) \in E(\gamma_1, \gamma_2) \right\} \nonumber \\
 & =  \mathbb P\left\{   \bigcup_{k} \{D(k,\bu^*) - D(k_0,\bu^*) <0,(\bU^*, \bU) \in E(\gamma_1, \gamma_2) \right\} \nonumber\\
 \label{equ::indicator}
 &  \leq \sum_k \mathbb P\left\{    D(k,\bu^*) - D(k_0,\bu^*) <0,(\bU^*, \bU) \in E(\gamma_1, \gamma_2) \right\}.
\end{align}

It remains to bound the right-hand side of \eqref{equ::indicator}. With an application of \eqref{eq:002}, we can deduce that 
\begin{align}   
\label{eq:dist_penalize}
& D(k, \bu^*) -D(k_0,\bu^*) \nonumber \\
& =\|(\bI - \bH_{k}-  \bO_{ k^{\bot}  \bu^*})\bY\|^2 - \|(\bI - \bH_{k_0}- \bO_{ k_0^{\bot} \bu^*})\bY\|^2 + \lambda(k - k_0 ) \nonumber\\
&  = \|(\bI - \bH_{k}- \bO_{ k^{\bot}  \bu^*}) \bX_{k_0} \bbeta_{k_0}^0 \|^2 - \bU^\top( \bH_{k}+ \bO_{ k^{\bot}  \bu^*} -  \bH_{k_0}-\bO_{ k_0^{\bot} \bu^*} )\bU \nonumber\\
& \quad   + 2\bU^\top(\bI - \bH_{k}-\bO_{ k^{\bot}  \bu^*})\bX_{k_0} \bbeta_{k_0}^0 + 2\lambda(k-k_0).\end{align}

Then we can show that 
\allowdisplaybreaks
\begin{align}\label{eqn:013}
  \nonumber & \Prob_{(\bU^*, \bU)}\left\{ D(k, \bU^*) -D(k_0,\bU^*)< 0, (\bU^*, \bU) \in E(\gamma_1, \gamma_2) \right\}\\
   \nonumber & \leq \Prob_{(\bU^*, \bU)}\big\{(1- \gamma^2_1) \|(\bI-\bH_k)\bX_{k_0} \bbeta_{k_0}^0\|^2 - \sigma_0^2\bU^\top( \bH_{k}+ \bO_{ k^{\bot}\bU^*}-  \bH_{k_0}- \bO_{ k_0^{\bot} \bU^*} )\bU \\
   \nonumber &\hspace{5.5em} + 2\sigma_0\bU^\top(\bI - \bH_{k}-\bO_{ k^{\bot}  \bU^*})\bX_{k_0} \bbeta_{k_0}^0 +  2\lambda(k-k_0) <0, (\bU^*, \bU) \in E(\gamma_1, \gamma_2) \big\}\\
   \nonumber & \leq  \Prob_{(\bU^*, \bU)}\big\{(1-\gamma^2_1)(1-\delta)\|(\bI-\bH_k)\bX_{k_0} \bbeta_{k_0}^0\|^2 - \sigma_0^2 \bU^\top(\bH_{k}+\bO_{ k^{\bot}  \bU^*} -  \bH_{k_0}-\bO_{ k_0^{\bot} \bU^*} )\bU \\
   \nonumber &\hspace{6.5em} + \lambda(k-k_0) <0, (\bU^*, \bU) \in E(\gamma_1, \gamma_2) \big\} \\
   \nonumber & \quad + \Prob_{(\bU^*, \bU)}\big\{(1-\gamma^2_1)\delta\|(\bI-\bH_k)\bX_{k_0} \bbeta_{k_0}^0\|^2 + 2\sigma_0\bU^\top(\bI - \bH_{k}-\bO_{ k^{\bot}  \bU^*})\bX_{k_0} \bbeta_{k_0}^0 \\
   \nonumber &\hspace{6.5em}  +\lambda(k-k_0) ) < 0, (\bU^*, \bU) \in E(\gamma_1, \gamma_2) \big\} \\
    &= (I_1) + (I_2)
\end{align}
for each $\delta \in (0,1)$.

To derive an upper bound for term $(I_1)$ above, observe that 
\begin{align}
\label{equ::bound-hlper-2-2}
  &\| (\bI - \bH_k- \bO_{ k^{\bot}  \bU^*}) \bU \|^2 
    \leq \gamma_2^2 \|\bU\|^2.
\end{align}
It holds that 
\begin{align}
\nonumber
    & \bU^\top(\bI - \bH_k-\bO_{k^{\bot}  \bU^*})\bX_{k_0} \bbeta_{k_0}^0   =  \bU^\top(\bI - \bH_k-\bO_{ k^{\bot}  \bU^*})(\bI - \bH_k)\bX_{k_0} \bbeta_{k_0}^0
    \\ \label{equ::bound-hlper-1}
    &\leq \|\bU^\top(\bI - \bH_k-\bO_{ k^{\bot}  \bU^*})\| \|(\bI - \bH_k)\bX_{k_0} \bbeta_{k_0}^0\|.
\end{align}
Then we can obtain that  
\begin{align}
\nonumber
     & \bU^\top( \bH_k+\bO_{ k^{\bot}  \bU^*} -  \bH_{k_0}-\bO_{ k_0^{\bot} \bU^*} )\bU \\\nonumber
     & = \bU^\top( \bH_k+\bO_{ k^{\bot}  \bU^*} - \bI -  \bH_{k_0}-\bO_{ k_0^{\bot} \bU^*}  + \bI)\bU  \\  \nonumber
     & = \bU^\top( \bH_k+\bO_{ k^{\bot}  \bU^*} - \bI)\bU -\bU^\top( \bH_{k_0}+\bO_{ k_{0}^{\bot}  \bU^*} - \bI)\bU \\ \label{equ::bound-hlper-2}
    & \leq \bU^\top(  \bI - \bH_{k_{0},\bu^{*}})\bU \leq\gamma^2_2\|\bU\|^2.
\end{align}

Note that $\|\bU\|^2$ follows a chi-square distribution because $\bU$ is a standard normal vector.
Then in view of 
\begin{align*}
     \|\bU^\top(\bI - \bH_k-\bO_{ k^{\bot}  \bU^*})\|^2 < \gamma^2_2  \|\bU\|^2,
\end{align*}
it follows that when $ k < k_0$ and $\frac{\lambda}{n_0} < \frac{1}{6}(1-\gamma_1^2)C_{\min}$,
from \eqref{equ::bound-hlper-1}, \eqref{equ::bound-hlper-2}, and the definition of $C_{\min}$, we have that
\begin{align} \nonumber
    (I_1) & \leq \Prob_{(\bU^*, \bU)}\left\{\|\bU\|^2 > \frac{(1-\gamma^2_1)(1-\delta)}{\gamma^2_2}\frac{\|(\bI-\bH_k)\bX_{k_0}\bbeta_{k_0}^0\|^2 }{\sigma_0^2} + \frac{\lambda(k-k_0)}{\gamma_2^2\sigma_0^2}\right\}\\
    \label{equ:: bound_I_1_first}
    &\leq  \exp\left\{-\frac{n_0}{2}\log(1-2t_1) - t_1\frac{(1-\gamma^2_1)(1-\delta-1/6  )}{\gamma^2_2}\frac{n_0(k_0 - k)C_{\min}}{\sigma_0^2}  \right\}
\end{align}
for any $0<t_1<1/2$. 

Otherwise when $k > k_0$, it follows from the Markov inequality and the moment generating function of the chi-square distribution that 
\begin{align} \nonumber
    (I_1)   &\leq \Prob_{(\bU^*, \bU)}\left\{\|\bU\|^2 > \frac{(1-\gamma^2_1)(1-\delta)}{\gamma^2_2}\frac{\|(\bI-\bH_k)\bX_{k_0}\bbeta_{k_0}^0\|^2 }{\sigma_0^2} + \frac{\lambda(k-k_0)}{\gamma_2^2\sigma_0^2}\right\} \\ \nonumber
    & \leq \Prob_{(\bU^*, \bU)}\left\{\|\bU\|^2 >  \frac{\lambda(k-k_0)}{\gamma_2^2\sigma_0^2}\right\} \\ \nonumber
    & \leq \Prob_{(\bU^*, \bU)}\left\{t_1 \|\bU\|^2 >  t_1 \frac{\lambda(k-k_0)}{\gamma_2^2\sigma_0^2}\right\} \\
  \label{equ::bound_I_1_second}
    & \leq \exp\left\{-\frac{n_0}{2}\log(1-2t_1) -  t_1 \frac{\lambda(k - k_0)}{\gamma_2^2\sigma_0^2} \right\},
\end{align}
where the second inequality above holds because $ \frac{(1-\gamma^2_1)(1-\delta)}{\gamma^2_2}\frac{\|(\bI-\bH_k)\bX_{k_0}\bbeta_{k_0}^0\|^2 }{\sigma_0^2} >0$, and the last step has used the fact that $\|\bU\|^2$ follows a chi-square distribution. 

For term $(I_2)$ above, we follow a similar argument as for $(I_1)$. If 
$\frac{\lambda}{n_0} < \frac{1}{6}(1-\gamma_1^2)C_{\min}$, 
by the definition of $C_{\min}$, the moment generating function of the chi-square distribution, and the Markov inequality,
an application of the Cauchy--Schwartz inequality gives that when $k < k_0$, 
\allowdisplaybreaks
\begin{align} \nonumber
    (I_2) \leq&~ \Prob_{(\bU^*, \bU)}\big\{(1- \gamma^2_1) \delta \|(\bI-\bH_k)\bX_{k_0}\bbeta_{k_0}^0\|^2  <  2\sigma_0\|\bU^\top(\bI - \bH_{k}-\bO_{ k^{\bot}  \bU^*})\| \\ \nonumber
    &\hspace{18em} \times\|(\bI-\bH_k)\bX_{k_0}\bbeta_{k_0}^0\| - \lambda(k-k_0) \big\} \\ \nonumber
    \leq &~ \Prob_{(\bU^*, \bU)}\big\{2\sigma_0\|\bU^\top(\bI - \bH_{k}-\bO_{ k^{\bot}  \bU^*})\| >(1-\gamma_1^2 ) (\delta-1/6) \|(\bI-\bH_k)\bX_{k_0}\bbeta_{k_0}^0\| \big\} \\ \nonumber
    \leq&~ \Prob_{(\bU^*, \bU)}\left\{\|\bU\|^2 > \frac{(1-\gamma^2_1)^2(\delta-1/6)^2}{4\gamma^2_2}\frac{\|(\bI-\bH_k)\bX_{k_0}\bbeta_{k_0}^0\|^2}{\sigma_0^2}\right\}\\ \nonumber
    \leq&~ \Prob \left\{\chi^2_{n_0} > \frac{(1-\gamma^2_1)^2(\delta-1/6)^2}{4\gamma^2_2}\frac{n_0(k_0 - k)C_{\min}}{\sigma_0^2}\right\} \\ \nonumber
     \leq &~ \mathbb{E}  \big\{ \exp(t_2 \chi^2_{n_0} )\big\}\exp\left\{-t_2 \frac{(1-\gamma^2_1)^2(\delta-1/6)^2}{4\gamma^2_2}\frac{n_0(k_0 - k)C_{\min}}{\sigma_0^2} \right\}  \\
     \label{equ::bound_I_2_first}
    =&~ \exp\left\{-\frac{n_0}{2}\log(1-2t_2)- t_2\frac{(1-\gamma^2_1)^2(\delta-1/6)^2}{4\gamma^2_2}\frac{n_0(k_0 - k)C_{\min}}{\sigma_0^2}\right\}
\end{align}
for any $0<t_2<1/2$. 
Specifically, the first inequality above follows from the Cauchy–Schwarz inequality and \eqref{equ::bound-hlper-1}; the second inequality is derived using the condition that $\frac{\lambda}{n_0} < \frac{1}{6}(1-\gamma_1^2)C_{\min}$; 
the third inequality follows from \eqref{equ::bound-hlper-2-2}; the fourth inequality is based on the definition of $C_{\min}$; and applying the Markov inequality, we can derive the fifth inequality above.

When $k  >  k_0$, it follows from the fact that $(1-\gamma^2_1)\delta\|(\bI-\bH_k)\bX_{k_0}\bbeta_{k_0}^0\|^2 + 2\bU^\top(\bI - \bH_{k}-\bO_{ k^{\bot}  \bU^*})\bX_{k_0}\bbeta_{k_0}^0 \geq -\frac{\|\bU^\top(\bI - \bH_{k}-\bO_{ k^{\bot}  \bU^*})\|^2}{(1-\gamma_1^2)\delta}$ that 
\begin{align} \nonumber
   (I_2) 
     &\leq \Prob_{(\bU^*, \bU)}\left\{\|\bU^\top(\bI - \bH_{k}-\bO_{ k^{\bot}  \bU^*})\|^2 > (1-\gamma_1^2)\delta\lambda(k-k_0)\right\} \\ \nonumber
    &\leq \Prob_{(\bU^*, \bU)}\left\{\|\bU\|^2> \frac{(1-\gamma_1^2)\delta\lambda(k-k_0)}{\gamma_2^2 \sigma_0^2}\right\}\\ \nonumber
    & \leq  \mathbb{E}  \big\{ \exp(t_2 \|\bU\|^2 ) \big\}\exp\left\{-t_2\frac{(1-\gamma_1^2)\delta\lambda(k-k_0)}{\gamma_2^2 \sigma_0^2} \right\}   \\   
    \label{equ::bound_I_2_second}
    & =  \exp\left\{-\frac{n_0}{2}\log(1-2t_2)-t_2\frac{(1-\gamma_1^2)\delta\lambda(k-k_0)}{\gamma_2^2 \sigma_0^2}\right\},
\end{align}
where the third inequality above employs the Markov inequality, and the last equality is calculated using the moment-generating function of the chi-square distribution. 

Now by setting $(1-\gamma^2_1)(1 - \delta-1/6) =  (1-\gamma^2_1)^2(\delta-1/6)^2/4$, we have 
$$\delta = \frac{2}{1- \gamma^2_1}(\sqrt{\frac{5}{3} - \frac{2}{3}\gamma_1^2} - 1) + \frac{1}{6}.$$ 
Further, let us choose $t_1=t_2= \frac{\gamma_2}{2.04}$, which entails that 
$$-\frac{n_0}{2}\log(1-2t_1)=-\frac{n}{2}\log(1-2t_2)=-\frac{n_0}{2}\log(1-\frac{\gamma_2}{1.02}) \leq 2n_0\gamma_2.$$
Then intersecting with event $\{(\bU^*, \bU) \in E(\gamma_1, \gamma_2)\}$, by \eqref{equ::indicator} and \eqref{eqn:013}, we can deduce that
\begin{align} \nonumber
    &  \Prob_{(\bU^*, \bU)}\big\{\hat k_{\bU^*} \neq k_0, \, (\bU^*, \bU) \in E(\gamma_1, \gamma_2)\big\} \\ \nonumber
    & \leq \sum_{k<k_0}\Prob_{(\bU^*, \bU)} \big\{D(k,\bu^{*}) - D(k_0,\bu^*) < 0, (\bU^*, \bU) \in E(\gamma_1, \gamma_2) \big\}  \\ \nonumber
    &\hspace{1.5em} + \sum_{k>k_0}\Prob_{(\bU^*, \bU)} \big\{D(k,\bu^{*}) - D(k_0,\bu^*) < 0, (\bU^*, \bU) \in E(\gamma_1, \gamma_2) \big\} \\ \nonumber
    & \leq 2\sum_{k=0}^{k_0-1} \exp\left\{ -(\sqrt{\frac{5}{3} - \frac{2}{3}\gamma^2_1} - 1)^2 \frac{n_0(k_0 - k)C_{\min}}{2.04 \gamma_2 \sigma^2_0} + 2n_0\gamma_2 \right\}\\ \nonumber
    &\hspace{1.5em} + 2\sum_{k=k_0+1}^{n} \exp \left\{ -\frac{\lambda (k-k_0) (\sqrt{\frac{5}{3} - \frac{2}{3}\gamma^2_1} - 1)}{1.02 \gamma_2 \sigma^2_0} + 2n_0 \gamma_2 \right\} \nonumber \displaybreak[1] \\ \nonumber
    & \leq 2 \exp\left\{ -(\sqrt{\frac{5}{3} - \frac{2}{3}\gamma^2_1} - 1)^2 \frac{n_0 k_0 C_{\min}}{2.04 \gamma_2 \sigma^2_0} + 2n_0\gamma_2 \right\} \frac{\exp\left\{ (\sqrt{\frac{5}{3} - \frac{2}{3}\gamma^2_1} - 1)^2 \frac{n_0 k_0 C_{\min}}{2.04 \gamma_2 \sigma^2_0}  \right\} - 1}{\exp\left\{ (\sqrt{\frac{5}{3} - \frac{2}{3}\gamma^2_1} - 1)^2 \frac{n_0 C_{\min}}{2.04 \gamma_2 \sigma^2_0}\right\} - 1}\\ 
    \label{equ::target_prob}
    &\hspace{1.4em} + 2 \exp\left\{ (\sqrt{\frac{5}{3} - \frac{2}{3}\gamma^2_1} - 1) \frac{ \lambda k_0}{1.02 \gamma_2 \sigma^2_0} + 2n_0\gamma_2 \right\} \frac{\exp\left\{- (\sqrt{\frac{5}{3} - \frac{2}{3}\gamma^2_1} - 1) \frac{\lambda (k_0+1)}{1.02 \gamma_2 \sigma^2_0}  \right\}}{1 - \exp\left\{ -(\sqrt{\frac{5}{3} - \frac{2}{3}\gamma^2_1} - 1) \frac{\lambda}{1.02 \gamma_2 \sigma^2_0}\right\} },
\end{align}
where the first inequality above simply uses $k_0$ to partition the range of $k$, the second inequality is a combination of the bounds for $(I_1)$ in \eqref{equ:: bound_I_1_first}--\eqref{equ::bound_I_1_second} and $(I_2)$ in \eqref{equ::bound_I_2_first}--\eqref{equ::bound_I_2_second}, and the last inequality is derived from the sum of a geometric sequence.

To bound the first term on the very right-hand side of \eqref{equ::target_prob}, we use the fact that $(\sqrt{\frac{5}{3} - \frac{2}{3}\gamma_1^2} - 1)^2 \geq 1.02 (1 - \gamma_1^2)^2/13$ with $\gamma_1^2 \in (0,1)$ and choose $-\frac{ C_{\min}(1 - \gamma^2_1)^2}{26 \gamma_2 \sigma^2_0} + 2 \gamma_2+ \frac{\log k_0}{n_0} < 0 $. Then we can show that 
\begin{align*}
  & 2 \exp\left\{ -(\sqrt{\frac{5}{3} - \frac{2}{3}\gamma^2_1} - 1)^2 \frac{n_0 k_0 C_{\min}}{2.04 \gamma_2 \sigma^2_0} + 2n_0\gamma_2 \right\} \frac{\exp\left\{ (\sqrt{\frac{5}{3} - \frac{2}{3}\gamma^2_1} - 1)^2 \frac{n_0 k_0 C_{\min}}{2.04 \gamma_2 \sigma^2_0}  \right\} - 1}{\exp\left\{ (\sqrt{\frac{5}{3} - \frac{2}{3}\gamma^2_1} - 1)^2 \frac{n_0 C_{\min}}{2.04 \gamma_2 \sigma^2_0}\right\} - 1}\\
  & \leq 2  \exp \left\{ n_0(-\frac{ C_{\min}(1 - \gamma^2_1)^2}{26 \gamma_2 \sigma^2_0} + 2 \gamma_2+ \frac{\log k_0}{n_0}
  ) \right\}.
\end{align*}
As for the second term on the very right-hand side of \eqref{equ::target_prob}, it holds that 
\begin{align*}
   & 2 \exp\left\{ (\sqrt{\frac{5}{3} - \frac{2}{3}\gamma^2_1} - 1) \frac{ \lambda k_0}{1.02 \gamma_2 \sigma^2_0} + 2n_0\gamma_2 \right\} \frac{\exp\left\{- (\sqrt{\frac{5}{3} - \frac{2}{3}\gamma^2_1} - 1) \frac{\lambda (k_0+1)}{1.02 \gamma_2 \sigma^2_0}  \right\}}{1 - \exp\left\{ -(\sqrt{\frac{5}{3} - \frac{2}{3}\gamma^2_1} - 1) \frac{\lambda}{1.02 \gamma_2 \sigma^2_0}\right\} } \\
    &\leq 4 \exp\left \{ - \frac{(1 - \gamma^2_1) \lambda}{4 \gamma_2 \sigma^2_0} + 2n_0 \gamma_2\right\},
\end{align*}
provided that $(\sqrt{\frac{5}{3} - \frac{2}{3}\gamma^2_1} - 1) \frac{\lambda}{1.02 \gamma_2 \sigma^2_0} >  \log 2$, i.e., $$\lambda > \frac{\gamma_2 \sigma^2_0 1.02 \log2}{\sqrt{\frac{5}{3} - \frac{2}{3}\gamma^2_1} - 1}.$$
Combining the above two results, we can obtain that
\begin{align}
   \nonumber &  \Prob_{(\bU^*, \bU)}\big\{\hat k_{\bU^*} \neq k_0, \, (\bU^*, \bU) \in E(\gamma_1, \gamma_2)\big\}\\
     &\leq 2  \exp \left\{ n_0(-\frac{ C_{\min}(1 - \gamma^2_1)^2}{26 \gamma_2 \sigma^2_0} + 2 \gamma_2+ \frac{\log k_0}{n_0}
  ) \right\} + 4 \exp\left \{ - \frac{(1 - \gamma^2_1) \lambda}{4 \gamma_2 \sigma^2_0} + 2n_0 \gamma_2\right\}.
\end{align}

In light of \eqref{equ::bound-helper-1}, to complete the proof, we will need to bound the second and third terms on the right-hand side therein. To this end,
we introduce one additional technical lemma below.

\begin{lemma}
    \label{lem:epsilon_bound}
    Assume that $n_0-d(k_0) > 4$. Then for any $0< \tilde\gamma_1, \gamma_2 < 1$, it holds that 
    \begin{align*}
        & \mathbb{P}\left(\bU \not\in  \left\{\max_{k<k_0} \rho_{k^\bot}(\bU, \bX_{k_0} \bbeta_{k_0}^0) < \tilde\gamma_1^2,  \max_{k<k_0} \rho(\bU, k) < 1- \gamma_2  \right\}\right)\\
        & \leq \mathbb{P}\left(\max_{k<k_0} \rho_{k^\bot}(\bU, \bX_{k_0} \bbeta_{k_0}^0) \geq \tilde\gamma_1^2\right) + \mathbb{P}\left(\max_{k<k_0} \rho(\bU, k) \geq 1- \gamma_2 \right)  \\
         & \leq 2(\arccos \tilde\gamma_1)^{n_0-1} \sum_{k<k_0} \frac{1}{(\arccos \tilde{\gamma}_1)^{d(k)}}  + k_0 (\frac{n_0-1}{2})^{\frac{d(k_0)}{2}} \gamma^{n_0-2}_2.
    \end{align*}
\end{lemma}     

The proof of Lemma \ref{lem:epsilon_bound} above is presented in Section \ref{sec.supp.B6}. By resorting to Lemmas Lemma \ref{lem::angle} and \ref{lem:epsilon_bound}, and in view of \eqref{equ::bound-helper-1}, we can deduce that 
\begin{align*} 
     & \Prob_{(\bU^*, \bU)}(\hat k_{\bU^*} \neq k_0, \rho(\bU^*, \bU) > 1- \gamma_2^2) \\
  \nonumber
     & \leq 2  \exp \left\{ n_0(-\frac{ C_{\min}(1 - \gamma^2_1)^2}{26 \gamma_2 \sigma^2_0} + 2 \gamma_2+ \frac{\log k_0}{n_0}) \right\} +  4 \exp\left \{ - \frac{(1 - \gamma^2_1) \lambda}{4 \gamma_2 \sigma^2_0} + 2n_0 \gamma_2\right\} \\ \nonumber
     &\hspace{1.4em} +  2(\arccos \tilde\gamma_1)^{n_0-1} \sum_{k<k_0} \frac{1}{(\arccos \tilde{\gamma}_1)^{d(k)}}  + k_0 (\frac{n_0-1}{2})^{\frac{d(k_0)}{2}} \gamma^{n_0-2}_2. 
\end{align*}

Let us now choose $\gamma_1 = \sqrt{1-\gamma_2^{1/4}}$, which leads to 
$$\tilde\gamma_1=(1-\sqrt{\gamma_2})\sqrt{1-\gamma_2^{1/4}}- \sqrt{\frac{2-2\sqrt{1-\gamma^2_2}}{\gamma_2}} \in [\sqrt{2}/2,1]$$ for $\gamma_2 \in [0,0.005]$. Then it follows that $$\arccos\tilde\gamma_1 \leq \arccos (\frac{\sqrt{2}}{2}) = \frac{\pi}{4} < 1. $$  
Hence, the probability bound above reduces to 
\begin{align*}
     & \Prob_{(\bU^*, \bU)}(\hat k_{\bU^*} \neq k_0, \rho(\bU^*, \bU) > 1- \gamma_2^2) \\
     & \leq 2  \exp \left\{ n_0(-\frac{ C_{\min}}{26 \sqrt{\gamma_2} \sigma^2_0} + 2 \gamma_2+ \frac{\log k_0}{n_0}
      ) \right\} +  4 \exp\left \{ - \frac{ \lambda}{4 \gamma^{\frac{3}{4}}_2 \sigma^2_0} + 2n_0 \gamma_2\right\} \\
     &\hspace{1.4em} +  2(\arccos \tilde\gamma_1)^{n_0-1} \sum_{k<k_0} \frac{1}{(\arccos \tilde{\gamma}_1)^{d(k)}}  + k_0 (\frac{n_0-1}{2})^{\frac{d(k_0)}{2}} \gamma^{n_0-2}_2,
\end{align*}
where $\tilde\gamma_1 = (1-\sqrt{\gamma_2})\sqrt{1-\gamma_2^{1/4}}- \sqrt{\frac{2-2\sqrt{1-\gamma^2_2}}{\gamma_2}}$. Therefore, the conclusion of Lemma \ref{lem::finite_penalize} follows immediately. This concludes the proof of Lemma \ref{lem::finite_penalize}.

\subsection{Proof of Lemma \ref{lem::finite_penalize_complete}} \label{sec.supp.B3}

Let us first introduce a technical Lemma \ref{lem:angle_ud} below.

\begin{lemma}
[\cite{wang2022finite}]
\label{lem:angle_ud}
Assume that $\bU^*_1, \ldots, \bU^*_B$ are $d$ i.i.d. copies of $\bU^* \sim N(0, \bI_{n_0})$. Then it holds that 
\begin{align*}
\Prob\left(\bigcap_{b=1}^B\{\rho(\bU^*_b, \bU) \leq  1- \gamma_2^2\} \right) \leq \left(1-\frac{\gamma_2^{n_0-1}}{n_0-1}\right)^B.
\end{align*}
\end{lemma}

We can decompose the event $\{k_0 \notin \mathcal S_B\}$ as 
\begin{align}
\label{eq:prob_decompose}
     & \Prob(k_0 \notin \mathcal S_B) \nonumber \\
     & = \Prob\left(k_0 \notin \mathcal S_B,\bigcup_{b=1}^B\{\rho(\bU_b^*, \bU) >  1- \gamma_2^2\} \right) + \Prob\left(k_0 \notin \mathcal S_B,\bigcap_{b=1}^B\{\rho(\bU_b^*, \bU) \leq  1- \gamma_2^2\} \right)\nonumber\\
     & \leq \Prob\left(\hat k_{\bU_b^*} \neq k_0,\rho(\bU_b^*, \bU) >  1- \gamma_2^2 \mbox{ for some } b \right) + \Prob\left(\bigcap_{b=1}^B\{\rho(\bU_b^*, \bU) \leq  1- \gamma_2^2\} \right)\nonumber\\
     & \leq \Prob\left(\hat k_{\bU^*} \neq k_0,\rho(\bU^*, \bU) >  1- \gamma_2^2 \right) + \Prob\left(\bigcap_{b=1}^B\{\rho(\bU_b^*, \bU) \leq  1- \gamma_2^2\} \right).
\end{align}
Then we see that the conclusion of Lemma \ref{lem::finite_penalize_complete} follows immediately from Lemmas \ref{lem::finite_penalize} and \ref{lem:angle_ud}, which completes the proof of Lemma \ref{lem::finite_penalize_complete}.

\subsection{Proof of Lemma \ref{lem::symptotic_bound_penalize0}} \label{sec.supp.B4}

Similar to the proof of Lemma \ref{lem::finite_penalize} in Section \ref{sec.supp.B2}, let us define 
$$D(k, \bu^*)=  \|(\bI - \bH_{k,\bu^*})\bY\|^2 + 2\lambda k =  \|(\bI-\bH_k-  O_{ k_0^{\bot} \bu^*})\bY\|^2 + 2 \lambda k$$
for each $ k< k_0$. 

In view of  \eqref{eq:dist_penalize}, for any $\delta \in (0,1)$ and any given $\bu^*$, by similar derivation as in \eqref{eqn:013}, it holds that
\begin{align} \nonumber
   & \Prob_{ \bU}\{ D(k, \bu^*) -D(k_0,\bu^*)< 0\}\\ \nonumber
    & \leq \Prob_{\bU}\{(1- \gamma^2_1) \|(\bI-\bH_k)\bX_{k_0}\bbeta_{k_0}^0\|^2 -\sigma_0^2\bU^\top( \bH_{k}+\bO_{ k^{\bot}  \bu^*} -  \bH_{k_0}-\bO_{ k_0^{\bot} \bu^*} )\bU \\ \nonumber
    &\hspace{3.4em} + 2\sigma_0\bU^\top(\bI - \bH_{k}-\bO_{ k^{\bot}  \bu^*})\bX_{k_0}\bbeta_{k_0}^0 +  2\lambda(k-k_0) <0 \}\\ \nonumber
    & \leq  \Prob_{\bU}\{(1-\gamma^2_1)(1-\delta)\|(\bI-\bH_k)\bX_{k_0}\bbeta_{k_0}^0\|^2 - \sigma_0^2\bU^\top( \bH_{k}+\bO_{ k^{\bot}  \bu^*} -  \bH_{k_0}-\bO_{ k_0^{\bot} \bu^*} )\bU\\ \nonumber
    &\hspace{3.4em} + \lambda(k-k_0) <0\} \\  \nonumber
    & \quad + \Prob_{\bU}\{(1-\gamma^2_1)\delta\|(\bI-\bH_k)\bX_{k_0}\bbeta_{k_0}^0\|^2 + 2\sigma_0\bU^\top(\bI - \bH_{k}-\bO_{ k^{\bot}  \bu^*})\bX_{k_0}\bbeta_{k_0}^0 \\ \nonumber
    &\hspace{3.4em} +\lambda(k-k_0) ) < 0\} \\ \label{I_1 and I_2}
    &= (I_1) + (I_2).
   \end{align}

Following Lemma 4 of \cite{shen2013constrained}, we can bound the log of the moment generating function $M(t)$ of $\bU^{T}(\bH_{k,\bu^*} - \bH_{k_0,\bu^*})\bU$ as 
\begin{align} \nonumber
\log\{M(t)\} &= \sum_{r=1}^{\infty} \frac{2^{r-1}t^r}{r} \tr\{(\bH_{k,\bu^*} - \bH_{k_0,\bu^*})^r\} \\ \nonumber
&\leq t\left(d(k) - d(k_0)\right) + \frac{t^2}{1-2t}\tr\{(\bH_{k,\bu^*} - \bH_{k_0,\bu^*})^2\} \\ \nonumber
&\leq t\left(d(k) - d(k_0)\right) +  \frac{t^2}{1-2t}\left(d(k) + d(k_0)+2\right) \\ 
\label{eq:log_mgf}
&\leq t\left(d(k) +2\right)
\end{align}
for any $0<t<1/3.$
By Lemma \ref{lem_proj}, we have that 
$$
\bH_{k}+\bO_{ k^{\bot}  \bu^*} -  \bH_{k_0}-\bO_{ k_0^{\bot} \bu^*} = \bH_{k}+\bO_{ k^{\bot}  \bu^*} - \bI -  \bH_{k_0}-\bO_{ k_0^{\bot} \bu^*} + \bI = \bH_{k,\bu^*} - \bH_{k_0,\bu^*}.
$$ 
Thus, with direct calculations and an application of the Markov inequality, we can deduce that 
\begin{align} \nonumber
    & (I_1)
     = \Prob_{\bU}\Big\{(1-\gamma^2_1)(1-\delta)\|(\bI-\bH_k)\bX_{k_0}\bbeta_{k_0}^0\|^2 + \lambda(k-k_0) \\ \nonumber 
    & \qquad \qquad\qquad \qquad \qquad <\sigma_0^2\bU^\top( \bH_{k}+\bO_{ k^{\bot}  \bu^*} -  \bH_{k_0}-\bO_{ k_0^{\bot} \bu^*} )\bU\Big\} \\ \nonumber
    & = \Prob_{\bU} \Bigg[\exp\left\{ t_1 (\bU^{T}(\bH_{k,\bu^*} - \bH_{k_0,\bu^*})\bU)\right\} \\ \nonumber
    & \qquad \qquad\qquad \qquad \qquad >\exp\Bigg\{ t_1 \frac{(1-\gamma^2_1)(1-\delta)\|(\bI-\bH_k)\bX_{k_0}\bbeta_{k_0}^0\|^2 + \lambda(k-k_0)}{\sigma^2_0}\Bigg\} \Bigg] \\ \label{I_1_lem8}
    & \leq \exp\left\{t_1\left( d(k) + 2\right) - \frac{t_1(1-\delta)(1-\gamma^2_1)n_0 (k_0 - k) C_{\min} + t_1 \lambda (k - k_0)}{\sigma_0^2} \right\}
\end{align} 
for any $0<t_1<1/3$, where 
the last inequality above also uses \eqref{eq:log_mgf} and the definition of $C_{\min}$.

Further, since $2\sigma_0\bU^\top(\bI - \bH_{k,\bu^*})\bX_{k_0}\bbeta_{k_0}^0$ follows the normal distribution $N(0, \sigma^2_0\|(\bI - \bH_{k,\bu^*})\bX_{k_0}\bbeta_{k_0}^0\|^2)$, it follows from the Markov inequality, the moment generating function of the normal distribution, and the definition of $C_{\min}$ that for any $0<t_2<1/3$, 
\allowdisplaybreaks
\begin{align}  \nonumber
  & (I_2) = \Prob_{\bU} \Bigg[ \exp\Big\{  t_2 \frac{2(1 - \gamma^2_1)\bU^\top(\bH_{ k,  \bu^*} - \bI)\bX_{k_0}\bbeta_{k_0}^0 }{\sigma_0}\Big\} \\ \nonumber
  &\hspace{5em} > \exp\Big\{t_2\frac{(1-\gamma^2_1)^2\delta\|(\bI-\bH_k)\bX_{k_0}\bbeta_{k_0}^0\|^2 + \lambda(k - k_0)}{ \sigma^2_0} \Big\} \Bigg] \\ \nonumber
  & \leq \mathbb{E}\Bigg[ \exp\Big\{  t_2 \frac{2(1 - \gamma^2_1)\bU^\top(\bH_{ k,  \bu^*} - \bI)\bX_{k_0}\bbeta_{k_0}^0 }{\sigma_0}\Big\} \Bigg]  \\ \nonumber
  & \hspace{5em} \times \exp\left\{ -t_2 \frac{(1-\gamma^2_1)^2\delta\|(\bI-\bH_k)\bX_{k_0}\bbeta_{k_0}^0\|^2 + \lambda(k - k_0)}{ \sigma^2_0} \right\} \\ \nonumber
  & = \exp\left\{ \frac{4 t^2_2(1 - \gamma^2_1)^2 \|(\bI - \bH_{k,\bu^*})\bX_{k_0}\bbeta_{k_0}^0\|^2}{2 \sigma^2_0}\right\} \\ \nonumber
  & \hspace{5em} \times\exp\left\{-t_2\frac{(1-\gamma^2_1)^2\delta\|(\bI-\bH_k)\bX_{k_0}\bbeta_{k_0}^0\|^2 + \lambda(k - k_0)}{ \sigma^2_0} \right\}  \\ \nonumber
   & \leq \exp\left\{ \frac{(2t^2_2 -\delta t_2) (1 - \gamma^2_1)^2}{\sigma^2_0} \|(\bI-\bH_k)\bX_{k_0}\bbeta_{k_0}^0\|^2 \right\} \exp\left\{ -t_2 \frac{\lambda(k-k_0)}{\sigma^2_0}\right\} \\ \label{I_2_lem8}
  & \leq \exp\left\{\frac{(2t_2^2- \delta t_2) (1-\gamma_1^2)^2 n_0 (k_0 - k) C_{\min} - t_2 \lambda (k- k_0)}{\sigma_0^2}\right\}.
\end{align} 

Then, from\eqref{equ:: union expression}, \eqref{I_1 and I_2}, \eqref{I_1_lem8} and \eqref{I_2_lem8}, by dividing the range of $k$ into two parts using $k_0$, we can show that
\begin{align*}
   & \Prob_{\bU}(\hat k_{\bu^*} \neq k_0)  \\
   & \leq\Prob_{\bU} \Bigg\{\bigcup_{k} \{D(k,\bu^*) - D(k_0,\bu^*) <0\} \Bigg\}  \\
   & \leq \sum_{k} \Prob_{\bU} \Bigg\{\{D(k,\bu^*) - D(k_0,\bu^*) <0\} \Bigg\}  \\
   & \leq  \sum_{k<k_0}  \Bigg[ \exp\left\{t_1\left( d(k) + 2\right) - \frac{t_1(1-\delta)(1-\gamma^2_1)n_0 (k_0 - k) C_{\min} + t_1 \lambda (k - k_0)}{\sigma_0^2} \right\}   \\
    & \hspace{4em}    + \exp\left\{\frac{(2t_2^2- \delta t_2) (1-\gamma_1^2)^2 n_0 (k_0 - k) C_{\min} - t_2 \lambda (k- k_0)}{\sigma_0^2}\right\} \Bigg]\\
    &  + \sum_{k>k_0}  \Bigg[ \exp\left\{t_1\left( d(k) + 2\right) - \frac{t_1(1-\delta)(1-\gamma^2_1)n_0 (k_0 - k) C_{\min} + t_1 \lambda (k - k_0)}{\sigma_0^2} \right\}  \\
    & \hspace{4em} + \exp\left\{\frac{(2t_2^2- \delta t_2) (1-\gamma_1^2)^2 n_0 (k_0 - k) C_{\min} - t_2 \lambda (k- k_0)}{\sigma_0^2}\right\} \Bigg].
\end{align*}

Let us set $\delta=5/6$ and $t_1=t_2=1/3$ in the last step above. Then when $$\frac{\lambda}{n_0} \in \left[ \sigma^2_0(1 + \frac{2}{n_0}) + m_1, \min \left \{ \frac{(1-\gamma^2_1)^2C_{\min}}{6} - \frac{3\sigma^2_0 \ln2 }{n_0}, \frac{(1-\gamma^2_1)^2C_{\min}}{6} - \frac{\sigma^2_0 d(k_0) }{n_0}\right\} - m_2\right],$$
we can show by some direct calculations that
\begin{align}  
\Prob_{\bU}(\hat k_{\bu^*} \neq k_0) \label{equ::bound_of_single_k_0}
  \leq 4 \exp \left\{ - \frac{n_0 m_1}{3 \sigma^2_0}   \right\} +  4 \exp \left\{\frac{2}{3}  - \frac{n_0 m_2}{18 \sigma^2_0}\right\}.
\end{align}
By the probability union bound, it holds that 
\begin{align*}
    \Prob_{({\cal U}^d, \bY)} (k_0 \notin \mathcal S_B )  \leq&~ \Prob_{({\cal U}^d, \bY)} \left\{k_0 \notin \mathcal S_B , \min_{1\leq i \leq d} \max_{ k< k_0} \rho_{k^\bot}(\bU^*_i, \bX_{k_0}\bbeta_{k_0}^0) < \gamma_1^2\right\}  \\ \nonumber 
    &  + \Prob_{({\cal U}^d, \bY)} \left\{k_0 \notin \mathcal S_B, \min_{1\leq i \leq B} \max_{k < k_0} \rho_{k^\bot}(\bU^*_i, \bX_{k_0}\bbeta_{k_0}^0) \geq \gamma_1^2\right\}.
\end{align*}

Denote by $i_{\min} = \arg\min_{1\leq i \leq B} \max_{k < k_0} \rho_{k^\bot}(\bU^*_i, \bX_{k_0}\bbeta_{k_0}^0)$. To bound the first term above, notice that $k_0 \notin \mathcal S_B $ implies that $k_0$ is not equal to any element in $\mathcal S_B$. From the definition of $i_{\min}$ and the conditional probability, we can deduce that 
\begin{align} 
  \nonumber  & \Prob_{({\cal U}^d, \bY)} \left\{k_0 \notin \mathcal S_B , \min_{1\leq i \leq d} \max_{ k< k_0} \rho_{k^\bot}(\bU^*_i, \bX_{k_0}\bbeta_{k_0}^0) < \gamma_1^2\right\} \\ \nonumber
    \leq &~ \Prob_{({\cal U}^d, \bY)} \left\{\hat k_{\bU^*_{i_{\min}}} \neq k_0 ,  \max_{k < k_0} \rho_{k^\bot}(\bU^*_{i_{\min}}, \bX_{k_0}\bbeta_{k_0}^0) < \gamma_1^2\right\} \\ 
    \leq &~ \Prob_{( \bU|{\cal U}^d)} \left\{\hat k_{\bU^*_{i_{\min}}} \neq k_0 \middle| \max_{k < k_0} \rho_{k^\bot}(\bU^*_{i_{\min}}, \bX_{k_0}\bbeta_{k_0}^0) < \gamma_1^2\right\}.\label{equ::first_bound_of_notinS_B}
\end{align}
We next bound the second term. It follows from the basic probability inequality that 
\begin{align} \nonumber
    & \Prob_{({\cal U}^d, \bY)} \left\{k_0 \notin \mathcal S_B, \min_{1\leq i \leq B} \max_{k < k_0} \rho_{k^\bot}(\bU^*_i, \bX_{k_0}\bbeta_{k_0}^0) \geq \gamma_1^2\right\} \\ \nonumber
    \leq &~ \Prob_{({\cal U}^d, \bY)} \left\{ \min_{1\leq i \leq B} \max_{ , k < k_0} \rho_{k^\bot}(\bU^*_i, \bX_{k_0}\bbeta_{k_0}^0) \geq \gamma_1^2\right\} \\ \label{equ::second_bound_of_notinS_B}
    \leq &~ \Prob_{({\cal U}^d, \bY)} \left\{ \bigcap_{1\leq i \leq B} \left\{\max_{ k < k_0} \rho_{k^\bot}(\bU^*_i, \bX_{k_0}\bbeta_{k_0}^0) \geq \gamma_1^2 \right\}\right\}.
\end{align}

Then combining \eqref{equ::first_bound_of_notinS_B} and \eqref{equ::second_bound_of_notinS_B} leads to 
\begin{align} \nonumber
     \Prob_{({\cal U}^d, \bY)} (k_0 \notin \mathcal S_B ) \leq&~  \Prob_{( \bU|{\cal U}^d)} \left\{\hat k_{\bU^*_{i_{\min}}} \neq k_0 \middle| \max_{k < k_0} \rho_{k^\bot}(\bU^*_{i_{\min}}, \bX_{k_0}\bbeta_{k_0}^0) < \gamma_1^2\right\} \\ \nonumber
    & \qquad \times \Prob_{{\cal U}^d} \left\{\max_{k < k_0} \rho_{k^\bot}(\bU^*_{i_{\min}}, \bX_{k_0}\bbeta_{k_0}^0) < \gamma_1^2\right\} \\ \label{equ::bound_first_part}
    &  + \Prob_{({\cal U}^d, \bY)} \left\{ \bigcap_{1\leq i \leq B}\left\{ \max_{k < k_0} \rho_{k^\bot}(\bU^*_i, \bX_{k_0}\bbeta_{k_0}^0) \geq \gamma_1^2 \right\} \right\}.
\end{align}
Therefore, from \eqref{equ::bound_first_part} we can deduce that 
\begin{align} \nonumber
    & \Prob_{({\cal U}^d, \bY)} (k_0 \notin \mathcal S_B ) \leq  \max\left\{ \Prob_{\bU}(\hat k_{\bU^*} \neq k_0) : \max_{k < k_0} \rho_{k^\bot}(\bU^*, \bX_{k_0}\bbeta_{k_0}^0) < \gamma_1^2 \right\} \\ \nonumber
    & \hspace{9em} + \prod^B_{i=1}  \Prob_{\bU^*_i} \left\{  \max_{k < k_0} \rho_{k^\bot}(\bU^*_i, \bX_{k_0}\bbeta_{k_0}^0) \geq \gamma_1^2\right\}\\ \label{equ::bounding k0 notin_SB}
    \leq&~  4 \exp \left\{ - \frac{n_0 m_1}{3 \sigma^2_0}   \right\} +  4 \exp \left\{\frac{2}{3}  - \frac{n_0 m_2}{18 \sigma^2_0}\right\} + \left[2 k_0\{\arccos (\gamma_1)\}^{n_0-d(k_0)-1 }\right]^B,
\end{align}
where the first inequality above is due to the independence of the $\bU^*_i$'s, and
the second inequality comes from \eqref{equ::bound_of_single_k_0} and Lemma \ref{lem::angle}. This concludes the proof of Lemma \ref{lem::symptotic_bound_penalize0}.

\subsection{Proof of Lemma \ref{lemma:u_d}} \label{sec.supp.B5}
Denote by 
$$g_k(\bU)= \frac{\|(\bI-\bH_k)\bU\|}{\|\bU\|} \ \text{ and } \ g_{k}(\bU^*)= \frac{\|(\bI-\bH_k)\bU^*\|}{\|\bU^*\|}.$$ 
For each given $\bU^*$, by adding and subtracting the term $\left((\bU^*)^\top/\|\bU^*\|-\bU^\top/\|\bU\|\right)(\bI-\bH_k)\bX_{k_0} \bbeta_{k_0}^0$ to $\frac{1}{\|(\bI-\bH_k)\bU^*\|}\left((\bU^*)^\top(\bI-\bH_k)\bX_{k_0} \bbeta_{k_0}^0\right)$, and applying the Cauchy--Schwarz inequality, we can deduce that 
\begin{align} \nonumber
     & \frac{1}{\|(\bI-\bH_k)\bU^*\|}\left((\bU^*)^\top(\bI-\bH_k)\bX_{k_0} \bbeta_{k_0}^0\right)\\  \nonumber
     & =   \frac{1}{\|\bU\|} \bU^\top(\bI-\bH_k)\bX_{k_0} \bbeta_{k_0}^0
     + \left((\bU^*)^\top/\|\bU^*\|-\bU^\top/\|\bU\|\right)(\bI-\bH_k)\bX_{k_0} \bbeta_{k_0}^0  \\ \nonumber & \quad + \left(\frac{1}{\|(\bI-\bH_k)\bU^*\|}
      -\frac{1}{\|\bU^*\|}\right)(\bU^*)^\top(\bI-\bH_k)\bX_{k_0} \bbeta_{k_0}^0\\ \label{equ::original_term_10}
     & \leq (L_1),
\end{align}
where we define 
\begin{align*}
     & (L_1) :=  \frac{\|\bI-\bH_k)\bU\|}{\|\bU\|}\frac{1}{\|(\bI-\bH_k)\bU\|}\bU^\top(\bI-\bH_k)\bX_{k_0} \bbeta_{k_0}^0 \\
     &   \quad + \left\|\frac{(\bU^*)^\top}{\|\bU^*\|}-\frac{\bU^\top}{\|\bU\|}\right\|\|(\bI-\bH_k)\bX_{k_0} \bbeta_{k_0}^0\| \nonumber \displaybreak[1]\\
     &   \quad  + \frac{\|\bU^*\|-\|(\bI-\bH_k)\bU^*\|}{\|\bU^*\|}\frac{1}{\|(\bI-\bH_k)\bU^*\|}(\bU^*)^\top(\bI-\bH_k)\bX_{k_0} \bbeta_{k_0}^0.
\end{align*}

From the definitions of $g_k(\bU)$ and $g_k(\bU^*)$, it holds that 
\begin{align} \nonumber
     (L_1)   \leq&~ g_{k}(\bU)\tilde\gamma_1\|(\bI-\bH_k)\bX_{k_0} \bbeta_{k_0}^0\|  +  \sqrt{2-2\frac{(\bU^*)^\top \bU}{\|(\bU^*)^\top\| \|\bU\|}}\|(\bI-\bH_k)\bX_{k_0} \bbeta_{k_0}^0\| \\ \nonumber
     &+ (1-g_{k}(\bU^*))\frac{1}{\|(\bI-\bH_k)\bU^*\|}(\bU^*)^\top(\bI-\bH_k)\bX_{k_0} \bbeta_{k_0}^0\\ \nonumber
     \leq&~ g_{k}(\bU) \tilde\gamma_1\|(\bI-\bH_k)\bX_{k_0} \bbeta_{k_0}^0\|  +  \sqrt{2-2\sqrt{1-\gamma^2_2}}\|(\bI-\bH_k)\bX_{k_0} \bbeta_{k_0}^0\|\\ \label{equ::bound_L_1}
     &+ (1-g_{k}(\bU^*))\frac{1}{\|(\bI-\bH_k)\bU^*\|}(\bU^*)^\top(\bI-\bH_k)\bX_{k_0} \bbeta_{k_0}^0,
\end{align}
where the first inequality above uses the condition $\max_{k<k_0} \rho_{k^\bot}(\bU, \bX_{k_0} \bbeta_{k_0}^0) < \tilde\gamma_1^2$ in \eqref{eq:def_subset_E} to incorporate $\tilde\gamma_1$ as an upper bound, and the second inequality uses the condition of $\rho(\bU^*, \bU)$ in \eqref{eq:def_subset_E}.

Then it follows from \eqref{equ::original_term_10} and \eqref{equ::bound_L_1} that
\begin{align}  \nonumber
    & \frac{1}{\|(\bI-\bH_k)\bU^*\|}(\bU^*)^\top(\bI-\bH_k)\bX_{k_0} \bbeta_{k_0}^0\\ 
    \label{equ::bound_U_Xbeta}
    & \leq  \frac{g_{k}(\bU)}{g_{k}(\bU^*)} \tilde\gamma_1\|(\bI-\bH_k)\bX_{k_0} \bbeta_{k_0}^0\|  + \frac{1}{g_{k}(\bU^*)} \sqrt{2-2\sqrt{1-\gamma^2_2}}\|(\bI-\bH_k)\bX_{k_0} \bbeta_{k_0}^0\|.
\end{align}
Further, let us define $\bH_{\bU}$ as the projection operator that projects any vector onto the one-dimensional subspace spanned by vector $\bU$. Since $\|(\bI - \bH_k)\bU^*\|  \leq \|(\bI - \bH_k \bH_{\bU})\bU^*\|$, if $(\bU^*, \bU)  \in \tilde{E}(\gamma_1,\gamma_2)$ it holds that 
\begin{align*}
    g_{k}(\bU^*) & \leq \frac{\|(\bI-\bH_{\bU})\bU^*\|}{\|\bU^*\|} + \frac{\|(\bH_{\bU}-\bH_k \bH_{\bU})\bU^*\|}{\|\bU^*\|} \\ &\leq \gamma_2 + \frac{\|(\bI-\bH_k)\bH_{\bU}\bU^*\|}{\|\bH_{\bU}\bU^*\|} \\
    &= \gamma_2 + g_{k}(\bU).
\end{align*}

Similarly, by the symmetry between $\bU$ and $\bU^*$, we can show that 
$$g_{k}(\bU)\leq  g_{k}(\bU^*) + \gamma_2.$$ Hence, by dividing both sides by $g_k(\bU)$, we can derive that 
\begin{align*}
\frac{g_{k}(\bU^*)}{g_{k}(\bU)}\geq 1 - \frac{\gamma_2}{g_{k}(\bU)}.     
\end{align*}
Now by substituting \eqref{equ::bound_tilde_gamma_1} into \eqref{equ::bound_U_Xbeta},  we can see that a sufficient condition for 
\begin{align}
    \label{equ::condition_gamma_1_inE}
    \frac{1}{\|(\bI-\bH_k)\bU^*\|}(\bU^*)^\top(\bI-\bH_k)\bX_{k_0} \bbeta_{k_0}^0 \leq \gamma_1
\end{align}
 is given by 
\begin{align}
\label{equ::bound_tilde_gamma_1}
\tilde\gamma_1\leq \left(1 - \frac{\gamma_2}{g_{k}(\bU)}\right) \gamma_1 - \sqrt{2-2\sqrt{1-\gamma^2_2}}\leq  \frac{g_{k}(\bU^*)}{g_{k}(\bU)}\gamma_1 - \sqrt{2-2\sqrt{1-\gamma^2_2}}. 
\end{align}

Observe that it follows from 
\begin{align}
    \rho(\bU, k) < 1- \gamma_2
    \label{equ::condition_rho}
\end{align}
that 
\begin{align*}
    g_{k}(\bU) = \sqrt{1-\rho^2(\bU, k) } > \sqrt{\gamma_2}.
\end{align*}
Then we can see that the sufficient condition introduced above is satisfied for the choice of  $$\tilde\gamma_1=(1-\sqrt{\gamma_2})\gamma_1 - \sqrt{\frac{2-2\sqrt{1-\gamma^2_2}}{\gamma_2}}.$$ 
Therefore, noticing that to derive \eqref{equ::condition_gamma_1_inE} we have applied conditions \eqref{equ::bound_tilde_gamma_1} and \eqref{equ::condition_rho},
we can obtain that
\begin{align*}
  & \{\rho(\bU^*, \bU)  > 1- \gamma_2^2\}\hspace{-0.5mm}\bigcap \hspace{-0.5mm}\bigg\{ \max_{k <k_0} \rho_{k^\bot}(\bU^*, \bX_{k_0} \bbeta_{k_0}^0) < \tilde\gamma_1^2, \max_{k<k_0} \rho(\bU, k) < 1- \gamma_2 \bigg\}\\
  & \quad \subset  E(\gamma_1, \gamma_2), 
\end{align*}
which completes the proof of Lemma \ref{lemma:u_d}.

\subsection{Proof of Lemma \ref{lem:epsilon_bound}} \label{sec.supp.B6}

First, note that the desired probability can be upper bounded as
\begin{align}\label{eq:014}
    & \mathbb{P}\left(\bU \not\in  \left\{\max_{k<k_0} \rho_{k^\bot}(\bU, \bX_{k_0} \bbeta_{k_0}^0) < \tilde\gamma_1^2,  \max_{k < k_0} \rho(\bU, k) < 1- \gamma_2  \right\}\right) \nonumber\\
    & \leq \mathbb{P}\left(\max_{k<k_0} \rho(\bU, k) \geq 1- \gamma_2 \right) + \mathbb{P}\left(\max_{k<k_0} \rho_{k^\bot}(\bU, \bX_{k_0} \bbeta_{k_0}^0) \geq \tilde\gamma_1^2\right)\nonumber\\ 
    & \leq \sum_{k<k_0}\mathbb{P}\left( \rho(\bU, k) \geq 1- \gamma_2 \right) + \sum_{k<k_0}\mathbb{P}\left( \rho_{k^\bot}(\bU, \bX_{k_0} \bbeta_{k_0}^0) \geq \tilde\gamma_1^2\right),
\end{align} 
where the first and second inequalities come from the property of the union bound. Hence, we need only to bound the two probabilities on the very right-hand side above.

Let us define 
$$g_{k}^2(\bu)=\frac{\|(\bI-\bH_k)\bu\|^2}{\|\bu\|^2} = \frac{\|(\bI-\bH_k)\bu\|^2}{\|(\bI-\bH_k)\bu\|^2 + \|\bH_k \bu\|^2},$$ 
and let $F_{a,b}$ be the CDF of the $F\text{-distribution}$ with degrees of freedom $a$ and $b$. When $n_0-d(k)>4$ and $\gamma_2<0.6$, it follows from the definitions of $\rho(\bU, k)$ and $g_k(\bU)$ that 
\begin{align} \nonumber
   \mathbb{P}_{\bU}\{\rho(\bU, k) \geq 1- \gamma_2 \}&=\mathbb{P}_{\bU}\{g_{k}(\bU) \leq \sqrt{\gamma_2}\}\\ \nonumber
   & = \mathbb{P}_{\bU}\left(\frac{\|(\bI-\bH_k)\bU\|^2}{\|\bH_{k}\bU\|^2} \leq \frac{\gamma_2}{1-\gamma_2}\right)\\ \nonumber
   &  =F_{n_0-d(k), d(k)}\left(\frac{\gamma_2/(n_0-d(k))}{(1-\gamma_2)/d(k)}\right) \\ 
   \label{equ::bound of rho}
   &\leq
   \left(\frac{n_0-d(k)}{2}\right)^{\frac{d(k)}{2}}\gamma_2^{\frac{n_0-d(k)}{2}-1}, 
\end{align}
where the third equality above holds because both $\|(\bI-\bH_k)\bU\|^2$ and $\|(\bI-\bH_k)\bU\|^2$ are chi-square distributed with degrees of freedom $n_0 - d(k)$ and $d(k)$, respectively, and the last inequality uses the fact that
\begin{align*}
    F_{n_0-d(k), d(k)}(\frac{\gamma_2/(n_0-d(k))}{(1-\gamma_2)/d(k)}) & = \frac{{\int}_0^{\gamma_2} t^{\frac{n_0-d(k)}{2}-1} (1-t)^{\frac{d(k)}{2}-1}dt}{\mathcal B(\frac{n_0-d(k)}{2}, \frac{d(k)}{2})} \leq \frac{\gamma_2^{\frac{n_0-d(k)}{2}-1}}{\mathcal B(\frac{n_0-d(k)}{2}, \frac{d(k)}{2})},
\end{align*}
with $\mathcal B(\frac{n_0-d(k)}{2}, \frac{d(k)}{2})$ the beta function satisfying that
\begin{align*}
    \mathcal B(\frac{n_0-d(k)}{2}, \frac{d(k)}{2}) \geq \left(\frac{n_0-d(k)}{2}\right)^{-\frac{d(k)}{2}}.
\end{align*}

Therefore, combining \eqref{eq:014}--\eqref{equ::bound of rho} with Lemmas \ref{lem::angle} and \ref{lemma:u_d} yields that 
\begin{align*}
    & \mathbb{P}\left(\bU \not\in  \left\{\max_{k<k_0} \rho_{k^\bot}(\bU, \bX_{k_0} \bbeta_{k_0}^0) < \tilde\gamma_1^2,  \max_{k < k_0} \rho(\bU, k) < 1- \gamma_2  \right\}\right)\\
    & \leq \sum_{k<k_0}\Bigg\{2(\arccos \tilde\gamma_1)^{n_0-d(k)-1} +   \left(\frac{n_0-d(k)}{2}\right)^{\frac{d(k)}{2}}\gamma_2^{\frac{n_0-d(k)}{2}-1}\Bigg\} \\
      & \leq   2(\arccos \tilde\gamma_1)^{n_0-1} \sum_{k<k_0} \frac{1}{(\arccos \tilde{\gamma}_1)^{d(k)}}  + \sum_{k<k_0} (\frac{n_0-1}{2})^{\frac{d(k_0)}{2}} \gamma^{n_0-d(k)-1}_2 \\
      & \leq   2(\arccos \tilde\gamma_1)^{n_0-1} \sum_{k<k_0} \frac{1}{(\arccos \tilde{\gamma}_1)^{d(k)}}  + k_0 (\frac{n_0-1}{2})^{\frac{d(k_0)}{2}} \gamma^{n_0-2}_2.
\end{align*}
This concludes the proof of Lemma \ref{lem:epsilon_bound}.

\section{Additional technical details and simulation results} \label{sec.supp.C}

\subsection{Additional technical details} \label{sec.supp.C1}

\textit{The ADMM algorithm for updating $\bbeta$}. We now provide detailed derivations for the alternating direction method of multipliers (ADMM) algorithm implementing the square-root fused clipped Lasso (SFL) method for inferring the values of interference functions. Observe that the SFL introduced in \eqref{equation::fused-lasso} for our setting involves a nonconvex optimization problem. To solve such a nonconvex problem, we employ the difference-of-convex (DC) programming \citep{LeThi1997SolvingAC} which helps deal with the discontinuities due to the use of the indicator functions in our SFL formulation. Specifically, the DC decomposition refers to expressing a nonconvex function as the difference of two convex functions, which enables solving nonconvex optimization problems using methods designed for convex problems. 
For our SFL formulation, we solve the nonconvex problem by iteratively solving sub-optimization problems, where in each iteration we break the task down to a convex problem that can be implemented with the ADMM algorithm \citep{boyd2011distributed}. 

Our goal is to find solution $\hat \bbeta^{grp}$ by minimizing the SFL objective function
\begin{align}
    \underset{\bbeta_{k} \in \mathbb R^{d(k)}}{\rm argmin}
    \left\{ \sqrt{(2n_0)^{-1}\|\by_{obs}- \bX_{k}\bbeta_{k}\|^2_2} + \lambda_1 J(\bbeta_{k}) \right\},
    \label{equation::fused-lasso-copy}
\end{align}
where
\begin{align*}
    J(\bbeta_{k})
    =
    \sum_{1 \leq i<j \leq d(k)} \Big\{ &~  |\beta_{k,i}-\beta_{k,j}|\mathbbm{1}(|\beta_{k,i}-\beta_{k,j}|<\lambda_2) + \\
    &~ \lambda_2 \mathbbm{1}(|\beta_{k,i}-\beta_{k,j}| \geq \lambda_2) \Big\}.
\end{align*}
Denote by 
$$\mathcal{S}(\bbeta) = \sqrt{(2n_0)^{-1}\|\by_{obs}- \bX_{k}\bbeta_{k}\|^2_2} + \lambda_1 J(\bbeta_{k}).$$
Using the idea of the DC decomposition, we can decompose $\mathcal{S}(\bbeta)$ above into the difference of two convex functions, $\mathcal{S}_1(\bbeta)$ and $\mathcal{S}_2(\bbeta)$; that is, 
$$\mathcal{S}(\bbeta) = \mathcal{S}_1(\bbeta) - \mathcal{S}_2(\bbeta). $$ 
Specifically, we define 
\begin{align}
    \mathcal{S}_1(\bbeta) = \sqrt{(2n_0)^{-1}\|\by_{obs}- \bX_{k}\bbeta_{k}\|^2_2} + \lambda_1 \sum_{1 \leq i < j \leq d(k)}|\beta_{k,i}-\beta_{k,j}|
\end{align} 
and 
\begin{align}
\mathcal{S}_2(\bbeta) = \lambda_1 \sum_{1 \leq i < j \leq d(k)} (|\beta_{k,i}-\beta_{k,j}| - \lambda_2)_{+},
\end{align}
where $(z)_+$ stands for the positive part of a given scalar. 

We then apply the DC programming to solve the SFL optimization problem \eqref{equation::fused-lasso-copy} above through iteratively solving the subproblem 
\begin{align}
    \nonumber
    \hat{\bbeta}^{s+1}_{k} = \underset{\bbeta_{k} \in \mathbb R^{d(k)}}{\rm argmin}  &~ \Bigg\{ \sqrt{(2n_0)^{-1}\|\by_{obs}- \bX_{k}\bbeta_{k}\|^2_2} \, + \\
    &~  \lambda_1 \sum_{1 \leq i < j \leq d(k)}|\beta_{k,i}-\beta_{k,j}| -  \langle \nabla{\mathcal{S}_2(\hat\bbeta^s_{k})}, \bbeta_{k} \rangle \Bigg\}, 
    \label{sub-optimization-problem: ADMM}
\end{align}
where $\nabla{\mathcal{S}_2(\bbeta^s)}$ denotes the subgradient with respect to $\bbeta^s$ satisfying
\begin{align*}
    \left[\nabla \mathcal{S}_2(\boldsymbol{\beta}^s)\right]_i = \lambda_1 \sum_{j:\, j \neq i}\begin{cases}
1 & \text{if } \beta^{s}_{k,i)}-\beta^{s}_{k,j} > \lambda_2, \\
-1 &  \text{if } \beta^{s}_{k,i)}-\beta^{s}_{k,j} < -\lambda_2,
\end{cases}
\end{align*}
and $s$ stands for the $s$th iteration. We see immediately that the new optimization problem (\ref{sub-optimization-problem: ADMM}) above is indeed a convex one since the last part $\langle \nabla{\mathcal{S}_2(\bbeta^s)}, \bbeta \rangle $ is linear in $\bbeta$. Hence, we can resort to the ADMM algorithm to solve the convex subproblem in \eqref{sub-optimization-problem: ADMM}, which is summarized in Algorithm \ref{alg:admm}.

\begin{algorithm}[ht!]
\caption{ADMM algorithm for solving convex subproblem in \eqref{sub-optimization-problem: ADMM}}
\label{alg:admm}
    \KwIn{Initialization $\bp^0$, $\bnu^0$, total iteration number $S'$, $\hat\bbeta^s_{k}$ from DC programming iteration, and $\lambda_1,\rho>0$.}

    \For{$s' \in \{0,1,\ldots,S'-1\}$}{

    Update $\bbeta^{s'+1}$ by solving 
    \begin{align*}
        \bbeta^{s'+1} = &~\underset{\bbeta_{k} \in \mathbb R^{d(k)}}{\rm argmin} \Big\{ \sqrt{(2n_0)^{-1}\|\by_{obs}- \bX_{k}\bbeta_{k}\|^2_2} - \langle \nabla{\mathcal{S}_2(\hat\bbeta^s_{k})}, \bbeta_{k} \rangle  + \\
        &~ \quad \langle \bT^T \bnu^{s'}, \bbeta_{k} \rangle + \frac{\rho}{2} \|\bT \bbeta_{k} - \bp^{s'} \|^2 \Big\}, 
    \end{align*}
    where $\bT$ satisfies $\bT \bbeta_{k} = \sum_{1 \leq i < j \leq d(k)} |\beta_{k,i}-\beta_{k,j}|$  as defined in \eqref{equ::def-T};
    
    Update $\bp^{s'+1}$ by computing
    $$
    \bp^{s'+1} = \mathcal{L}_{\frac{\lambda_1}{\rho}}(\bT \bbeta^{s'+1} + \frac{1}{\rho} \bnu^{s'}),
    $$
    where $\mathcal{L}_{\lambda}(\cdot)$ is the soft-thresholding operator applied componentwise with $\mathcal{L}_{\lambda}(x) = \operatorname{sign}(x) (|x|-\lambda)_{+}$\;
    
    Update the dual variable $\bnu$ as 
    $
    \bnu^{s'+1} = \bnu^{s'} + \rho(\bT \bbeta^{s'+1} - \bp^{s'+1});
    $
    
    }

    \KwOut{$\bbeta^{S'}$ as the solution of \eqref{sub-optimization-problem: ADMM}.}
\end{algorithm}

We next provide additional technical details for the ADMM algorithm given in Algorithm \ref{alg:admm}.
First, we can derive the augmented Lagrangian for the convex optimization problem \eqref{sub-optimization-problem: ADMM} as 
\begin{align*}
 \bL_{\rho}(\bbeta_{k},\bp,\gamma)
    & = \sqrt{(2n_0)^{-1}\|\by_{obs}- \bX_{k}\bbeta_{k}\|^2_2}  \\
    & \quad- \langle \nabla{\mathcal{S}_2(\hat{\bbeta}^s_{k})}, \bbeta_{k} \rangle  +\lambda_1\| \bp\|_1 \\
    & \quad+\langle \bnu, \bT \bbeta_{k} - \bp \rangle +\frac{\rho}{2} \|\bT \bbeta_{k} - \bp \|^2,
\end{align*}
where $\nabla{\mathcal{S}_2(\hat{\bbeta}^s_{k})}$ is a deterministic vector carried over from the previous DC programming iteration and 
\begin{align}
    \bT =
    \begin{bmatrix}
    1 & -1 & 0 & \cdots & 0 & 0 \\
    1 & 0 & -1 & \cdots & 0 & 0 \\
    \vdots & \vdots & \vdots & \ddots & \vdots & \vdots \\
    1 & 0 & 0 & \cdots & 0 & -1 \\
    0 & 1 & -1 & \cdots & 0 & 0 \\
    \vdots & \vdots & \vdots & \ddots & \vdots & \vdots \\
    0 & 0 & 0 & \cdots & 1 & -1
    \end{bmatrix}
    \in \mathbb{R}^{\binom{d(k)}{2} \times d(k)}
    \label{equ::def-T}
\end{align}
satisfying $\bT \bbeta_{k} = \sum_{1 \leq i < j \leq d(k)} |\beta_{k,i}-\beta_{k,j}|.$

To update $\bbeta$, we invoke the Karush--Kuhn--Tucker (KKT) conditions by setting $$\frac{\partial \bL_p}{\partial \bbeta_{k}} = 0$$ with 
\begin{align*}
    \frac{\partial \bL_p}{\partial \bbeta_{k}}
    = \frac{\bX_{k} ^{T} (\bX_{k}  \bbeta_{k}  - \by_{obs})}{\sqrt{2 n_0} \| \by_{obs} - \bX_{k} \bbeta_{k}  \|_2} - \nabla{\mathcal{S}_2(\hat{\bbeta}^s)} + \bT^{T} \bnu + \rho \bT^{T} (\bT \bbeta_{k}  - \bp).
\end{align*}
Notice that the expression above is nonlinear with respect to $\bbeta_{k}$. We numerically solve it and obtain the optimizer $\bbeta^{s'+1}$ in  Algorithm \ref{alg:admm}.
We further update $\bp$ by finding the minimizer of
\begin{align}
    \lambda_1 \| \bp\|_1 - \bnu^T \bp + \frac{\rho}{2} \| \bT \bbeta_{k}- \bp\|^2.
    \label{equ::update-p}
\end{align}
As shown in Section 2.1 of \cite{yang2012feature}, the minimizer of \eqref{equ::update-p} above takes the form
\[
\bp^{s'+1} = \mathcal{L}_{\frac{\lambda_1}{\rho}}(\bT \bbeta^{s'+1} + \frac{1}{\rho} \bnu^{s'}).
\]

Finally, to update the dual variable $\bnu$, we follow the standard practice for the ADMM algorithm and set 
\[
\bnu^{s'+1} = \bnu^{s'} + \rho(\bT \bbeta^{s'+1} - \bp^{s'+1}).
\]
This complete the derivations for the ADMM algorithm in Algorithm \ref{alg:admm}.

\subsection{Additional simulation results} \label{section::appendix-simulations}

\subsubsection{Additional results for inferring the ADET in Section \ref{sec::infer-adet}}

\begin{figure}[h]
    \centering
    \includegraphics[width=0.92\textwidth]{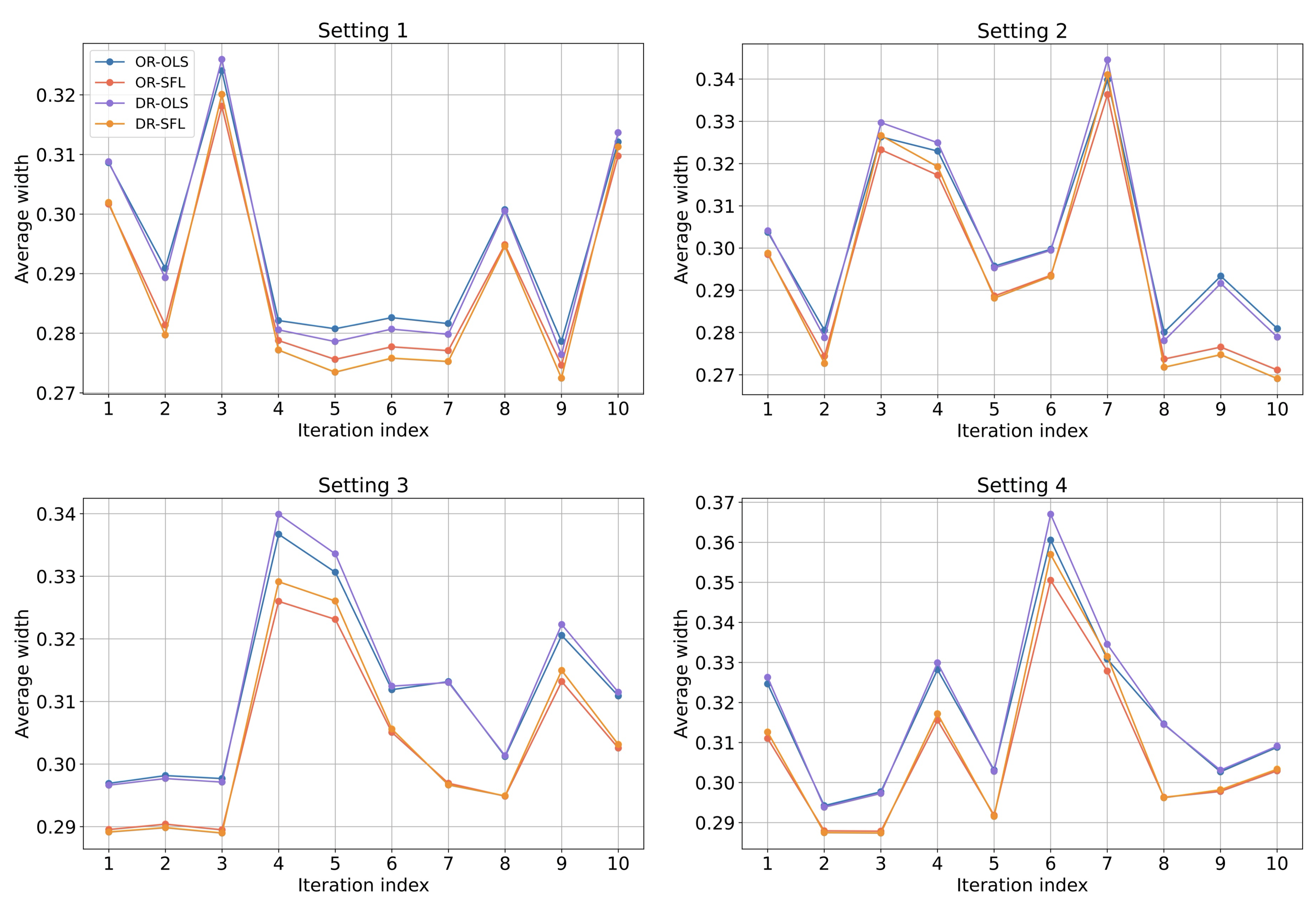}
    \vspace{-2em}
    \caption{Average widths of $1000$ confidence intervals for different methods across ten repetitions when $k=4$. 
    }
    \label{fig::er.sbm_g1.2_CIwidth}
\end{figure}

As depicted in Section \ref{sec::infer-adet}, the SFL method is preferred in overfitting scenarios with valid inference procedure and shorter CIs.
More explicitly, we demonstrate this using the first $10$ repetitions in our experiments.
For each repetition, we compute the average width of CIs within the $1000$ replications.
The results are displayed in Figure \ref{fig::er.sbm_g1.2_CIwidth}. 
It is seen that the SFL method consistently exibits shorter CIs than the OLS method across the $10$ repetitions.

\subsubsection{No exact matching scenario}

We now consider the settings \textit{without exact matching} where the interference function values of two nodes within the same group (given $k_0$) are \textit{approximately} centered around a common value, as opposed to being identical.
We employ the same settings as in Section \ref{sec::infer-adet} and modify the interference function by adding a perturbation around the original value 
$$f_\text{new}(T_{i,1}, T_{i,2}) = f(T_{i,1}, T_{i,2}) + \delta_{g(i)},$$ where $g(i)$ specifies the group that node $i$ belongs to and $\delta_{g(i)} \sim \mathrm{N}(0,\sigma_{g(i)})$ for $\sigma_{g(i)} \sim \mathrm{Uniform}(0,1)$.
For the four settings, we compute $1000$ CIs based on $1000$ replications of potential outcomes with each method, and repeat the entire procedure $100$ times.
Note that our methods and theoretical results naturally handle the case when $\sigma_{g(i)}$ is the same across all nodes, as the noise term $\delta_{g(i)}$ can be absorbed into $\epsilon_i$.
Our simulation results empirically demonstrate the good performance of our methods under heterogeneous $\delta_{g(i)}$.

\begin{figure}[H]
    \centering
    \includegraphics[width=0.91\textwidth]{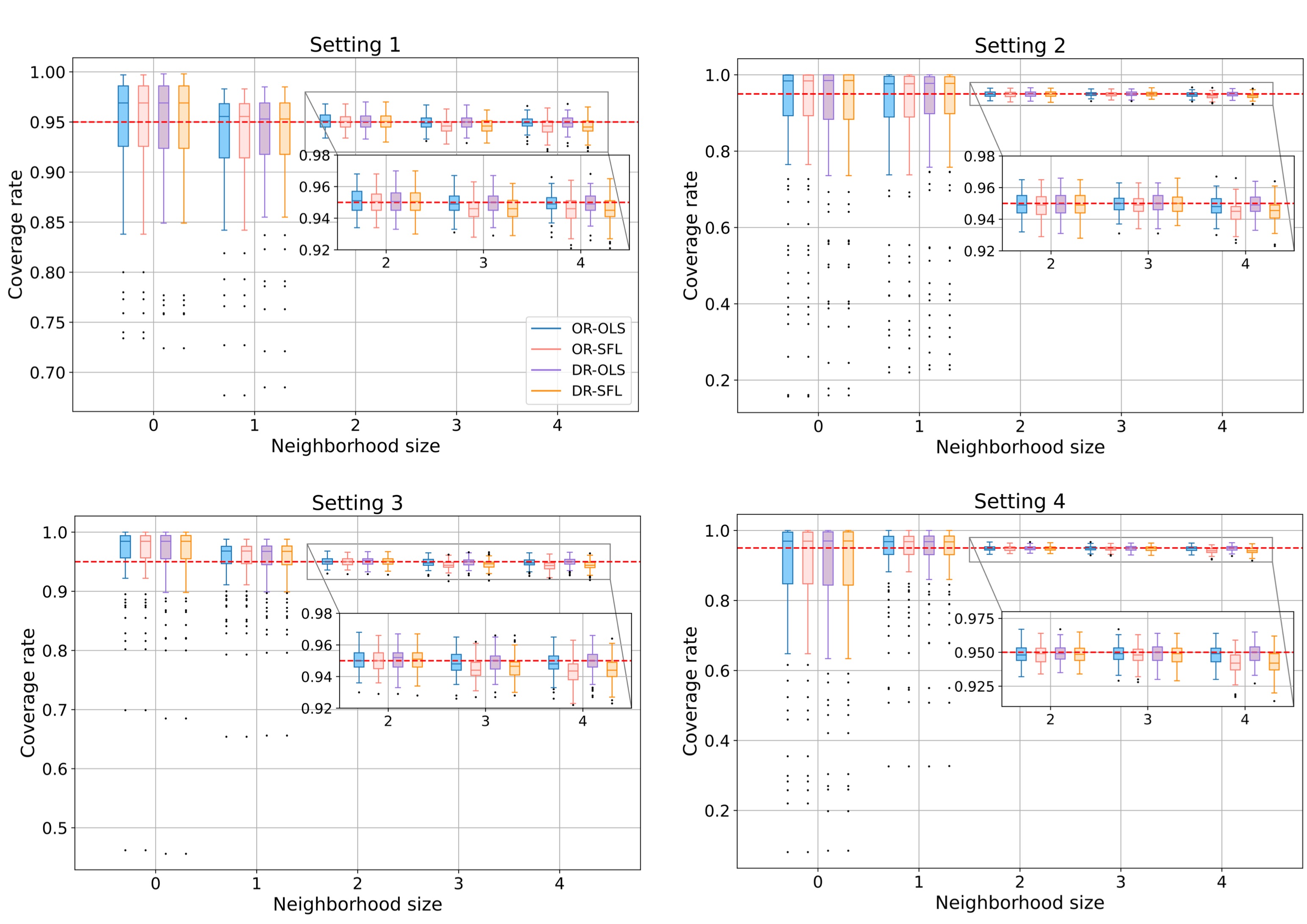}
    \vspace{-2em}
    \caption{Empirical coverage probabilities of different methods across $100$ repetitions without exact matching.
    }
    \label{fig::er.sbm_g1.2_cp_mismatch}
\end{figure}

As shown in Figure \ref{fig::er.sbm_g1.2_cp_mismatch}, when $k \geq k_0$, our methods consistently achieve good coverage rates. 
For $k < k_0$, the empirical coverage probability is closer to the nominal level compared to that in Figure \ref{fig::er.sbm_g1.2_cp}, with fewer values close to zero.
This could result from the added perturbation under the setting without exact matching, and consequently, leading to a much larger average CI width.
The average widths of CIs under settings without exact matching are shown in Table \ref{tab::average-mean-CI-length-across-100-iterations-mismatch}.

\begin{table}[h] 
    \caption {Average confidence interval widths of different methods without exact matching.} 
    \vspace{0.5em}
    \centering
    \resizebox{\textwidth}{!}{
    \begin{tabular}{ll|ccccc||ccccc}
    \toprule
    \multirow{2}{*}{} & \multirow{2}{*}{Method} & \multicolumn{5}{c||}{Mapping $\gamma_1$}  & \multicolumn{5}{c}{Mapping $\gamma_2$} \\
    \cmidrule{3-12}
      & & $k=0$ & $k=1$ & $k=2$ & $k=3$ & $k=4$ & $k=0$ & $k=1$ & $k=2$ & $k=3$ & $k=4$ \\
    \midrule
    \multirow{4}{*}{Graphon 1} & OR - OLS & 0.6181 & 0.5158 & 0.457 & 0.4647 & 0.4646 & 0.9284 & 0.7719 & 0.4627 & 0.4633 & 0.4684 \\
                               & OR - SFL & 0.6181 & 0.5158 & 0.4567 & 0.4579 & 0.4576 & 0.9284 & 0.7718 & 0.4621 & 0.4616 & 0.4600 \\
                               & DR - OLS & 0.6191 & 0.5169 & 0.4580 & 0.4657 & 0.4656 & 0.9310 & 0.7738 & 0.4643 & 0.4648 & 0.4699 \\
                               & DR - SFL & 0.6191 & 0.5169 & 0.4576 & 0.4588 & 0.4585 & 0.9310 & 0.7738 & 0.4636 & 0.4631 & 0.4615 \\
    \midrule
    \multirow{4}{*}{Graphon 2} & OR - OLS & 0.6980 & 0.5631 & 0.4730 & 0.4792 & 0.4830 & 0.8095 & 0.6712 & 0.4661 & 0.4665 & 0.4742 \\
                               & OR - SFL & 0.6980 & 0.5631 & 0.4726 & 0.4722 & 0.4717 & 0.8095 & 0.6711 & 0.4654 & 0.4648 & 0.4624 \\
                               & DR - OLS & 0.7016 & 0.5664 & 0.4757 & 0.4820 & 0.4858 & 0.8128 & 0.6738 & 0.4680 & 0.4684 & 0.4761 \\
                               & DR - SFL & 0.7016 & 0.5664 & 0.4753 & 0.4749 & 0.4744 & 0.8128 & 0.6737 & 0.4674 & 0.4668 & 0.4643 \\
    \bottomrule
    \end{tabular}
    }
\label{tab::average-mean-CI-length-across-100-iterations-mismatch}
\end{table}

We observe similar trends in Table \ref{tab::average-mean-CI-length-across-100-iterations-mismatch} and Table \ref{tab::average-mean-CI-length-across-100-iterations}. 
Specifically, whether under the exact matching scenario or not, the square-root fused clipped Lasso (SFL) method generally produces shorter CIs compared to the OLS, particularly for larger values of $k$. 
As noted above, due to the added perturbation in the setting without exact matching, the CIs in Table \ref{tab::average-mean-CI-length-across-100-iterations-mismatch} are consistently wider than those in Table \ref{tab::average-mean-CI-length-across-100-iterations}. This contributes to higher empirical coverage probability when $k < k_0$ in Figure \ref{fig::er.sbm_g1.2_cp_mismatch} compared to Figure \ref{fig::er.sbm_g1.2_cp}.

Although our methods are not guaranteed to be robust in scenarios without exact matching (according to the current theoretical results), the numerical studies indicate that they may handle certain types of mismatches and extend to heterogeneous noise settings by treating $\delta_{g(i)}$ as part of $\epsilon_i$.

\subsubsection{Misspecified propensity score}

\begin{figure}[h]
    \centering
    \includegraphics[width=\textwidth]{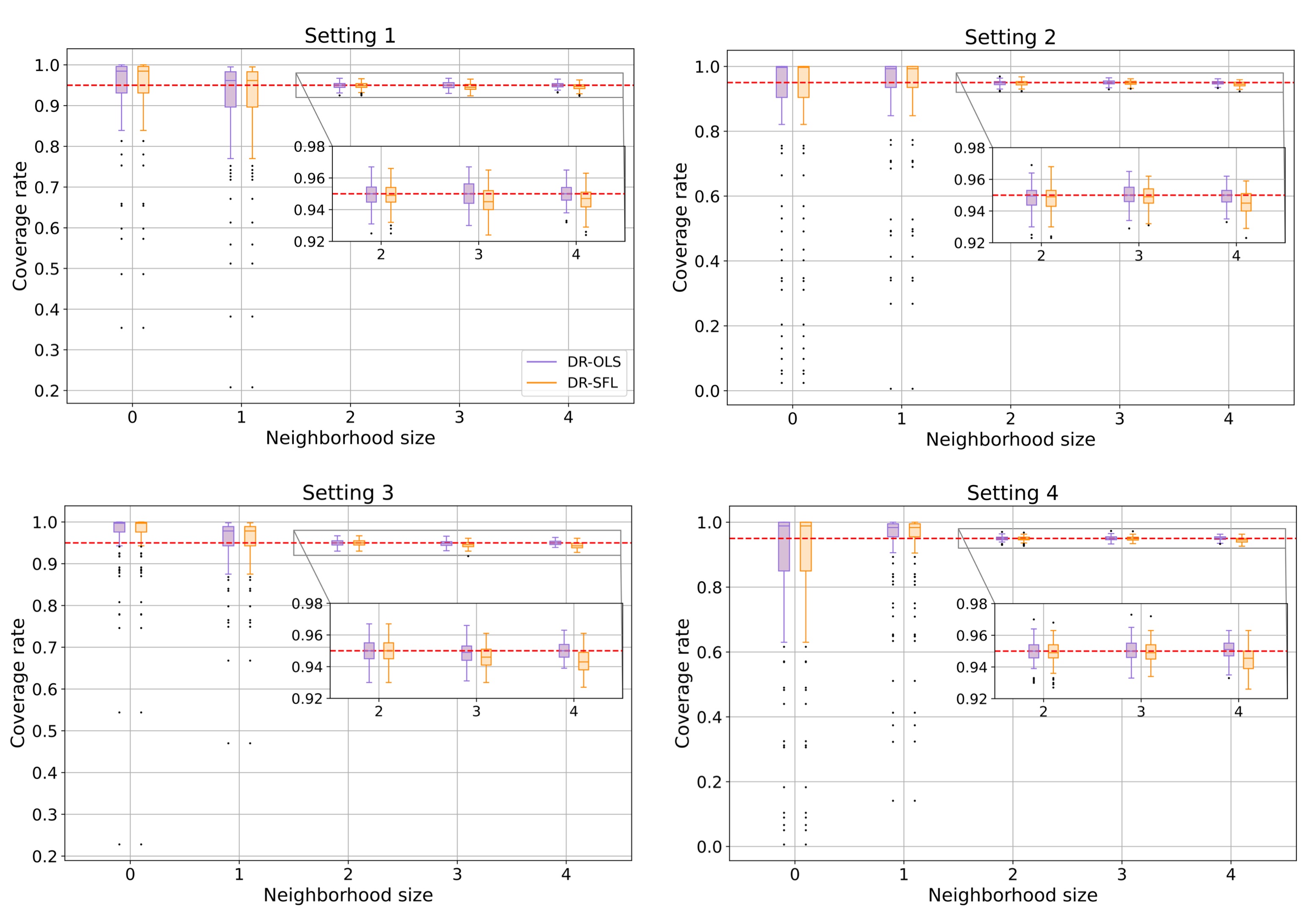}
    \vspace{-2em}
    \caption{Empirical coverage probability of the DR estimator across $100$ repetitions.
    }
    \label{fig::er.sbm_g1.2_cp_avg_p}
\end{figure}

Since specifying a model for the propensity scores is beyond the scope of the current paper, we now investigate the performance of the DR estimator when setting the overall treated proportion in the network as the propensity score shared across all nodes.
Again, we employ the four settings as in Section \ref{sec::infer-adet}, generate $1000$ replications of potential outcomes $\{Y_i\}$, and infer the ADET by applying different methods.
The procedure is repeated $100$ times  and the empirical coverage probabilities of the DR estimator are shown in Figure \ref{fig::er.sbm_g1.2_cp_avg_p}.

The inference performance based on the DR estimator in Figure \ref{fig::er.sbm_g1.2_cp_avg_p} is comparable to its performance depicted in Figure \ref{fig::er.sbm_g1.2_cp}.
Specifically, when $k \geq k_0$, the empirical coverage probability of the DR estimator is close to the nominal level across various network and treatment assignment configurations, while the inference results can be poor under certain designs when $k < k_0$.
Note that when $k < k_0$, given the interference network and treatment assignments $\bZ$, the DR estimator is not unbiased due to (i) the estimation of the interference function values can be biased and (ii) the ADET depends on $\bZ$ by definition.
In general, when $k \geq k_0$, we expect that the DR estimator performs well as long as the heterogeneity among the propensity scores is not severe.
Under this scenario, the propensity scores can be approximated by the overall treated proportion in the network, as shown in Figure \ref{fig::er.sbm_g1.2_cp_avg_p}.

\section{Additional real data example details}
\label{sec.supp.realdata}

In this section, we provide additional details of the real data application.
The cohort study conducted at a secondary school in Glasgow between 1995 and 1997 \citep{michell1997girls} collected three waves of survey data, focusing on changes in the smoking behaviors and substance use over time, as well as the influence of social interactions. 
Different from the focus in the previous studies, we are interested in making inference on the \textit{network} causal effect of being in a romantic relation on the substance use (alcohol, tobacco, cannabis) during the last wave of the study.
The raw data is publicly available at \url{https://www.stats.ox.ac.uk/~snijders/siena/Glasgow_data.htm}.

\begin{figure}[h]
    \centering
    \includegraphics[width=\textwidth]{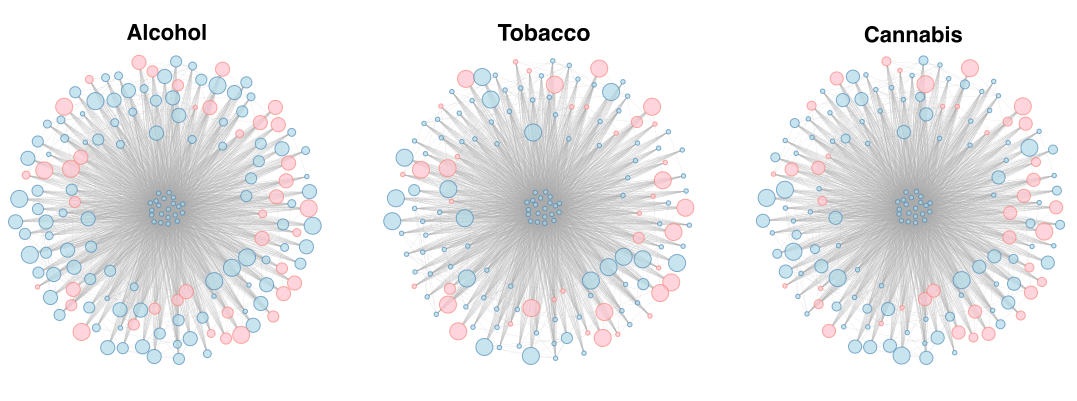} 
    \vspace*{-1cm}
    \caption {Visualization of the interference network with treated nodes in pink and untreated nodes in blue.
    A larger node size within the network signifies a higher frequency of alcohol consumption/tobacco use/cannabis use of an individual.} \label{fig::glasgow-w3} 
\end{figure}

For visualization, we temporarily rescale the substance use variables to $[1,5]$-valued through a linear transformation.
The friendship network between teenagers together with their treatment and substance use information is depicted in Figure \ref{fig::glasgow-w3}.

We observe a set of untreated nodes that never drink alcohol or use tobacco/cannabis, located at the center of the three plots in Figure \ref{fig::glasgow-w3}.
Besides them, the substance use of each individual varies across the three types of substance use and exhibits heterogeneous patterns for the same substance, particularly for the tobacco and cannabis use.
From the plots alone, it is challenging to determine whether the ADET of the romantic relationship on the frequency of substance use differs from zero with confidence due to the complex network dependence structure.

\end{document}